# Miniature soft robot with magnetically reprogrammable surgical functions

Chelsea Shan Xian Ng[1]†, Yu Xuan Yeoh[1]†, Nicholas Yong Wei Foo[1]†,
Keerthana Radhakrishnan[2], Guo Zhan Lum[1,2]*

[1]School of Mechanical and Aerospace Engineering, Nanyang Technological University,
50 Nanyang Avenue, Singapore 639798, Singapore

[2]Rehabilitation Research Institute of Singapore, Nanyang Technological University,
11 Mandalay Road, Singapore 308232, Singapore

†Equally contributing authors

*Correspondence to: gzlum@ntu.edu.sg

## Abstract

Miniature robots are untethered actuators, which have significant potential to make existing minimally invasive surgery considerably safer and painless, and enable unprecedented treatments because they are much smaller and dexterous than existing surgical robots. Of the miniature robots, the magnetically actuated ones are the most functional and dexterous. However, existing magnetic miniature robots are currently impractical for surgery because they are either restricted to possessing at most two on-board functionalities or having limited five degrees-of-freedom (DOF) locomotion. Some of these actuators are also only operational under specialized environments where actuation from strong external magnets must be at very close proximity (< 4 cm away). Here we present a millimeter-scale soft robot where its magnetization profile can be reprogrammed upon command to perform five surgical functionalities: drug-dispensing, cutting through biological tissues (simulated with gelatin), gripping, storing (biological) samples and remote heating. By possessing full six-DOF motions, including the sixth-DOF rotation about its net magnetic moment, our soft robot can also roll and two-anchor crawl across challenging unstructured environments, which are impassable by its five-DOF counterparts. Because our actuating magnetic fields are relatively uniform and weak (at most 65 mT and 1.5 T/m), such fields can theoretically penetrate through biological tissues harmlessly and allow our soft robot to remain controllable within the depths of the human body. We envision that this work marks a major milestone for the advancement of soft actuators, and towards revolutionizing minimally invasive treatments with untethered miniature robots that have unprecedented functionalities.

## Significance

Magnetic miniature robots have great potential to transform minimally invasive surgery because they are much smaller and more dexterous than existing surgical robots. However, such actuators are currently impractical for surgery because they are either restricted to possessing at most two on-board functions, having five degrees-of-freedom (DOF) locomotion, or are only operational in specialized environments. Here we present a millimeter-scale soft robot, which can reprogram its magnetization profile to dispense drugs, cut, grip and store biological samples, and heat remotely. Our six-DOF actuator can also negotiate across unstructured environments, which are impassable by its five-DOF counterparts. We envision that this study presents a significant advancement for soft actuators and miniature robots, and it marks a major milestone towards revolutionizing medicine.

## Main text

Miniature robots are untethered actuators of millimeter length scale or smaller (*1–16*). Because these actuators are much smaller and more dexterous than existing macro-scale surgical robots, they have significant potential to make minimally invasive surgery considerably safer and painless (*7, 17–19*). Furthermore, as miniature robots have greater access to highly confined and enclosed spaces in the human body than their macro-scale counterparts (*1, 7, 10, 17, 18*), they can prospectively enable unprecedented minimally invasive treatments in the future too. Of the miniature robots, the magnetically actuated ones are especially promising since they are known to be the most functional and dexterous (*1, 6, 8, 9, 11, 14, 16, 19–24*). Such attributes have been observed because unlike other classes of miniature robots that are actuated by heat (*25*), electricity (*4*), light (*12*), chemicals (*26*) and pressure (*27*), the actuating fields of these magnetic miniature robots can be specified not only in magnitude but also in their directions and spatial gradients (*8, 28, 29*). Since magnetic fields can harmlessly penetrate through biological tissues, such an actuation method is also highly compatible for the targeted medical applications of miniature robots (*7, 17–19, 30*).

There exist two classes of magnetic miniature robots. The first class possesses slime- or liquid-based bodies, and these miniature robots can be actuated by external magnetic fields to actively deform their bodies and locomote (*31–36*). Although such emerging actuators have attracted significant attention within the robotics community, their non-solid bodies do not have the necessary structural integrity to output sufficient forces for surgery and they may leave residues of their body in the environment (*31–36*). Such attributes are generally undesirable for minimally invasive treatments (*30*). While there exist slime- or liquid-based robots that can stiffen their body via phase change (*32*) or under actuation by strong magnetic fields (*33*), such mechanisms have several critical drawbacks which render them impractical for surgery. Specifically, these actuators require very strong, non-uniform magnetic fields (20–150 mT, 8–10 T/m) to stiffen up (*33*) or to reach a desired geometry before they are frozen (*32*).

Therefore, they can only operate in specialized environments where actuation from strong external magnets must be at very close proximity (< 4 cm away) (*32, 33*). Again, such constraints are undesirable for surgery since miniature robots may have to access in-depth regions within the human body for such treatments (> 5 cm from the skin surface) (*37*). For the slime-based robots to induce their phase-changing mechanisms, they must be heated beyond 60°C (*32*) and such temperatures may also cause cell death within the patients' body (*38, 39*). Conversely, the second class of magnetic miniature robots have solid elastomeric bodies which do not leave residues of their body in the environment, and they can be actuated wirelessly from a distance (< 30 cm) (*40*). While such features are fundamentally compatible for medical treatments, magnetic miniature robots with elastomeric bodies are currently impractical for surgery as well. This is because such miniature robots cannot possess more than two on-board functions (*1–3, 5, 8, 9, 11, 14, 16, 20–25, 28, 29, 41–47*) and thus they are unable to execute complicated medical procedures in a surgery. Due to the size of miniature robots, it is very difficult to increase their number of on-board functions (*7, 20*). Indeed, even larger magnetic capsule robots, which are in the centimeter-scale, cannot realize more than two on-board functions too; for example, each magnetic capsule robot from Sun et al. can at most execute two of the following functions: dispense drugs, store samples or heat remotely (*24*). Moreover, most of the magnetic miniature robots with elastomeric bodies can only achieve a maximum of five degrees-of-freedom (DOF) motions because they cannot control their sixth-DOF rotation about the net magnetic moment (*1–3, 5, 9, 11, 14, 16, 20–25, 41–47*). Hence, while these actuators can translate along three axes, they can only rotate about two axes (*1–3, 5, 9, 11, 14, 16, 20–25, 41–47*). Without the sixth-DOF rotation, it is very challenging for the five-DOF actuators to negotiate across the unstructured environments in the human body (*8, 15, 28*). In comparison, miniature robots with six-DOF are much more dexterous as they can move across obstacles which are otherwise impassable by their five-DOF counterparts (*8, 15, 28*). Although there exist magnetic miniature robots with six-DOF, these rare actuators can only grasp (*8, 28*) or dispense drugs (*15*), and such simple functionalities are insufficient to perform complicated surgeries. It remains a great challenge to create magnetic miniature robots, which can operate in unstructured environments while possessing more than two on-board functions and full six-DOF motions. If such actuators can be created, they would have great potential to transform surgery.

Here we present a millimeter-scale soft robot where its magnetization profile can be reprogrammed upon command to perform five surgical functions: drug-dispensing, cutting through biological tissues (gelatin), gripping, storing (biological) samples and remote heating. By possessing full six-DOF motions, our soft robot can also negotiate across challenging, unstructured environments, which are impassable by its five-DOF counterparts (*1–3, 5, 9, 11, 14, 16, 20–25, 41–47*). Our soft robot is made of smart magnetic composites that have the necessary structural integrity to potentially output sufficient forces for surgery and they will not leave body residuals in the environment (Supplementary

Information (SI) Section S1A). Hence, it is more practical for medical treatments than those with slime- or liquid-based bodies (*31–36*). Furthermore, the external surfaces of our soft robot can potentially be made biocompatible via coating techniques (SI Section S1B).

Our soft robot has a main body, two soft tentacles, a reprogrammable module, and a remote heating component (Fig. 1**a** and SI Section S1A). Within the main body, there are three soft deformable beams that are arranged in a rotary symmetrical configuration (Fig. 1**b**). These soft beams can be selectively activated to dispense drugs, cut, grip and store objects (Fig.s 1**b-c**, S5). The magnetization profiles within the soft robot's main body and tentacles are pre-programmed while the magnetic moments of the reprogrammable module and remote heating component can be magnetized or demagnetized upon command (Fig. 1**b**). Based on the magnetic moments of the reprogrammable module and remote heating component, the proposed soft robot can operate under several modes: its locomotion or specific function modes. When the reprogrammable module and remote heating component are demagnetized, the soft robot will assume its locomotion mode such that it can actively control the deformation of the soft tentacles to adopt different gaits (Fig. 1**d** and SI Video S1). The soft robot can also quickly switch to a specific function mode by magnetizing its reprogrammable module and remote heating component with a uniform magnetic field ($\vec{B}$) of 60 mT (< 1 s) (Fig. 1**d**(ii), SI Video S1 and SI Section S1C). An angle $\varphi$ is defined to describe the direction of these components' magnetic moment (Fig. 1**b**(ii)). When the reprogrammable module and remote heating component are magnetized along the axes where $\varphi$ is equal to $90°$, $330°$ or $210°$, our soft robot is reprogrammed to its drug-dispensing, cutting or gripping/storage modes, respectively (Fig. 1**b**). For example, after the reprogrammable module and remote heating component possess magnetic moments along the axis where $\varphi$ is equal to $330°$ (Fig. 1**c**(i), **d**(ii) and SI Video S1), our soft robot is reprogrammed to its cutting mode and it can be actuated by a magnetic field of 15 mT to expose a retractable cutting tool (Fig. 1**c**(ii), **d**(iii), SI Video S1 and SI Section S1C). To revert to the locomotion mode, the reprogrammable module and heating component can be demagnetized quickly (< 1 s) via an alternating $\vec{B}$ of 45 Hz, which has a linearly decaying amplitude (Fig. 1**d**(iv), (vii), (x), SI Video S1 and SI Section S1C). Using similar reprogramming strategies, our soft robot can also be reprogrammed to activate its drug-dispensing, gripping and storage functions (Fig.s 1**d**(v)-(vi), (viii)-(ix), S5, SI Video S1 and SI Section S1C). For each function mode, a corresponding pair of soft beams in the main body will bend elastically to activate a surgical function when the applied $\vec{B}$ has a component parallel to the magnetic moments of the reprogrammable module and remote heating component (SI Sections S2A, S3B). This component of $\vec{B}$ is denoted as $\vec{B}_{\mathrm{Func}}$ (Fig.s 1**c**, S5). The soft beams will bend because the magnetic moments located at their free ends will always tend to align with the applied $\vec{B}_{\mathrm{Func}}$, and this will in turn induce a magnetic torque necessary for these deformations (SI Section S2A). Except for Fig. 1 and SI Video S1, we document all our actuating $\vec{B}$ in Fig.s S19-35 to allow the figures in the main text and videos to better illustrate our actuator's motions.

During the reprogramming processes, our soft robot will remain stationary and not produce uncontrollable motions (SI Video S1). This is a critical feature because the actuator must remain controllable by surgeons during treatments. The magnetization profile within the main body and tentacles of our soft robot will not be re-magnetized during the reprogramming processes because these components have magnetic coercivities of 93.3–614 mT (SI Section S1A), which are higher than those of our reprogramming $|\vec{B}|$ (60–65 mT). Once magnetized, the reprogrammable module and the remote heating component will have functional magnetization magnitudes of 7.18 kA/m and 6.52 kA/m, respectively (SI Section S1A). The magnetization magnitudes of the reprogrammable module and the remote heating component are reduced by 6.03- and 8.51-fold, respectively, upon demagnetization and become negligible (SI Section S1A). The key difference between the remote heating component and the reprogrammable module is that the induction heating power of the former is fourfold of the latter when an alternating $\vec{B}$ of high frequency (amplitude: 9.34 mT, frequency: 75.4 kHz) is applied (SI Section S1A). Therefore, the remote heating component is placed at the top of our soft robot to better facilitate potential hyperthermia treatments while the reprogrammable module can serve as a heat insulator base to store the drugs and biological samples (Fig. 1**a**). By having a diverse range of magnetic responses, our soft robot can therefore realize mechanical functionalities that are far beyond those of all the existing miniature robots (*1–5, 8, 9, 11–16, 20–25, 28, 29, 31–36, 41–47*).

To describe the motions of our soft robot when it is in the locomotion mode, we first attach a local coordinate frame to its body where all the axes in this frame are denoted with the subscript: $\mathrm{Loco},\{L\}$ (Fig. 2**a**(i)). When a $\vec{B}$ is applied along the soft robot's $Z_{\mathrm{Loco},\{L\}}$-axis, the actuator's tentacles will deform into either an upright or inverted 'U'-shape (SI Section S2A). In the deformed configurations, the soft robot possesses a net magnetic moment, which is parallel to the $Z_{\mathrm{Loco},\{L\}}$-axis (SI Section S2). The soft robot's net magnetic moment will increase when the tentacles deform more, and there is a positive correlation between the tentacles' curvature and the magnitude of the applied $\vec{B}$ (SI Sections S2A, S3A). As the soft robot's net magnetic moment will always tend to align with $\vec{B}$, this actuator will be able to generate two axes of rotation via controlling the direction of $\vec{B}$ (e.g., magnitude: 15 mT in Fig. 2**a**, SI Video S2 and SI Sections S2B, S4A). However, to rotate about the soft robot's net magnetic moment (i.e., the sixth-DOF axis (*8, 28, 29*)), appropriate magnetic spatial gradients must be applied (Fig. 2**b**, SI Video S2 and SI Sections S2B, S4A). Our soft robot can achieve angular velocities of 1.56–16.5 rad/s when it rotates about its three independent axes (Fig.s 2**a-b**, S13**a**, SI Video S2 and SI Section S4A). By applying other magnetic spatial gradients (0.21–0.31 T/m), the soft robot can also translate along three independent axes with velocities of 0.126–0.911 mm/s (Fig. 2**c**, SI Video S2 and SI Section S4A). The experiments in Fig. 2 show that our actuator has six-DOF motions since it can produce three independent axes of rotations and translations (*8, 15, 28, 29*). To levitate the soft robot against gravity in an aquatic environment, a buoyant component is added to this actuator to increase its buoyancy (Fig. 2**c**, SI Video S2 and SI Section S4A). The buoyant component is required due to the limitations of our electromagnetic coil system (SI Section S1D), and not by the working principles of our soft robot. In theory, our soft robot can levitate against gravity with a magnetic spatial gradient of 4.39 T/m in air and such actuation signals can be generated by stronger coil systems (SI Section S4A).

In its locomotion mode, our soft robot can roll and two-anchor crawl. Rolling is one of the fastest modes of terrestrial locomotion (*1, 10, 28*). Because our soft robot has six-DOF, it can choose to roll along its length or width, and steer its rolling direction via controlling its sixth-DOF angular displacement (Fig. 2**d-e**, SI Video S3 and SI Section S4B). The soft robot can achieve speeds of 1.29–6.75 mm/s with this gait (a rotating $\vec{B}$ of 20 mT and 0.1–1 Hz, SI Section S4B). Rolling along the width of the actuator is slower than rolling along its length but its smaller rolling radius can allow the soft robot to squeeze through narrower openings. Our soft robot is therefore very robust; it can roll along its length to quickly maneuver across obstacle-free terrains, or it can roll along its width if it wishes to overcome barriers with narrow openings (Fig. 2**f**, SI Video S3 and SI Section S4B). Our soft robot can also two-anchor crawl by alternating the free ends of its tentacles as anchors (Fig.s 2**g**, S16**a**, SI Video S4 and SI Section S4C). This gait can be activated by rotating $\vec{B}$ about the soft robot's $X_{\mathrm{Loco},\{L\}}$-axis with varying magnitudes (8–22 mT). Based on this gait, the soft robot can achieve speeds of 0.4–1.06 mm/s (SI Section S4C). Because our soft robot has six-DOF and its stride length can be precisely controlled via the two-anchor crawling locomotion, it can steer its crawling direction, climb up a slope and move across obstacles with strict shape constraints (Fig. 2**g-i**, SI Video S4 and SI Section S4C). By having such high dexterity, our soft robot will have great potential to negotiate across the highly unstructured environments within the human body.

After our soft robot reaches its destination via the locomotion mode, it can be reprogrammed to a desired surgical function mode. While our soft robot is unable to execute the two-anchor crawling locomotion in its function modes, it can still realize rolling or spin-walking gaits with six-DOF (SI Sections S2B, S5C). Rolling is faster than spin-walking, but the latter gait can allow the soft robot to maintain an upright orientation during locomotion. Using such gaits, our soft robot can roll to a targeted location and dispense a simulated solid drug when it is reprogrammed to its drug-dispensing mode (Fig. 3**a**, SI Video S5 and SI Section S5A). If the soft robot is required to deliver liquid medicine, such drugs can potentially be stored in a biodegradable capsule so that the actuator can deliver this solid cargo to the targeted location (*13*). From the medical perspective, it is highly advantageous to deliver drugs with miniature robots because these actuators can potentially increase the delivery efficiency by 78.6-fold compared to traditional delivery modes that transport the drugs via the human's circulatory system (*15,*

*43, 48*). The maximum capacity of drugs that our soft robot can carry is theoretically sufficient for targeted drug delivery treatments as the amount of drugs that it can carry is twofold of what is required for practical targeted chemotherapy applications (*43*) (SI Section S5A).

By reprogramming the soft robot to its cutting mode, it can potentially perform incision to remove tumors or unwanted tissues. A retractable cutting tool hidden in the soft robot's main body can protrude when a $\vec{B}_{\mathrm{Func}}$ of 29.2 mT is applied (Fig. 3**b**(ii), SI Video S6 and SI Sections S3B, S5B). After the cutting tool protrudes, our actuator can use this tool to penetrate through gelatin, which is derived from biological tissues (applied magnetic spatial gradients: 1.5 T/m in Fig. 3**b**(iii), SI Video S6 and SI Section S5B). Since this gelatin possesses similar fracture toughness to those of the soft tissues in the human brain and liver (*49*) (SI Section S5B), this suggests that our soft robot has great potential to generate sufficient pressure for its cutting tool to perform incision on human tissues. A notable advantage of our cutting mechanism is that the retractable cutting tool can be safely hidden within the soft robot's body. Hence, unlike existing miniature robots that expose the sharp edges of their blades (*44*), our soft robot can ensure that its cutting tool will not accidently cut any tissue as it moves in the human body.

Our soft robot can also be reprogrammed to enable gripping functions; a pair of grippers will extend out of the soft robot's main body when a $\vec{B}_{\mathrm{Func}}$ of 25 mT is applied (Fig. 3**c**-**d**). We can control the grippers accurately because we have characterized their extensions with respect to different $|\vec{B}_{\mathrm{Func}}|$ (SI Sections S3B, S5C). To demonstrate the gripping function, we command the soft robot to pick up an object and subsequently transport it to the targeted location via the spin-walking locomotion (Fig. 3**c**, SI Video S7 and SI Section S5C). The gripping mechanism can potentially assist an incision procedure by adjusting a partially carved-out tissue to a more favorable orientation before resuming the cutting process. This can be an important step during incision since it can help surgeons to avoid removing healthy tissues from the patients. Furthermore, the gripping function can also allow our soft robot to pick up and subsequently encase the excised tissue (Fig. 3**d**(i)-(iii), SI Video S7 and SI Section S5C). After the excised tissue is stored within the actuator's main body, the soft robot can be reprogrammed to its locomotion mode again (Fig. 3**d**(iii)-(iv), SI Video S7 and SI Section S5C). This ensures that the gripping function will not be accidentally activated to release the encased tissue, and that the soft robot can safely store and transport this sample out for biopsy. The volume of samples that our soft robot can carry is theoretically also sufficient for biopsy purposes; our sample's volume in Fig. 3**d** is 1.78-fold of those required for deoxyribonucleic acid (DNA) extraction and polymerase chain reaction (PCR) amplification (*25*) (SI Section S5C).

Our actuator's remote heating function can be executed across all its locomotion or function modes (SI Section S5D). To enable this heating function, alternating $\vec{B}$ of high frequency is applied on the soft robot (amplitude: 9.34 mT, frequency: 75.4 kHz). Because the frequency of the actuating signals is too high for the soft robot to respond and produce any motion, we can effectively decouple the actuator's heating function from its locomotion via controlling the frequency of the applied $\vec{B}$. As demonstrated in Fig. 3**e** and SI Video S8, our soft robot can remain stationary while its temperature is raised from 26.3°C to 40°C within 35 s to trigger a color change in a thermochromic dye at 40°C (SI Section S5D). It is important to be able to raise the local temperature of a surface to 40°C as this is a necessary requirement to execute a wide range of hyperthermia treatments (*38, 39*).

The potential of deploying our soft robot for surgery is demonstrated via in vitro biological phantoms (Fig. 4, SI Videos S9-S10 and SI Section S6). In its locomotion mode, the six-DOF soft robot can roll across unstructured environments (Fig. 4**a**(i)-(ii), SI Video S9 and SI Section S6A). Once the soft robot reaches its destination, it can be reprogrammed to dispense simulated drugs at another targeted location (Fig. 4**a**(iii), SI Video S9 and SI Section S6A). Subsequently, the soft robot can be reprogrammed to its locomotion mode again so that it can two-anchor crawl towards a synthetic tumor (Fig. 4**a**(iv), SI Video S9 and SI Section S6A). By reprogramming to its cutting mode, the soft robot can then perform an incision to remove this tumor (Fig. 4**a**(v), SI Video S9 and SI Section S6A). To transport the excised tumor out for biopsy, the soft robot can be reprogrammed to its gripping mode to pick up and encase the biological sample within its main body (Fig. 4**a**(vi)-(vii), SI Video S9 and SI Section S6A). Finally, our soft robot can assume its locomotion mode and execute the two-anchor crawling gait to exit the phantom by backtracking to the entrance (Fig. 4**a**(viii), SI Video S9 and SI Section S6A). Likewise, the soft robot can also roll to a desired location and perform hyperthermia treatments (simulated by the color change of a thermochromic dye at 40°C in Fig. 4**b**, SI Video S10 and SI Section S6B). These demonstrations show that our six-DOF soft robot can navigate well in unstructured terrains, and it possesses the required surgical functions for performing minimally invasive treatments. Such abilities are unprecedented for magnetic miniature robots with elastomeric bodies; specifically, such existing actuators have at most two on-board functions (*1–3, 5, 8, 9, 11, 14–16, 20–25, 28, 29, 41–47*) and most of them possess limited five-DOF motions (*1–3, 5, 9, 11, 14, 16, 20–25, 41–47*).

## Discussion

A notable advantage of our soft robot is that it can operate under relatively weak, uniform fields of up to 65 mT and 1.5 T/m (Fig.s S20-35 and SI Sections S4-6), which are much easier to generate from a distance (*9*). As a result, our soft robot is not constrained to lab-based environments where actuation from external magnets must be at very close proximity like those with slime- or liquid-based bodies (< 4 cm away) (*32, 33*). Instead, our soft robot has great potential to be wirelessly controlled while it operates within the depths of the human body. Indeed, it is possible to wirelessly control miniature robots in a large spherical workspace (radius: 78.3 cm) with magnetic fields of 20–80 mT via remote magnets, which are placed at least five centimeters away from the actuators (*40, 50*). For this first study, we used a lab-based electromagnetic coil system (SI Section S1D), but we aim to redesign it to become more compatible for biomedical applications in subsequent studies. Our calculations indicate that our soft robot's actuating $\vec{B}$ can be made harmless to the patients (SI Section S7A), suggesting that it is indeed compatible for medical treatments.

For easy characterization purposes, we have created a larger prototype of our soft robot in this first study (length: 4.4 mm, SI Section S7B). Nonetheless, this prototype can potentially be deployed for treatments in the gastrointestinal tract since such medical procedures can be administered by centimeter-scale robots (*18*). It is possible to miniaturize our soft robot until its length is 2.5 mm, while still allowing it to execute all the functions and locomotion with an actuating $\vec{B}$ which is safe for patients and easy to generate at a distance (SI Section S7B). Using this miniaturized version, our soft robot can potentially access the central nervous system of approximately 50% of the human population (*7*), and it will have great potential to enable unprecedented minimally invasive surgery such as those for the brain. In contrast to existing surgical robots that have difficulties reaching a patient's inner brain (*7*), our soft robot can potentially non-invasively access this organ via the following procedure: first, inject this actuator into the spine, and subsequently navigate it through the central nervous system to reach the interior and exterior part of the brain via the cerebral aqueduct (*7*). If such novel procedures can be realized, surgeons can potentially exploit our actuator to develop a wide range of unprecedented brain surgeries. Our soft robot can potentially enable other types of radical minimally invasive surgery for the heart, ear and uterus too, and such hypotheses have been discussed by Nelson et al. (*7*).

In this study, we observe our soft robot's motions via a standard camera. However, we will use imaging techniques such as ultrasound, X-radiation (X-rays), computed tomography (CT) scans and magnetic resonance imaging (MRI) to observe its position and orientation when it is operating within the enclosed human body (*1, 7, 17, 18, 22, 30*). Ultrasound is especially promising as previous studies had shown that this feedback strategy has minimal interference with the soft robot's actuating magnetic fields (*1, 22*). We will explore these feedback strategies when our soft robot is deployed for future in vivo experiments. Using such imaging techniques, we also aim to automate our soft robot to further enhance its speed and positioning accuracy, which may be currently limited by our human operator's manual control.

Although our soft robot can also roll with six-DOF in its function modes, it will be better to execute this gait when the actuator is in its locomotion mode. This is because it guarantees that the surgical functions of the soft robot will not be accidentally activated when the actuator rolls in its locomotion mode. Nonetheless, we report that our soft robot is programmed in such a way that it will not activate the surgical functions when $|\vec{B}|$ is 3 mT or smaller (SI Section S3B). This feature allows us to decouple the soft robot's surgical functions and locomotion when it is in the function modes. The external surfaces of our soft robot can be coated with biocompatible thin films to become less adhesive in the future (SI Section S1B). For proof-of-concept purposes, here we have minimized the surface energy of the soft robot by conducting most of the experiments in oil-based environments.

In summary, here we present a miniature soft robot that is highly dexterous, and it can be magnetically reprogrammed to dispense drugs, cut, grip and store samples, and heat remotely. Our actuator has great potential to revolutionize surgery, and it also represents a significant advancement for soft actuators and small-scale robotics.

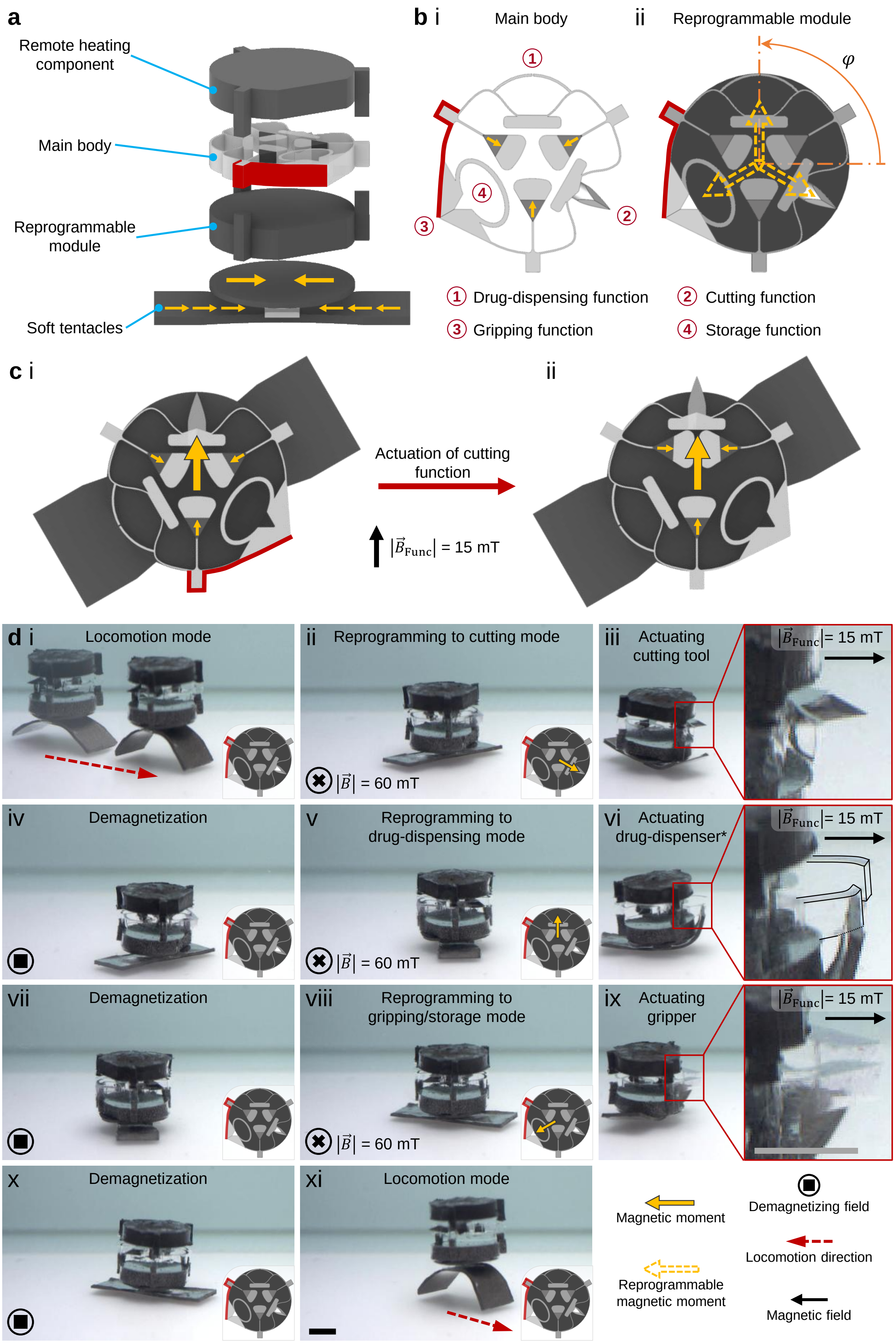

**Fig. 1 | The proposed soft robot and its reprogramming strategies. a,** The auxiliary exploded view of our soft robot. An edge on the main body is highlighted in red for easier identification of the soft robot's orientation from the top view. **b,** The top cross-sectional views of our soft robot where the red edge corresponds to the one shown in **a**. (i) Its main body has three soft beams that are arranged in a rotary symmetrical configuration. These deformable beams have a magnetized free end, and they are used to activate the soft robot's drug-dispensing, cutting, gripping and storage functions. (ii) The magnetic moments of the reprogrammable module and the remote heating component can be described by the angle, $\varphi$. When $\varphi$ is equal to 90°, 330° or 210°, the soft robot will be in its drug-dispensing mode, cutting mode or gripping/storage mode, respectively. The soft robot will assume its locomotion mode when the reprogrammable module and remote heating component have negligible magnetic moment. **c,** (i) As an example, the soft robot can be reprogrammed to its cutting mode when the magnetic moments of the reprogrammable module and remote heating component are magnetized to the axis in which $\varphi$ is equal to 330°. (ii) When $\vec{B}_{\mathrm{Func}}$ is applied, two beams in the actuator's main body will deform and extend the cutting tool out of its body. The red edge on the soft robot corresponds to the one shown in **a**. The orientation of the soft robot is adjusted such that its cutting tool can extend vertically, allowing such extensions to be visualized more easily. **d,** A demonstration of the soft robot's reprogramming process. In its function modes, the soft robot can be actuated to activate a respective surgical function ((iii), (vi), (ix)). Likewise, our actuator can actively control its soft tentacles when it assumes the locomotion mode ((i), (xi)). The demagnetizing fields are represented by a square in a circle, while the actuating and magnetizing fields are represented by the black arrows and a cross in a circle, respectively. The solid yellow arrows represent the magnetic moments in our soft robot while the dashed yellow arrows represent the potential magnetic moments that its reprogrammable module and remotely heating component should possess so that the soft robot can be reprogrammed to the desired function mode. The dashed red arrows represent the trajectory of our soft robot. To indicate the soft robot's operational mode, we include the magnetic moments of its reprogrammable module and remote heating component at the bottom right corner in the sub-panels of **d**. Scale bars, 1 mm.

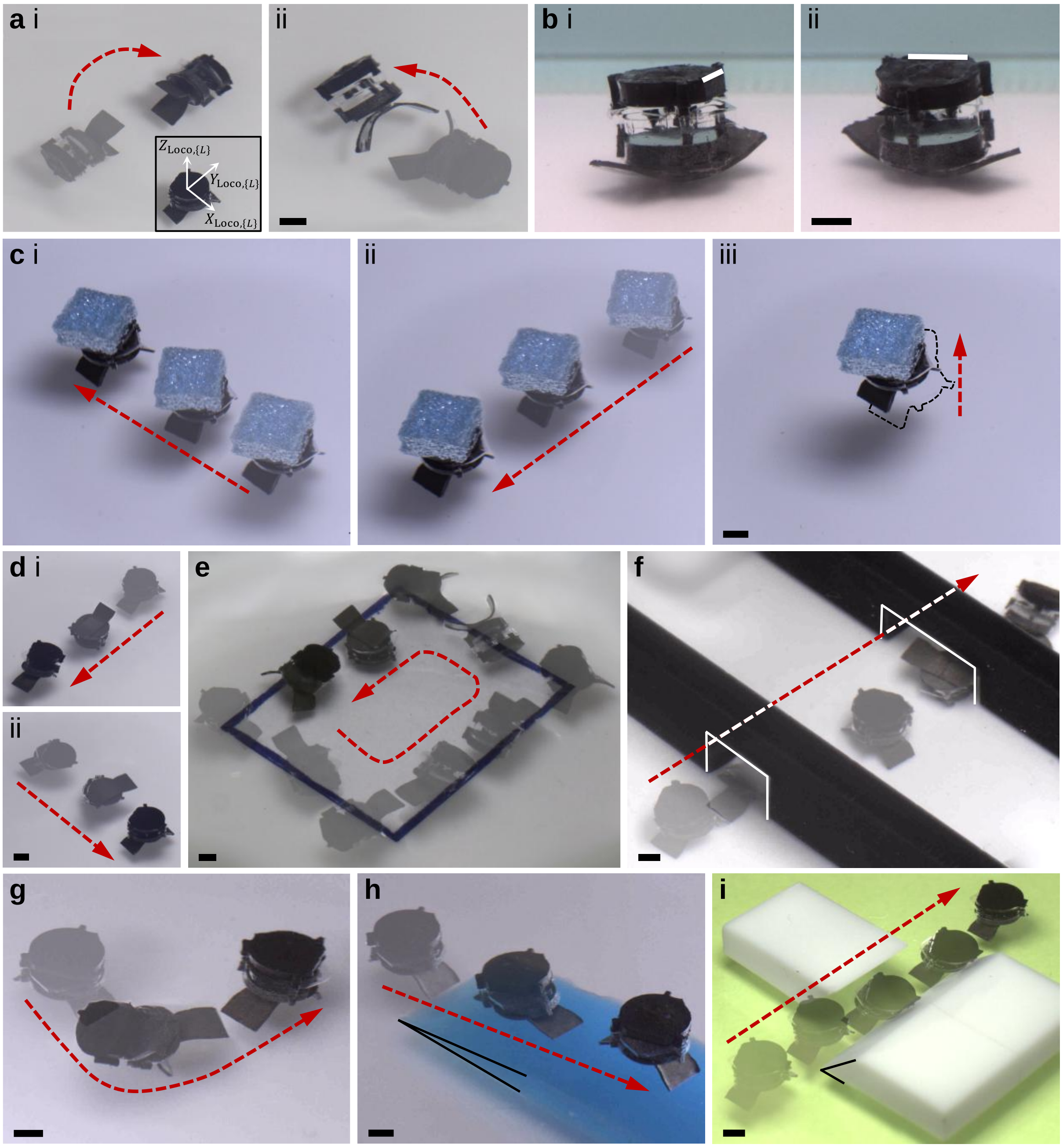

**Fig. 2 | Locomotion of the soft robot when it is reprogrammed to its locomotion mode. a,** (i) and (ii) demonstrate that the soft robot can rotate about its $X_{\mathrm{Loco},\{L\}}$- and $Y_{\mathrm{Loco},\{L\}}$-axes, respectively. The local coordinate frame of the soft robot is shown at the bottom right corner of (i). **b,** To illustrate the soft robot's sixth-DOF rotation (i.e., about its $Z_{\mathrm{Loco},\{L\}}$-axis), we add a white line to the top of the actuator, highlighting its sixth-DOF angular displacements from two different snapshots in SI Video S2. **c,** (i)-(iii) illustrate that the soft robot can translate along its $X_{\mathrm{Loco},\{L\}}$-, $Y_{\mathrm{Loco},\{L\}}$- and $Z_{\mathrm{Loco},\{L\}}$-axes, respectively. The dotted outline in (iii) represents the actuator's initial position. **d,** (i)-(ii) show that the soft robot can roll along its length and width, respectively. **e,** The soft robot can steer its rolling direction as it rolls along its length and width to track a path outlined by the dark blue lines. **f,** Using its six-DOF motions, the soft robot can roll across challenging barriers. **g,** When the soft robot executes the two-anchor crawling locomotion, its propulsion direction can be steered too. **h,** The soft robot can use its two-anchor crawling gait to ascend a slope of 15°. **i,** Using its two-anchor crawling locomotion, our six-DOF soft robot can negotiate across a challenging barrier with strict shape constraints. The dashed red arrows represent the trajectory of the soft robot. Scale bars, 1 mm.

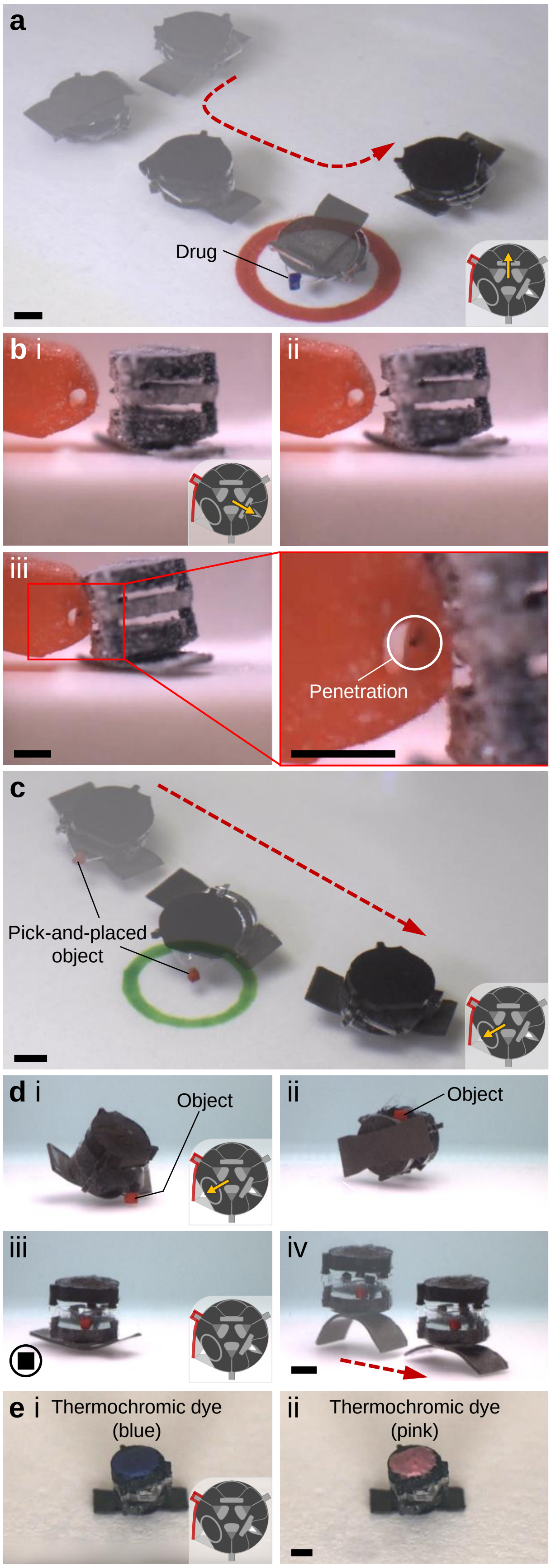

**Fig. 3 | Surgical functions of the soft robot when it is reprogrammed to its respective function modes.** To highlight the soft robot's function mode, we add the magnetic moments of its reprogrammable module and remote heating component at the bottom right corner of selected sub-figures. These illustrations are similar to those in Fig. 1. **a,** In its drug-dispensing mode, the soft robot can roll towards the targeted location (marked by a red circle) and dispense a simulated solid drug. The simulated drug is represented by a blue PVC object, which has a dimension of 0.24 mm × 0.4 mm × 0.4 mm. **b,** After reprogramming the soft robot to its cutting mode, it can penetrate a 10 wt% gelatin structure with the extended cutting tool. **c,** When the soft robot is in its gripping/storage mode, it can use a spin-walking locomotion to pick-and-place a red PVC object (0.4 mm × 0.4 mm × 0.4 mm) to the targeted location marked by the green circle. **d,** In its gripping/storage mode, the soft robot can also pick up a red PVC object (0.4 mm × 0.4 mm × 0.4 mm) and store this object in its main body. By reprogramming the soft robot to its locomotion mode, this actuator can subsequently safely transport the object via the two-anchor crawling gait. **e,** The soft robot can remotely heat up a layer of thermochromic dye that is placed on its remote heating component. This surgical function is performed with the soft robot reprogrammed to its locomotion mode. The thermochromic dye is blue when its temperature is below 40°C, and it will change to pink when its temperature is 40°C or above. Since the thermochromic dye changes from blue to pink, this implies that it has been heated up to 40°C or above by our soft robot. During the heating process, the soft robot remains stationary. The dashed red arrows represent the trajectory of our soft robot. Scale bars, 1 mm.

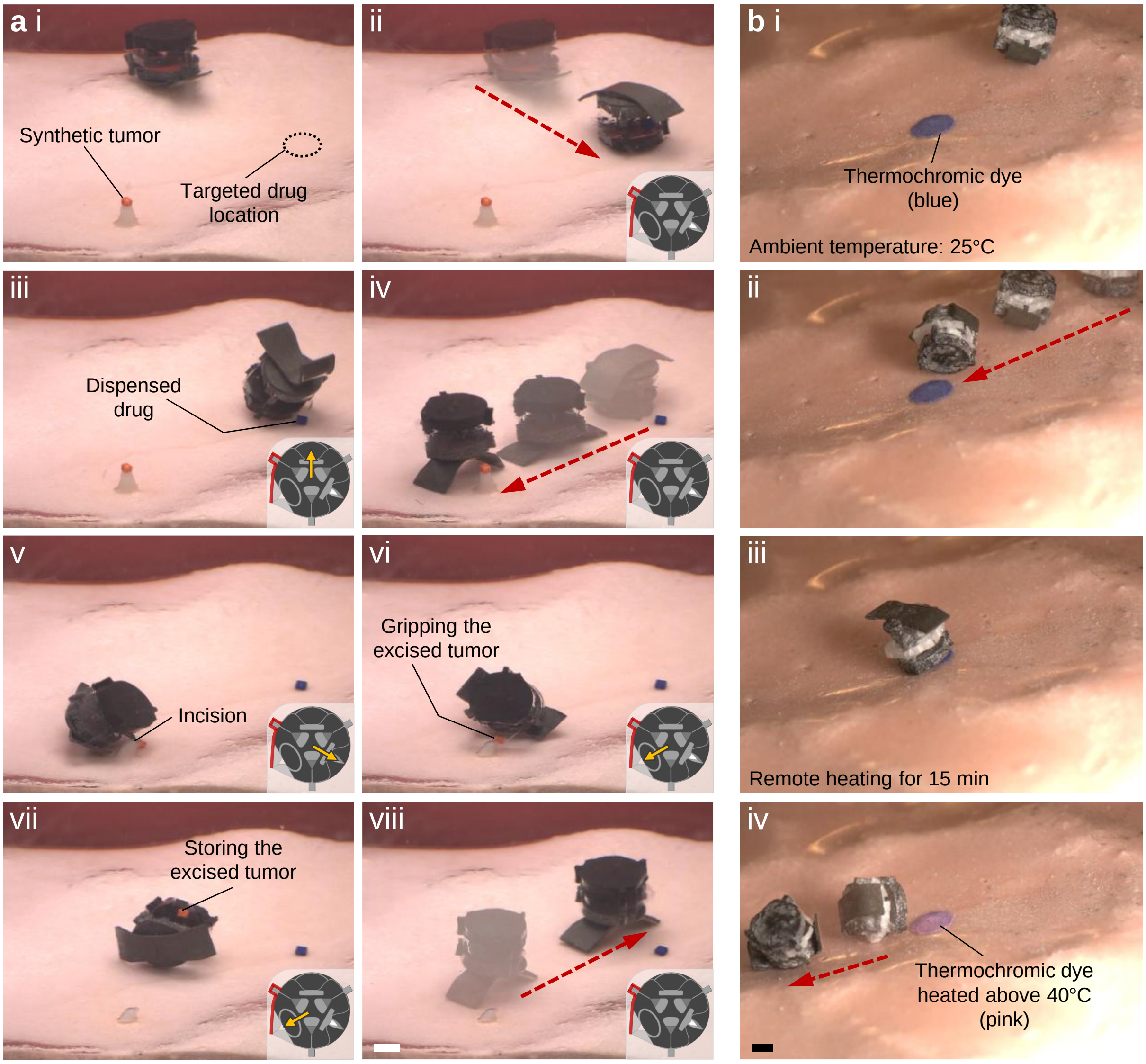

**Fig. 4 | In vitro experiments on biological phantoms. a,** Demonstration of the soft robot's locomotion and four reprogrammable surgical functionalities. The magnetic moments of the actuator's reprogrammable module and remote heating component are illustrated at the bottom right corner of selected sub-figures to indicate its function mode. These illustrations are similar to those in Fig. 1. (i)-(ii) In its locomotion mode, the soft robot can roll along its length to reach the targeted location for drug-dispensing. (iii) By reprogramming the soft robot to its drug-dispensing mode, it can be actuated to dispense a simulated drug, which is represented by a blue object made of PVC (0.24 mm × 0.4 mm × 0.4 mm). (iv) Subsequently, the soft robot is reprogrammed back to its locomotion mode so that it can two-anchor crawl towards the synthetic tumor (represented by a red PVC object with a dimension of 0.4 mm × 0.4 mm × 0.4 mm). (v) At close proximity to the synthetic tumor, the soft robot is then reprogrammed to its cutting mode. Upon actuation, the cutting tool in the soft robot can be extended out before making an incision to remove the tumor. For proof-of-concept purposes, the synthetic tumor is connected to the substrate via soft butter so that incision can be more easily performed in this experiment. (vi)-(vii) Thereafter, the soft robot is reprogrammed to the gripping/storage mode to pick and store the excised tumor in its body. (viii) Finally, the soft robot is reprogrammed back to its locomotion mode before transporting the stored excised tumor back to the entrance via the two-anchor crawling gait. **b,** Demonstration of the actuator's locomotion and remote heating function. The actuator is reprogrammed to its locomotion mode for this entire experiment. (i)-(ii) The soft robot rolls along its width to the targeted heating location (marked out by the thermochromic dye). The thermochromic dye is blue when its temperature is below 40°C and it will change to pink when the temperature rises to 40°C or above. (iii) After the remote heating component of the soft robot has established contact with the dye, the soft robot is actuated by an alternating magnetic field (amplitude: 9.34 mT, frequency: 75.4 kHz) for 15 minutes. This causes the soft robot's remote heating component to heat up the thermochromic dye, causing the dye to eventually change from blue to pink. During the heating process, the soft robot remains stationary. (iv) Finally, the soft robot rolls away from the targeted location. The dashed red arrows represent the trajectory of our soft robot. Scale bars, 1 mm.

Supplementary Information

# Miniature soft robot with magnetically reprogrammable surgical functions

Chelsea Shan Xian Ng[1†], Yu Xuan Yeoh[1†], Nicholas Yong Wei Foo[1†], Keerthana Radhakrishnan[2], Guo Zhan Lum[1,2*]

[1]School of Mechanical and Aerospace Engineering, Nanyang Technological University, 50 Nanyang Avenue, Singapore 639798, Singapore

[2]Rehabilitation Research Institute of Singapore, Nanyang Technological University, 11 Mandalay Road, Singapore 308232, Singapore

[†]Equally contributing authors

*Correspondence to: gzlum@ntu.edu.sg

**This document includes:**

**Nomenclature**

| | |
|---|---|
| | **SI Section S1 - Materials and methods** |
| $\vec{m}_{\mathrm{Loco}}$ | Net magnetic moment of the soft robot (locomotion mode) |
| $\vec{m}_{\mathrm{Heat}}$ | Magnetic moment of the remote heating component |
| $\vec{m}_{\mathrm{Rprog}}$ | Magnetic moment of the reprogrammable module |
| $M$ | Magnetization magnitude |
| $M_{\mathrm{Sample}}$ | Magnetization magnitude of the sample |
| $\vec{B}$ | Magnetic field |
| $\vec{B}_{\mathrm{Test}}$ | Applied magnetic fields for the customized actuation tests |
| $\vec{B}_{\mathrm{Demag}}$ | Demagnetizing magnetic field |
| $\vec{B}_{\mathrm{Func}}$ | The component of $\vec{B}$ that is along $\vec{m}_{\mathrm{Rprog}}$ and $\vec{m}_{\mathrm{Heat}}$ |
| $B_{\mathrm{Demag}}$ | Maximum amplitude of $\vec{B}_{\mathrm{Demag}}$ |
| $K_{\mathrm{Demag}}$ | Decaying magnitude of $\vec{B}_{\mathrm{Demag}}$ (per period) |
| $H_{\mathrm{ci}}$ | Intrinsic coercivity |
| $V$ | Volume of magnetic component |
| $V_{\mathrm{Sample}}$ | Volume of the sample |
| $E$ | Young's modulus |
| $I$ | Second moment of area |
| $\gamma$ | Angular deflection of the beam |
| $\gamma_{\mathrm{Beamtip}}$ | Value of $\gamma$ at the beam's free end |
| $s$ | Arc length along the beam's length |
| $l$ | Length of beam |
| $l_{\mathrm{Beam}}$ | Length of the test beam |
| $\varphi$ | Angle for describing the soft robot's operating mode |
| $\varphi_{\mathrm{Disp}}$ | When $\varphi$ is 90° |
| $\varphi_{\mathrm{Cut}}$ | When $\varphi$ is 330° |
| $\varphi_{\mathrm{Grip}}$ | When $\varphi$ is 210° |
| $f$ | Frequency of $\vec{B}$ |
| $f_{\mathrm{Demag}}$ | Frequency of $\vec{B}_{\mathrm{Demag}}$ |
| | **SI Section S2 - Theory** |
| $\vec{B}_{\mathrm{Grad}}$ | Independent spatial gradients of $\vec{B}$ |
| $\vec{B}_{\mathrm{Tent}}$ | The component of the applied $\vec{B}$ along $\vec{m}_{\mathrm{Tent(soft)}}$ |
| $\vec{M}$ | Magnetization profile of the soft robot |
| $\vec{M}_{\mathrm{Tent(soft)}}$ | Magnetization profile of the soft part of the tentacles |
| $\vec{M}_{\mathrm{Tent(rigid)}}$ | Magnetization profile of the sixth-DOF enhancement module |
| $\vec{M}_{\mathrm{Main}}$ | Magnetization profile of the main body |
| $\vec{M}_{\mathrm{Rprog}}$ | Magnetization profile of the reprogrammable module |
| $\vec{M}_{\mathrm{Heat}}$ | Magnetization profile of the remote heating component |
| $\vec{M}_{\mathrm{Loco}}$ | Magnetization profile of the soft robot in the locomotion mode |
| $M_{\mathrm{Tent(soft)}}$ | Magnetization magnitude of the soft tentacles |
| $M_{\mathrm{Tent(rigid)}}$ | Magnetization magnitude of the sixth-DOF enhancement module |
| $M_{\mathrm{Rprog}}$ | Magnetization magnitude of the reprogrammable module |
| $M_{\mathrm{Heat}}$ | Magnetization magnitude of the remote heating component |
| $M_{\mathrm{Inner}}$ | Magnetization magnitude of an inner magnet |
| $\vec{m}_{\mathrm{Main}}$ | Magnetic moment of the soft robot's main body |
| $\vec{m}_{\mathrm{Cut}}$ | Net magnetic moment of the soft robot (cutting mode) |
| $\vec{m}_{\mathrm{Disp}}$ | Net magnetic moment of the soft robot (drug-dispensing mode) |
| $\vec{m}_{\mathrm{Grip}}$ | Net magnetic moment of the soft robot (gripping/storage mode) |
| $u$ | Heaviside step function |
| $\delta$ | Dirac delta function |
| $\mathbf{R}_x$ | Standard $X$-axis rotation matrix |
| $\mathbf{R}_y$ | Standard $Y$-axis rotation matrix |
| $\mathbf{R}_z$ | Standard $Z$-axis rotation matrix |
| $s_{\mathrm{Tent}}$ | Arc length along the soft tentacle's length |
| $s_{\mathrm{Inner}}$ | Arc length along the inner beam's length |
| $\gamma_{\mathrm{Tent}}$ | Angular deflection of the soft tentacles |
| $\gamma_{\mathrm{Inner}}$ | Angular deflection of the inner beam |
| $\gamma_{\mathrm{Intip}}$ | The value of $\gamma_{\mathrm{Inner}}$ at the inner beam's free end |
| $l_{\mathrm{Inner}}$ | Length of the inner beam |
| $M_{\mathrm{b}}$ | Bending moment |
| $A$ | Cross-sectional area |
| $A_{\mathrm{Tent}}$ | Cross-sectional area of the soft tentacles |
| $\tau$ | Distributed magnetic torque (per volume) |
| $T$ | Magnitude of a magnetic torque |
| $\vec{T}$ | Magnetic torque |
| $\vec{F}$ | Magnetic force |
| $\vec{r}$ | Displacement vector from the soft robot's center of mass to a point of interest |
| $\mathbf{D}_{\mathrm{Loco}}$ | Design matrix (locomotion mode) |
| $\mathbf{D}_{\mathrm{Disp}}$ | Design matrix (drug-dispensing mode) |
| $\mathbf{D}_{\mathrm{Cut}}$ | Design matrix (cutting mode) |
| $\mathbf{D}_{\mathrm{Grip}}$ | Design matrix (gripping/storage mode) |
| $\mathbf{C}_{\mathrm{Loco}}$ | Control matrix (locomotion mode) |
| $\mathbf{C}_{\mathrm{Disp}}$ | Control matrix (drug-dispensing mode) |
| $\mathbf{C}_{\mathrm{Cut}}$ | Control matrix (cutting mode) |
| $\mathbf{C}_{\mathrm{Grip}}$ | Control matrix (gripping/storage mode) |
| $d_i$ | $i^{\mathrm{th}}$ element of the design matrix |
| $\theta$ | Desired angular displacement about the $Z_{\{I\}}$-axis |
| $k_1, k_2$ | Scale factors |
| $\alpha, \beta$ | Angles required to define the soft robot's intermediate coordinate frame |
| **SI Section S3 – Experimental characterization for the deformation of beams** | |
| $\xi_{\mathrm{Disp},1}$ | Deviation angle of $\vec{m}_{\mathrm{Disp}}$ (drug-dispensing mode with inverted 'U'-shaped tentacles) |
| $\xi_{\mathrm{Cut},1}$ | Deviation angle of $\vec{m}_{\mathrm{Cut}}$ (cutting mode with inverted 'U'-shaped tentacles) |
| $\xi_{\mathrm{Grip},1}$ | Deviation angle of $\vec{m}_{\mathrm{Grip}}$ (gripping/storage mode with inverted 'U'-shaped tentacles) |
| $\xi_{\mathrm{Disp},2}$ | Deviation angle of $\vec{m}_{\mathrm{Disp}}$ (drug-dispensing mode with upright 'U'-shaped tentacles) |
| $\xi_{\mathrm{Cut},2}$ | Deviation angle of $\vec{m}_{\mathrm{Cut}}$ (cutting mode with upright 'U'-shaped tentacles) |
| $\xi_{\mathrm{Grip},2}$ | Deviation angle of $\vec{m}_{\mathrm{Grip}}$ (gripping/storage mode with upright 'U'-shaped tentacles) |
| $a_{\mathrm{Disp}}$ | Shortest distance between the doors of the drug chamber (during actuation) |
| $a_{\mathrm{Cut}}$ | Extension of the cutting tool (during actuation) |
| $a_{\mathrm{Grip}}$ | Shortest distance between the grippers (during actuation) |
| **SI Section S4 – Locomotion** | |
| $f_{\vec{B}}$ | Frequency of the rotating $\vec{B}$ |
| $f_{\vec{B}_{\mathrm{Grad}}}$ | Frequency of varying $\theta$ |
| $v_{\mathrm{Roll}}$ | Rolling speed |
| $v_{\mathrm{Crawl}}$ | Two-anchor crawling speed |
| $f_{\mathrm{Roll}}$ | Frequency of the rotating $\vec{B}$ (for rolling) |
| $f_{\mathrm{Crawl}}$ | Frequency of $\vec{B}$ (for two-anchor crawling) |
| **SI Section S7 – Additional Discussion** | |
| $B_{\mathrm{Rotate}}$ | Magnitude of $\vec{B}$ (for rotation) |
| $B_{\mathrm{Act}}$ | Magnitude of the step output $\vec{B}$ by the main electromagnetic coil system |
| $B_{\mathrm{Mag}}$ | Magnitude of the step magnetizing field |
| $B_{\mathrm{Heat}}$ | Amplitude of the alternating $\vec{B}$ (for remote heating) |
| $f_{\mathrm{Heat}}$ | Frequency of the alternating $\vec{B}$ (for remote heating) |
| $\frac{\mathrm{d}\lvert\vec{B}\rvert}{\mathrm{d}t}$ | Temporal rate of change of the applied magnetic field |
| $\frac{\mathrm{d}\lvert\vec{B}\rvert}{\mathrm{d}t}_{\mathrm{Highest}}$ | Highest temporal rate of change of the applied magnetic field |
| $\frac{\mathrm{d}\lvert\vec{B}\rvert}{\mathrm{d}t}_{\mathrm{Max}}$ | Maximum allowable temporal rate of change of the applied magnetic field |
| $L_{\mathrm{EC}}$ | Inductance of the main electromagnetic coil system |
| $L_{\mathrm{RC}}$ | Inductance of the reprogramming coil |
| $R_{\mathrm{EC}}$ | Resistance of the main electromagnetic coil system |
| $R_{\mathrm{RC}}$ | Resistance of the reprogramming coil |
| $K$ | Gradient of the ramp magnetizing field |
| $\eta$ | Duration of a monotonic increasing or decreasing gradient |
| $\vec{H}$ | Magnetic field intensity |
| $\lvert\vec{H}\rvert f_{\mathrm{Highest}}$ | The highest calculated $\lvert\vec{H}\rvert f$ value of our applied magnetic fields |
| $l_{\mathrm{c}}$ | Characteristic length of the soft robot |
| $\lambda_{\mathrm{Body}}$ | Scale factor for the body of the soft robot |
| $\lambda_{\mathrm{Tent}}$ | Scale factor for the soft tentacles of the soft robot |
| $b$ | Width of beam |
| $h$ | Thickness of beam |
| $\dot{Q}_{\mathrm{Gen}}$ | Heat generation rate |
| $\dot{Q}_{\mathrm{Trf}}$ | Heat transfer rate |

**SI Section S1. Materials and methods**

In this section, we will describe the material properties and fabrication process of our soft robot in Section S1A and discuss about feasible methods to make it fully biocompatible in Section S1B. Subsequently, we will describe the required magnetic fields to reprogram and actuate our soft robot in its function modes (Section S1C). Our experimental setup is also reported in Section S1D.

A. Material properties and fabrication process

As shown in Fig. 1**a**, our proposed soft robot can be decomposed into its remote heating component, main body, reprogrammable module, and tentacles. Its tentacles can be further broken down into the sixth-degree-of-freedom (DOF) enhancement module, spacer, and soft parts of the tentacles (Fig. S1**a**). The function of the sixth-DOF enhancement module is to increase the producible torque about the soft robot's net magnetic moment ($\vec{m}_{\mathrm{Loco}}$) when the robot is in its locomotion mode (see Fig. S7**b** for an illustration of $\vec{m}_{\mathrm{Loco}}$). Because our soft robot has a polymer-based body, it can endure much higher mechanical stress than existing magnetic miniature robots that have liquid- or slime-based bodies (*1*–*6*). As a result, our soft robot is much less susceptible than their counterparts with liquid or slime-based bodies in breaking its body due to the adhesion between itself and the environment's substrate (*1*–*6*). Indeed, we report that our soft robot does not leave body residuals in the environment for all of our experiments.

In general, all of the robot's components are created by a combination of soft materials so that our soft robot can possess a diverse range of magnetic response. Specifically, the sixth-DOF enhancement module and the magnetized components in the main body are created by a polydimethylsiloxane (PDMS) polymer matrix (Dow Corning) embedded with NdFeB microparticles (average size: 5 μm), while the tentacles' soft components are created by embedding the Ecoflex 00-10 polymer matrix with NdFeB microparticles (average size: 5 μm). As the embedded NdFeB microparticles of these components have a high intrinsic magnetic coercivity (93.3-614 mT) after they are magnetized by a uniform field of 1.1 T, they can only be demagnetized with field strengths greater than 93.3 mT (Fig. S2). Thus, their magnetization profiles will not be altered when the soft robot is actuated or reprogrammed by $\vec{B}$ with magnitudes of 10-65 mT. By having high magnetic remanence (per volume) of 37.5-108 kA/m (Fig. S2), the aforementioned components can also generate relatively large magnetic torques and forces during actuation. The reprogrammable module is made by embedding PDMS with AlNiCo microparticles (average size: 5 μm) while the remote heating component is made by embedding PDMS with $Fe_3O_4$ microparticles (average size: 5 μm). These two components can be easily magnetized by a $\vec{B}$ of 60 mT and demagnetized by an alternating $\vec{B}$ with a linearly decaying amplitude (maximum amplitude: 65 mT). In theory, we can achieve magnetizing and demagnetizing fields that are harmless to the human body, allowing our reprogramming strategies to be compatible for biomedical applications (SI Section S7A). Furthermore, both the reprogrammable module and remote heating component have functional magnetization magnitudes (per volume) of 7.18 kA/m and 6.52 kA/m, respectively, after they are magnetized by a spatially uniform field of 60 mT. Therefore, these components will contribute an effective magnetic moment of $3.56 \times 10^{-5}$ A·m$^2$ for the proposed robot when it is in the function modes. While the remaining components of our soft robot are made of pristine PDMS which are non-responsive to magnetic fields, they are essential for maintaining the robot's structural integrity. The Young's moduli of our soft materials are evaluated via standard compression tests (SHIMADZU AG-X plus, 10 kN). For these experiments, we have created corresponding cylindrical samples for each type of polymer composite and the stress-strain relationship of every sample is measured five times. In general, the soft components of the tentacles are relatively more compliant than the remaining components of our robot and thus they will be able to produce large deformations, which are essential for generating the soft robot's gaits. The polymer matrix to particles mass ratios of all our robot's components and their respective Young's moduli ($E$) are summarized in Table S1.

The magnetic properties of our soft robot's materials are evaluated by a vibrating-sample magnetometer (VSM, Lake Shore Cryotronics 8600 Series) or via customized actuation tests. The VSM is used for the materials that have hard magnetic properties, i.e., those that are used for our soft tentacles and the magnetized components in the main body. For these experiments, corresponding samples have been created and their magnetic hysteresis loop are evaluated by the VSM based on their magnetizing field of 1.1 T (Fig. S2 and Table S1). In these hysteresis loops, the magnetization magnitude ($M$) and intrinsic coercivity ($H_{\mathrm{ci}}$) of these soft materials are measured accordingly and these data are summarized in Table S1.

The magnetic properties of our reprogrammable module and the remote heating component are measured via customized actuation tests. For these experiments, we use a fixed-free beam that is created with pristine PDMS and its free end can be adhesively attached with different magnetic samples (Fig. S3**a**(i)). When the beam is undeformed, the magnetic moment of the samples is parallel to the beam. After applying a $\vec{B}$ that is perpendicular to the beam (denoted as $\vec{B}_{\mathrm{Test}}$ in Fig. S3**a**(ii)), the magnetic moment of the samples will tend to align with the applied field, generating a magnetic torque that will induce a bending moment across the beam to deform it. In theory, the bending moment across a fixed-free beam has a constant magnitude and it is equal to the magnetic torque generated at the beam's free end (*7*). Two fixed-free beams of equal length, $l_{\mathrm{Beam}}$ (9.3 mm), are created for our customized actuation tests. The first beam is used to evaluate the magnetization magnitudes of the materials used for the reprogrammable module and remote heating component after they have been magnetized by a uniform $|\vec{B}|$ of 60 mT, while the second beam is used to evaluate their magnetization magnitude after they have been demagnetized. Since we expect the magnetization magnitudes of the demagnetized components to be significantly lower, the second beam is made to be more compliant compared to the first beam so that the deflections caused by the smaller magnetization magnitudes can be observed. We make the second beam more compliant than the first beam by reducing its thickness. To evaluate the stiffness of our fixed-free beams, we first conduct a series of tests based on a sample that is constructed by the material, which we use for the magnetized components in our soft robot' main body (denoted as inner magnets). Prior to the experiment, this sample has been magnetized by 1.1 T and our VSM results indicate that such materials will possess a magnetization magnitude (per volume) of 108 kA/m (Fig. S2**a**). Using the Euler-Bernoulli equation (*8*), the beams' deformation in Fig. S3**a** can be described as:

$$\left|\vec{B}_{\mathrm{Test}}\right| M_{\mathrm{Sample}} V_{\mathrm{Sample}} \sin\left(90^{\circ} - \gamma_{\mathrm{Beamtip}}\right) = EI\frac{d\gamma}{ds}, \tag{S1.1}$$

where $M_{\mathrm{Sample}}$ and $V_{\mathrm{Sample}}$ correspond to the magnetization magnitude and volume of the sample, while $E$ and $I$ represent the Young's modulus and the second moment of area of a corresponding beam, respectively. The remaining variables in Eq. (S1.1), $\gamma$ and $\gamma_{\mathrm{Beamtip}}$, represent the deflection of the corresponding beam at the arc length ($s$) and the free end, respectively. The left side of Eq. (S1.1) represents the magnetic torque and thus the bending moment of the beam at $s$, while the right side of Eq. (S1.1) represents the deformation of the beam. By integrating Eq. (S1.1) with respect to $s$ and subsequently applying the fixed-free boundary conditions, the obtained equation can be rearranged to establish the relationship between the measured $\gamma_{\mathrm{Beamtip}}$ and applied $\left|\vec{B}_{\mathrm{Test}}\right|$:

$$\gamma_{\mathrm{Beamtip}} \sec\gamma_{\mathrm{Beamtip}} = M_{\mathrm{Sample}} \frac{V_{\mathrm{Sample}} l_{\mathrm{Beam}}}{EI} \left|\vec{B}_{\mathrm{Test}}\right|. \tag{S1.2}$$

Based on Eq. (S1.2), the stiffness of the fixed-free beams is first characterized by varying $\left|\vec{B}_{\mathrm{Test}}\right|$ and measuring their corresponding $\gamma_{\mathrm{Beamtip}}$. For the first beam, $\left|\vec{B}_{\mathrm{Test}}\right|$ is specified to be 5 mT, 10 mT, and 15 mT (Fig. S3**b**(i)), while the second beam's $\left|\vec{B}_{\mathrm{Test}}\right|$ is specified to be 2 mT, 4 mT, and 6 mT (Fig. S3**c**(i)). We make five measurements for each data point in Fig. S3**b**(i), **c**(i). Based on these data, we use Eq. (S1.2) to plot $\gamma_{\mathrm{Beamtip}} \sec\gamma_{\mathrm{Beamtip}}$ against $\left|\vec{B}_{\mathrm{Test}}\right|$(Fig. S3**b**(i), **c**(i)). Since $V_{\mathrm{Sample}}$ is 7.07 mm$^3$, we use the gradient of the slopes in Fig. S3**b**(i), **c**(i) to determine that the flexural rigidity ($EI$) of our first and second fixed-free beams are $1.21 \times 10^{-7}$ (kg·m$^3$)/s$^2$ and $2.01 \times 10^{-8}$ (kg·m$^3$)/s$^2$, respectively.

After the stiffness of our fixed-free beams have been characterized, we proceed to adhesively attach the free end of these beams with samples that are created by the materials of our reprogrammable module and remote heating component. The magnetization magnitudes of these samples are then characterized when they are magnetized by a uniform $\left|\vec{B}\right|$ of 60 mT and demagnetized. For all of these tests, we vary $\left|\vec{B}_{\mathrm{Test}}\right|$ and measured the deformed beam's $\gamma_{\mathrm{Beamtip}}$ (Fig. S3**b-c**). For each data point in Fig. S3**b-c**, five trials are measured. Using these data, we can plot $\gamma_{\mathrm{Beamtip}} \sec\gamma_{\mathrm{Beamtip}}$ against $\left|\vec{B}_{\mathrm{Test}}\right|$ based on Eq. (S1.2) (Fig. S3**b**(ii)-(iii), **c**(ii)-(iii)). Using the known values of $V_{\mathrm{Sample}}$, $l_{\mathrm{Beam}}$, $EI$ and the gradient of the slopes in Fig. S3**b**(ii)-(iii), **c**(ii)-(iii), we can therefore determine the samples' $M_{\mathrm{Sample}}$ via Eq. (S1.2). In general, when our reprogrammable module and remote heating component are magnetized, they have magnetization magnitude (per volume) of 7.18 kA/m and 6.52 kA/m, respectively. However, when they are demagnetized by an alternating magnetic field with a linearly decaying amplitude (SI Section S1C), the reprogrammable module and remote heating component have a magnetization magnitude of 1.19 kA/m and 0.766 kA/m, respectively. Because their respective magnetization magnitude is 6.03- and 8.51-fold smaller when they are demagnetized, their effects become negligible when our soft robot assumes its locomotion mode.

When the reprogramming module and remote heating component are subjected to an alternating magnetic field of high frequency (amplitude: 9.34 mT, frequency: 75.4 kHz), they will generate heat via magnetic hysteresis losses. To evaluate their heat generation capabilities, respective cylindrical samples are created, where each sample has a diameter of 2.5 mm and a thickness of 1 mm. The samples are coated with a thermochromic dye that will turn pink when its temperature reaches 40°C and above. Otherwise, the thermochromic dye will remain blue. By subjecting the samples with the specified alternating magnetic field, we measure the required time for each sample to raise its temperature from the ambient temperature (26.3°C) to 40°C. Our experimental results indicate that the sample which is made by the material of the remote heating component requires 15 s to do so, while the sample that is made by the reprogrammable module requires 60 s. This implies that the heating power of the remote heating component is fourfold of the reprogrammable module. Because the reprogrammable module has relatively lower heat generation capabilities, it is used as an insulator base to minimize undesired heating to the encased drugs and excised tissue in the soft robot's main body. The remote heating component, however, is placed at the top of the soft robot so that it can better enable potential hyperthermia treatments with its relatively high heat generation capabilities.

To fabricate the proposed soft robot, all of its components are first molded separately via techniques that are similar to those reported in Xu et al. (*7*, *9*). During these molding processes, the components are cured at 80°C in an oven. Except for the remote heating component and the soft components of the tentacles that required 12 h and 1 h to cure, respectively, the remaining components are cured within 0.5 h. After they are cured, those that have embedded NdFeB microparticles are magnetized individually with 1.1 T (Fig. S1). Subsequently, all the components of our soft robot are assembled manually and bonded adhesively together. During the assembly process, a cutting tool is adhesively attached to the robot (Fig. S1**b**). This cutting tool is a tungsten carbide rod that has a diameter of 0.25 mm and length of 0.6 mm (from the base to the tip). Its cutting edge has an angle of 45° and the cutting surface area is estimated to be $1.47 \times 10^{-4}$ $\mathrm{mm}^2$. A solid drug can also be inserted into the robot's main body post-assembly.

B. Biocompatibility

The proposed soft robot is generally made of biocompatible materials, i.e., the PDMS and Ecoflex 00-10 polymer matrices (*10*, *11*), and the $Fe_3O_4$ and AlNiCo microparticles. While the components with embedded NdFeB microparticles are partially biocompatible (*12*) (i.e., the sixth-DOF enhancement module, soft components of the tentacles and the magnetized components in the main body), full biocompatibility can be recovered by safely sealing these microparticles within the robot's body (*13*). For instance, we can potentially coat an additional thin layer of PDMS or Ecoflex 00-10 polymer matrix to these components so that the embedded NdFeB particles can be hermetically sealed within them. This proposed solution will theoretically not affect the soft robot's drug-dispensing, cutting, gripping and storage functions as the magnetized components in the main body are designed to be rigid with negligible deformation. Likewise, since the sixth-DOF enhancement module is also designed to be rigid, having an additional layer of pristine PDMS and Ecoflex 00-10 will not affect the robot's soft-bodied functionalities. Furthermore, the soft components of the tentacles should still be sufficiently compliant to execute all of the proposed locomotive gaits as long as the additional layer of pristine Ecoflex 00-10 is sufficiently thin. To reduce the surface energy of the soft robot, we can also chemically treat the additional layer of PDMS or Ecoflex with a thin film of polytetrafluoroethylene (PTFE) (*14*), which are fully biocompatible (*15*). As the first study, our aim is to report the functionalities of the proposed robot. Therefore, we have not implemented the proposed coating solution yet. In the future, we aim to use this solution on our soft robot before deploying it for in vivo experiments.

C. Magnetic fields for actuation

Although $\vec{B}$ and $\vec{B}_{\mathrm{Func}}$ are generally used for describing all the actuating magnetic fields in the main text, we will use other variables in the SI to describe specific fields that are applied to the proposed soft robot so that our explanations can become clearer. In general, our soft robot can switch between its locomotion and function modes via magnetizing or demagnetizing its reprogrammable module and remote heating component (Fig. S4 and SI Video S1). When these components are demagnetized, the soft robot will be in its locomotion mode. Conversely, the soft robot can be reprogrammed to a function mode by magnetizing it with a spatially uniform $\vec{B}$ of $60$ mT for 5 ms (Fig. S4**a**-**b**(i)) along one of the three principal directions specified in Fig. 1**b**(ii). Using the angle $\varphi$ in Fig. 1**b**(ii) as a reference, if the magnetic moments of the reprogrammable module ($\vec{m}_{\mathrm{Rprog}}$) and remote heating component ($\vec{m}_{\mathrm{Heat}}$) are parallel to the direction in which $\varphi$ is $90°$, $330°$ or $210°$, the soft robot will be in its drug-dispensing, cutting or gripping/storage modes, respectively. To facilitate our discussion, these respective values of $\varphi$ are represented by $\varphi_{\mathrm{Disp}}$, $\varphi_{\mathrm{Cut}}$, and $\varphi_{\mathrm{Grip}}$. At all times, the magnetic moments of the reprogrammable module and remote heating component can also be demagnetized upon command via an alternating magnetic field with a linearly decaying magnitude ($\vec{B}_{\mathrm{Demag}}$) applied along them (Fig. S4**c**-**d**):

$$|\vec{B}_{\mathrm{Demag}}| = \left(B_{\mathrm{Demag}} - K_{\mathrm{Demag}} f_{\mathrm{Demag}} t\right) \cos\left(2\pi f_{\mathrm{Demag}} t\right), \tag{S1.3}$$

where $f_{\mathrm{Demag}}$, $B_{\mathrm{Demag}}$ and $K_{\mathrm{Demag}}$ represent the frequency (45 Hz), maximum amplitude (65 mT) and decaying magnitude (2 mT per period) of the demagnetizing field, respectively. Equation (S1.3) implies that the demagnetizing field will start off at its peak value of 65 mT and it will continuously decrease by 2 mT after each period until the amplitude of this alternating field reaches 0 mT (Fig. S4**d**).

When the proposed robot is in a function mode, applying a $\vec{B}_{\mathrm{Func}}$ of up to 34 mT along $\vec{m}_{\mathrm{Rprog}}$ and $\vec{m}_{\mathrm{Heat}}$ will generate a magnetic torque at the free ends of a corresponding pair of beams in the main body (Fig.s 1**c**, S5 and SI Video S1). The generated magnetic torque will in turn cause these beams to bend and activate a specific mechanical function (Fig.s 1**c**(ii), 1**d**(iii), (vi), (ix), S5**a**(ii), **b**(ii) and SI Video S1). For instance, when the soft robot is in its drug-dispensing mode, applying a $\vec{B}_{\mathrm{Func}}$ of 15 mT will deform a pair of soft beams, which will in turn push and open the doors of the drug chamber, creating an opening that can allow our soft robot to release the encased drug (Fig.s 1**d**(vi), S5**a**). Similarly, when the soft robot is in its cutting mode, applying a $\vec{B}_{\mathrm{Func}}$ of 15 mT will deform a pair of soft beams to push the cutting tool out of the robot's body (Fig. 1**c**, 1**d**(iii)). Applying a $\vec{B}_{\mathrm{Func}}$ of 15 mT will also push a pair of soft grippers out of the main body when the soft robot is in its gripping mode (Fig.s 1**d**(ix), S5**b**). The grippers can be used to pick up and eventually store an object in the robot's body (Fig. 3**d**). Due to the stored elastic energy within the deformed robot, the robot will always spring back to its undeformed configuration once the actuating magnetic field is removed. This is true regardless of whether the soft robot is in its locomotion or function modes.

D. Experimental setup

Unless specified otherwise, all of our experiments are conducted in a paraffin oil reservoir (viscosity: 11 $mm^2/s$), which is placed in the workspace of a customized electromagnetic coil system (Fig. S6**a**). Our electromagnetic coil system has a similar configuration to those reported by Xu et al. (*7*, *9*), and it has a workspace of 1.6 cm × 1.6 cm × 1.6 cm that can generate a homogenous field. Using this electromagnetic coil system, we can apply $\vec{B}$ and its magnetic spatial gradients of up to 34 mT and 0.4 T/m to our soft robot, respectively. These magnetic control signals are mainly used for actuating the proposed robot to produce its locomotion and mechanical functionalities. Within the workspace of the electromagnetic coil system, we can also place a reprogramming coil (single electromagnetic coil) that can output a $\vec{B}$ of up to 67 mT (Fig. S6**b**). The reprogramming coil is used for magnetizing or demagnetizing the reprogrammable module and remote heating component of our soft robot, which is placed at the center of the reprogramming coil (Fig. S6**b**). However, as the reprogramming coil can only generate magnetic fields along one specified direction, we can only reprogram our soft robot along this direction during our experiments. To perform the remote heating experiments, we must replace the reprogramming coil with a heating coil that can produce alternating magnetic fields of high frequency (amplitude: 9.34 mT, frequency: 75.4 Hz). The setup of the heating coil is shown in Fig. S6**c**. The reprogramming and heating coils cannot be used concurrently in our experiments due to the space constraints of our customized electromagnetic coil system. However, this limitation can be easily overcome by redesigning the configuration of our electromagnetic coil system and we will explore this approach in the future.

**SI Section S2. Theory**

In this section, we first analyze the deformation mechanism of our soft robot in Section S2A, and then discuss how the robot can be actuated with six degrees-of-freedom (DOFs) motions in Section S2B. For these analyses, we assume that $\vec{B}$ and its spatial gradients are uniform across the soft robot because it is very challenging to vary them spatially at small scales (*16–18*). This is especially true when the actuating magnetic signals are generated at a distance (*8*, *12*, *17*). Furthermore, the spatial gradients of $\vec{B}$ obey the Gauss's law and Ampere's law accordingly (*7*, *16–18*). The Gauss's law states that the divergence of $\vec{B}$ is equal to zero for all the coordinate frames:

$$\frac{\partial B_x}{\partial x} + \frac{\partial B_y}{\partial y} + \frac{\partial B_z}{\partial z} = 0, \tag{S2.1}$$

where $x$, $y$, and $z$ represent the coordinates of their respective Cartesian axes while $B_x$, $B_y$, and $B_z$ are the components of $\vec{B}$ along these Cartesian axes. Assuming that there are no electrical currents flowing in the workspace of our soft robot, the Ampere's law dictates that the following relationship must be enforced for all the coordinate frames:

$$\frac{\partial B_z}{\partial y} = \frac{\partial B_y}{\partial z}, \qquad \frac{\partial B_x}{\partial z} = \frac{\partial B_z}{\partial x}, \qquad \frac{\partial B_y}{\partial x} = \frac{\partial B_x}{\partial y}. \tag{S2.2}$$

Equations (S2.1-2) therefore dictate that there are only five independent spatial gradients of $\vec{B}$. Although there exist multiple ways to select and represent the independent spatial gradients of $\vec{B}$, (*16*), here we use the vector, $\vec{B}_{\text{Grad}}$, to represent these actuating signals:

$$\vec{B}_{\text{Grad}} = \left[\frac{\partial B_z}{\partial x} \; \frac{\partial B_z}{\partial y} \; \frac{\partial B_z}{\partial z} \; \frac{\partial B_y}{\partial y} \; \frac{\partial B_x}{\partial y}\right]^T. \tag{S2.3}$$

Establishing the key assumptions and independent spatial gradients of $\vec{B}$ will be beneficial for our subsequent analyses in this section.

A. Deformation analysis

To analyze the deformation characteristics of our soft robot, we first state the robot's magnetization profile, $\vec{M}_{\{M\}}$, according to its material coordinate frame (Fig. S1**a**):

$$\vec{M}_{\{M\}} = \vec{M}_{\text{Tent(soft)},\{M\}} + \vec{M}_{\text{Tent(rigid)},\{M\}} + \vec{M}_{\text{Main},\{M\}} + \vec{M}_{\text{Rprog},\{M\}} + \vec{M}_{\text{Heat},\{M\}}, \quad \text{(S2.4)}$$

where $\vec{M}_{\text{Tent(soft)},\{M\}}$ , $\vec{M}_{\text{Tent(rigid)},\{M\}}$ , $\vec{M}_{\text{Main},\{M\}}$ , $\vec{M}_{\text{Rprog},\{M\}}$ and $\vec{M}_{\text{Heat},\{M\}}$ represent the magnetization profile of the soft tentacles, sixth-DOF enhancement module, all the magnetized components in the main body, the reprogrammable module and the remote heating component, respectively. We denote vectors and axes that are expressed in the material coordinate frame with a subscript $\{M\}$ (Fig. S1**a**). In general, the magnetization profile of the soft robot can be approximated as a collection of magnetic moments embedded in it. The magnetization profile, $\vec{M}_{\text{Main},\{M\}}$, is made of three identical magnets that are placed at the free end of respective beams in the main body. These magnets are denoted as inner magnets, where each inner magnet's magnetization magnitude (per volume) is represented by $M_{\text{Inner},\{M\}}$.

The magnetization profile of $\vec{M}_{\text{Tent(soft)},\{M\}}$, $\vec{M}_{\text{Tent(rigid)},\{M\}}$, $\vec{M}_{\text{Main},\{M\}}$, $\vec{M}_{\text{Rprog},\{M\}}$ and $\vec{M}_{\text{Heat},\{M\}}$ can be expressed mathematically (Fig. S1**a**):

$$\begin{aligned}\vec{M}_{\text{Tent(soft)},\{M\}} = {}& M_{\text{Tent(soft)}}\left[u\left(Z_{\{M\}} - (Z_{\text{Tent(soft)},\{M\}} + t_{\text{Tent(soft)},\{M\}})\right) - u\left(Z_{\{M\}} - Z_{\text{Tent(soft)},\{M\}}\right)\right] \\ & \left[u\left(X_{\{M\}} + \frac{b_{\text{Tent}}}{2}\right) - u\left(X_{\{M\}} - \frac{b_{\text{Tent}}}{2}\right)\right] \\ & \left\{\left[u\left(Y_{\{M\}} + \frac{l_{\text{Tent}}}{2}\right) - u\left(Y_{\{M\}}\right)\right]\begin{bmatrix}0\\1\\0\end{bmatrix} + \left[u\left(Y_{\{M\}}\right) - u\left(Y_{\{M\}} - \frac{l_{\text{Tent}}}{2}\right)\right]\begin{bmatrix}0\\-1\\0\end{bmatrix}\right\},\end{aligned} \quad \text{(S2.5A)}$$

$$\begin{aligned}\vec{M}_{\text{Tent(rigid)},\{M\}} = {}& M_{\text{Tent(rigid)}}\frac{V_{\text{Six}}}{2} \\ & \left\{\sum_{i=1}^{2}\delta\left(Z_{\{M\}} - Z_{\text{Tent(rigid)},\{M\}}\right)\delta\left(Y_{\{M\}} - Y_{\text{Tent(rigid)},i,\{M\}}\right)\ \delta\left(X_{\{M\}}\right)\begin{bmatrix}0\\(-1)^{i+1}\\0\end{bmatrix}\right\},\end{aligned} \quad \text{(S2.5B)}$$

$$\begin{aligned}\vec{M}_{\text{Main}\{M\}} = {}& M_{\text{Inner}}V_{\text{Inner}} \\ & \left\{\sum_{i=1}^{3}\delta\left(Z_{\{M\}} - Z_{\text{Main},\{M\}}\right)\delta\left(Y_{\{M\}} - Y_{\text{Inner},i,\{M\}}\right)\delta\left(X_{\{M\}} - X_{\text{Inner},i,\{M\}}\right)\right. \\ & \left.\begin{bmatrix}\sin\left(\frac{2\pi}{3}i - \frac{\pi}{3}\right)\\-\cos\left(\frac{2\pi}{3}i - \frac{\pi}{3}\right)\\0\end{bmatrix}\right\},\end{aligned} \quad \text{(S2.5C)}$$

$$\vec{M}_{\text{Rprog},\{M\}} = M_{\text{Rprog}}V_{\text{Rprog}}\delta\left(Z_{\{M\}} - Z_{\text{Rprog},\{M\}}\right)\delta\left(Y_{\{M\}}\right)\delta\left(X_{\{M\}}\right)\begin{pmatrix}\cos\varphi\\\sin\varphi\\0\end{pmatrix}, \quad \text{(S2.5D)}$$

$$\vec{M}_{\mathrm{Heat},\{M\}} = M_{\mathrm{Heat}} V_{\mathrm{Heat}} \delta\big(Z_{\{M\}} - Z_{\mathrm{Heat},\{M\}}\big) \delta\big(Y_{\{M\}}\big) \delta\big(X_{\{M\}}\big) \begin{pmatrix} \cos\varphi \\ \sin\varphi \\ 0 \end{pmatrix}. \tag{S2.5E}$$

In Eq. (S2.5), $u$ represents the Heaviside step function, $\delta$ represents the Dirac delta function, and the variables $X_{\{M\}}$, $Y_{\{M\}}$, and $Z_{\{M\}}$ represent the coordinates of their respective Cartesian axes according to the soft robot's material coordinate frame. The following parameters are constants, and their physical descriptions and numerical values are summarized in Table S2: $Z_{\mathrm{Tent(soft)},\{M\}}$, $t_{\mathrm{Tent(soft)},\{M\}}$, $b_{\mathrm{Tent}}$, $l_{\mathrm{Tent}}$, $Z_{\mathrm{Tent(rigid)},\{M\}}$, $Y_{\mathrm{Tent(rigid)},1,\{M\}}$, $Y_{\mathrm{Tent(rigid)},2,\{M\}}$, $V_{\mathrm{Six}}$, $Z_{\mathrm{Main},\{M\}}$, $X_{\mathrm{Inner},1,\{M\}}$, $X_{\mathrm{Inner},2,\{M\}}$, $X_{\mathrm{Inner},3,\{M\}}$, $Y_{\mathrm{Inner},1,\{M\}}$, $Y_{\mathrm{Inner},2,\{M\}}$, $Y_{\mathrm{Inner},3,\{M\}}$, $V_{\mathrm{Inner}}$, $Z_{\mathrm{Rprog},\{M\}}$, $V_{\mathrm{Rprog}}$, $Z_{\mathrm{Heat},\{M\}}$, and $V_{\mathrm{Heat}}$. Although the width of our soft robot's tentacles is tapered, the deformation of such magnetic structures can still be modelled as a uniform beam (*8*). Hence, we have made this simplification in Eq. (S2.5A). The magnetization magnitudes of $M_{\mathrm{Tent(soft)}}$, $M_{\mathrm{Tent(rigid)}}$ and $M_{\mathrm{Inner}}$ are characterized in SI Section S1A to be 37.5 kA/m, 88.7 kA/m and 108 kA/m, respectively. Likewise, $M_{\mathrm{Rprog}}$ and $M_{\mathrm{Heat}}$ are evaluated in SI Section S1A to be 7.18 kA/m and 6.52 kA/m, respectively, when the soft robot is reprogrammed to its function modes. When the soft robot is in its locomotion mode, $M_{\mathrm{Rprog}}$ and $M_{\mathrm{Heat}}$ will be negligible (SI Section S1A).

By programming the stiffness profile of the tentacles, these soft components will be able to produce large bending deformation about the $X_{\{M\}}$-axis (Fig. S7). Likewise, the inner beams in the main body of our soft robot are also programmed to have a stiffness profile that will allow them to bend effectively about the $Z_{\{M\}}$-axis (Fig. S8). In general, all of these soft components can be modelled as large deflection beams and the geometries of the deformed robot can be expressed in its local coordinate frame (Fig. S9). Vectors that are expressed in the local coordinate frame are indicated with a subscript $\{L\}$. The local coordinate frame is always assigned in such a way that its $Z_{\{L\}}$-axis is parallel to the soft robot's net magnetic moment. Since the soft robot's net magnetic moment will vary when it is reprogrammed to different modes, we will use $\vec{m}_{\mathrm{Loco}}$, $\vec{m}_{\mathrm{Cut}}$, $\vec{m}_{\mathrm{Disp}}$ and $\vec{m}_{\mathrm{Grip}}$ to represent this vector when the robot is in its locomotion, cutting, drug-dispensing and gripping/storage mode, respectively. Because the deformation of the soft robot will differ when it is in its locomotion and function modes, here we will begin our analysis by assuming that the soft robot is in its locomotion mode.

When the soft robot is reprogrammed to its locomotion mode, $\vec{M}_{\mathrm{Rprog},\{M\}}$ and $\vec{M}_{\mathrm{Heat},\{M\}}$ are assumed to be null vectors. In this mode, only the tentacles of the soft robot will produce bending deformation when an upright $\vec{B}$ is applied (Fig. S7**b**(i)). This bending deformation will occur because the magnetic dipoles in the tentacles will tend to align with the applied $\vec{B}$ (Fig. S7**b**). As the deformed tentacles will endow the soft robot with a net magnetic moment, $\vec{m}_{\mathrm{Loco}}$, which is parallel to $\vec{B}$, the soft robot will not produce any rigid body rotation during the deformation process (Fig. S7). Although the inner beams of the soft robot will also experience a magnetic torque about the $X_{\{M\}}$- and $Y_{\{M\}}$-axes when an upright $\vec{B}$ is applied (Fig. S8**d**), they have negligible deformation about these axes due to their programmed stiffness profile. Furthermore, as the inner beams are arranged in a rotary symmetrical configuration (Fig. S1), they will not contribute to the soft robot's $\vec{m}_{\text{Loco}}$ too.

To analyze the deformation of the soft robot's tentacles (its soft part), we will focus on $\vec{M}_{\text{Tent (soft)},\{M\}}$. Since the tentacles are symmetrical about the $Z_{\{M\}}$-axis of the soft robot, we can simplify the analysis by only considering the tentacles' deformation in the right half space and model this structure as a large deflection beam (Fig. S7). Due to symmetry, the left end of this beam can be considered as a fixed end (Fig. S7). The boundary condition of the beam's right end is modelled as a free end (Fig. S7). To facilitate our subsequent derivations, we also specify a spatial variable, $s_{\text{Tent}}$, along the beam's length (Fig. S7). Upon actuation by an upright $\vec{B}_{\{L\}}$, $\vec{M}_{\text{Tent(soft)},\{M\}}$ will generate a distribution of magnetic torque, $\tau_{x,\{L\}}(s_{\text{Tent}})$, along the beam (Fig. S7):

$$\tau_{x,\{L\}}(s_{\text{Tent}}) = [1 \quad 0 \quad 0]\left\{\left(\mathbf{R}_x(\gamma_{\text{Tent}})\vec{M}_{\text{Tent(soft)},\{M\}}(s_{\text{Tent}})\right) \times \vec{B}_{\{L\}}\right\},$$

where

$$\mathbf{R}_x(\gamma_{\text{Tent}}) = \begin{pmatrix} 1 & 0 & 0 \\ 0 & \cos\gamma_{\text{Tent}} & -\sin\gamma_{\text{Tent}} \\ 0 & \sin\gamma_{\text{Tent}} & \cos\gamma_{\text{Tent}} \end{pmatrix}, \tag{S2.6}$$

and it is a rotation matrix that accounts for the large rotary deformation, $\gamma_{\text{Tent}}(s_{\text{Tent}})$, of the beam (Fig. S7). Based on the free-body diagram of an arbitrary infinitesimal element along the beam (Fig. S7), the following torque balance equation can be derived:

$$-\tau_{x,\{L\}}A_{\text{Tent}}\,\mathrm{d}s_{\text{Tent}} = \frac{\partial M_{\text{b}}}{\partial s_{\text{Tent}}}\,\mathrm{d}s_{\text{Tent}}, \tag{S2.7}$$

where $M_{\text{b}}$ and $A_{\text{Tent}}$ represent the bending moment applied at $s_{\text{Tent}}$ and the cross-sectional area of the beam. It is noteworthy that the Euler-Bernoulli equation also dictates that the following relationship between $M_{\text{b}}$ and $\gamma_{\text{Tent}}$ will hold (*8*):

$$M_{\text{b}} = EI\frac{\partial\gamma_{\text{Tent}}}{\partial s_{\text{Tent}}}, \tag{S2.8}$$

where $E$ and $I$ represent the Young's modulus and the second moment of area of the beam, respectively. By substituting Eq.s (S2.6) and (S2.8) into Eq. (S2.7), we can therefore obtain the governing equation which describes the deformation of this beam:

$$-[1 \quad 0 \quad 0]\left(\left(\mathbf{R}_x(\gamma_{\text{Tent}})\vec{M}_{\text{Tent(soft)},\{M\}}(s_{\text{Tent}})\right) \times \vec{B}_{\{L\}}\right)A_{\text{Tent}} = EI\frac{\partial^2\gamma_{\text{Tent}}}{\partial {s_{\text{Tent}}}^2}. \tag{S2.9}$$

Equation (S2.9) can be solved numerically with the following fixed-free boundary conditions, i.e., $\gamma_{\mathrm{Tent}}(s_{\mathrm{Tent}} = 0) = \frac{\partial \gamma_{\mathrm{Tent}}}{\partial s_{\mathrm{Tent}}}\left(s_{\mathrm{Tent}} = \frac{l_{\mathrm{Tent}}}{2}\right) = 0$, where $l_{\mathrm{Tent}}$ is the total length of the tentacles. It is noteworthy that we have solved the curvatures of the beam by substituting the experimentally obtained $E = 3.96 \times 10^5$ Pa, $|\vec{M}_{\mathrm{Tent(soft)},\{M\}}| = 37.5$ kA/m, and $|\vec{B}_{\{L\}}|$ is specified to be between 5 and 34 mT in Eq. (S2.9). The simulation predictions are compared to the experimental results in SI Section S3A (Fig. S10). The left side of the deformed tentacles can be obtained by mirroring their deformed right beam about the $Z_{\{M\}}$-axis (Fig. S7). Together, the tentacles are predicted to form a 'U'-shaped configuration for the soft robot (Fig. S7**b**). Because Eq. (S2.9) suggests that larger magnetic torques can be generated on the tentacles when $|\vec{B}_{\{L\}}|$ is increased, this implies that the soft tentacles should deform with greater curvatures in such scenarios. It will be beneficial to predict the curvatures of the tentacles accurately because this can in turn allow us to generate reliable locomotion for the soft robot, especially for its two-anchor crawling locomotion. By obtaining its deformed configuration, the soft robot's $\vec{m}_{\mathrm{Loco},\{L\}}$ can also be computed:

$$\vec{m}_{\mathrm{Loco},\{L\}} = \iiint \mathbf{R}_x(\gamma_{\mathrm{Tent}})\vec{M}_{\mathrm{Tent(soft)},\{M\}}\,\mathrm{d}V + \iiint \vec{M}_{\mathrm{Tent(rigid)},\{M\}}\,\mathrm{d}V + \iiint \vec{M}_{\mathrm{Main},\{M\}}\,\mathrm{d}V, \quad \text{(S2.10A)}$$

$$\rightarrow \vec{m}_{\mathrm{Loco},\{L\}} = \iiint \mathbf{R}_x(\gamma_{\mathrm{Tent}})\vec{M}_{\mathrm{Tent(soft)},\{M\}}\,\mathrm{d}V, \quad \text{(S2.10B)}$$

where $V$ represents the volume of each corresponding magnetic component. On the right of Eq. (S2.10A), the first to third integrals represent the net magnetic moment contributed by the deformed tentacles, the sixth-DOF enhancement module and the main body of the soft robot, respectively. Because the reprogrammable module and remote heating component have very small magnetic moments, we assume that they have negligible effects when the soft robot is in its locomotion mode and therefore they are excluded in Eq. (S2.10A). It is noteworthy that the second and third integrals on the right of Eq. (S2.10A) are null vectors because these magnetization profiles have symmetrical configurations. Thus, only the deformed tentacles of the soft robot will contribute to its net magnetic moment when the soft robot is in its locomotion mode (Eq. S2.10B) and this is an important information for describing the soft robot's locomotion in SI Section S2B. Another important point to note is that when a component of the soft robot is undeformed, its magnetization profile will be identical in the material and local coordinate frames. Equations (S2.9-10) imply that the soft robot's $\vec{m}_{\mathrm{Loco},\{L\}}$ will become parallel and anti-parallel to its $Z_{\{M\}}$-axis when its deformed tentacles assume an inverted and upright 'U'-shaped configurations, respectively.

When the soft robot is reprogrammed to a specific function mode, it will theoretically have two types of deformation upon actuation by $\vec{B}$. The first type of deformation will occur when $\vec{B}$ is applied perfectly along the $\vec{m}_{\mathrm{Rprog}}$ and $\vec{m}_{\mathrm{Heat}}$ (Fig. S8**a**). Under such actuation, the soft robot will not produce any rigid body rotations because its net magnetic moment is already aligned to the applied $\vec{B}$. However, its soft inner beams in the main body will still deform upon such actuation because the magnetic moments at the free ends of these beams will tend to align with the applied $\vec{B}$. The second type of deformation will occur when $\vec{B}$ is non-parallel to $\vec{m}_{\text{Rprog}}$ and $\vec{m}_{\text{Heat}}$ (Fig. S8**b-c**). For this type of actuation, the soft tentacles and the beams in the main body will deform concurrently (Fig. S8**b-c**). Together with the magnetic moments from the reprogrammable module and remote heating component, these deformed components will endow the soft robot with a net magnetic moment, which will also tend to align with the applied $\vec{B}$ (Fig. S8**b-c**). In general, it is more challenging to actuate the soft robot based on the first type of deformation because the constructed soft robot's $\vec{M}_{\{M\}}$ may deviate from its theoretical magnetization profile due to fabrications errors. As a result, the second type of deformation is generally observed when we actuate our soft robot in its function modes. Unless specified otherwise, our subsequent discussions will focus on the soft robot's second type of deformation.

Although the soft robot can be reprogrammed to three distinct function modes (Fig.s 1**c**, S5), the physics of the deformation for all these modes are similar. Hence, here we will only analyze the drug-dispensing mode of the soft robot as an example (Fig. S8). When the soft robot is actuated by $\vec{B}$, the deformed robot will be expected to rotate until its $\vec{m}_{\text{Disp}}$ is aligned to the applied field. Once aligned, $\vec{B}$ can be decomposed into two components according to the soft robot's local coordinate frame: $\vec{B}_{\text{Tent},\{L\}}$ and $\vec{B}_{\text{Func},\{L\}}$, which are parallel to the robot's $Z_{\{L\}}$-axis and $\vec{m}_{\text{Rprog}}$, respectively (Fig. S8**b-c**). The first component ($\vec{B}_{\text{Tent},\{L\}}$) will deform the tentacles of the soft robot and their deformation can be predicted by replacing $\vec{B}_{\{L\}}$ with $\vec{B}_{\text{Tent},\{L\}}$ in Eq. (S2.9). On the other hand, the deformation for the inner beams will be dictated by $\vec{B}_{\text{Func},\{L\}}$. When $\vec{B}_{\text{Func},\{L\}}$ is applied, an inner beam in the soft robot's main body, which is magnetized to be parallel to $\vec{B}_{\text{Func},\{L\}}$, will not produce magnetic torques (Fig. S8**b-c**). However, because the magnetic moments of the remaining two inner beams in the main body are non-parallel to $\vec{B}_{\text{Func},\{L\}}$, they will generate a magnetic torque ($T_{z,\{L\}}$) about the $Z_{\{M\}}$-axis at the free end of the beam as they tend to align with $\vec{B}_{\text{Func},\{L\}}$. Since the deformation of these twin beams is similar, we will only analyze the top beam in Fig. S8**b**(i) as an example. This beam will experience a magnetic torque ($T_{z,\{L\}}$) at its free end (Fig. S8**b**(i)):

$$T_{z,\{L\}} = [0 \quad 0 \quad 1]\left\{\left(\iiint \mathbf{R}_z(\gamma_{\text{Intip}})\vec{M}_{\text{Inner},1,\{M\}}\,\mathrm{d}V\right) \times \vec{B}_{\text{Func},\{L\}}\right\},$$

where the magnetization profile of the top beam ($\vec{M}_{\text{Inner},1,\{M\}}$) is given as:

$$\vec{M}_{\text{Inner},1,\{M\}} = M_{\text{Inner}} V_{\text{Inner}} \left\{ \delta\left(Z_{\{M\}} - Z_{\text{Inner},\{M\}}\right) \delta\left(Y_{\{M\}} - Y_{\text{Inner},1,\{M\}}\right) \delta\left(X_{\{M\}} - X_{\text{Inner},1,\{M\}}\right) \begin{bmatrix} \sin\left(\frac{\pi}{3}\right) \\ -\cos\left(\frac{\pi}{3}\right) \\ 0 \end{bmatrix} \right\}$$

and

$$\mathbf{R}_z\left(\gamma_{\text{Intip}}\right) = \begin{pmatrix} \cos\gamma_{\text{Intip}} & -\sin\gamma_{\text{Intip}} & 0 \\ \sin\gamma_{\text{Intip}} & \cos\gamma_{\text{Intip}} & 0 \\ 0 & 0 & 1 \end{pmatrix}, \tag{S2.11}$$

and it is a standard $Z$-axis rotation matrix that accounts for the orientation change of the magnetic moment, which is located at the beam's free end as this beam undergoes rotary deformation ($\gamma_{\text{Intip}}$). In theory, $T_{z,\{L\}}$ will induce a constant bending moment across the inner beams, which has a magnitude equal to $T_{z,\{L\}}$. By using the Euler-Bernoulli equation, we can obtain the governing equation that describes the bending deflection of the inner beams:

$$\begin{bmatrix} 0 & 0 & 1 \end{bmatrix} \left( \left( \iiint \mathbf{R}_z\left(\gamma_{\text{Intip}}\right) \vec{M}_{\text{Inner},1,\{M\}} \, \mathrm{d}V \right) \times \vec{B}_{\text{Func},\{L\}} \right) = EI \frac{\partial \gamma_{\text{Inner}}}{\partial s_{\text{Inner}}}, \tag{S2.12}$$

where $s_{\text{Inner}}$ is a spatial variable along the beam's length, and $\gamma_{\text{Inner}}$ is the rotary deformation at $s_{\text{Inner}}$ (Fig. S8**d**). By integrating Eq. (S2.12) with respect to $s_{\text{Inner}}$ and subsequently substituting the boundary condition at the beam's fixed and free ends, we can obtain the governing equation that describes the beam's deformation:

$$\frac{1}{2} M_{\text{Inner}} \left|\vec{B}_{\text{Func},\{L\}}\right| \left(\sqrt{3}\cos\gamma_{\text{Intip}} + \sin\gamma_{\text{Intip}}\right) = EI \frac{\gamma_{\text{Intip}}}{l_{\text{Inner}}}, \tag{S2.13}$$

where $l_{\text{Inner}}$ represents the length of the beam, respectively. The deformed beams will not immediately make the soft robot dispense its drugs upon actuation because there is a space between the inner beams and the drug-dispensing mechanism (modelled as non-uniform beams). By numerically solving Eq. (S2.13) with $E = 5.70 \times 10^5$ Pa and $I = 2.43 \times 10^{-17}$ m$^4$, it is predicted that $\left|\vec{B}\right| > 1.63$ mT will be required for the deformed inner beams to establish contact with the drug-dispensing mechanism. Upon contact, the deformed inner beams will also have to transmit sufficient forces to the drug-dispensing mechanism to activate it. However, it is very challenging to describe such interactions because the drug-dispensing mechanism has very complex geometries. Nonetheless, we highlight that $\left|\vec{B}\right|$ needs to be higher than a specific threshold before the drug-dispensing mechanism can be activated. This is an advantageous feature because it can allow us to effectively decouple our soft robot's locomotion and surgical functions, allowing our soft robot to successfully locomote without activating its function by restricting $\left|\vec{B}\right|$ to be smaller than this threshold. To evaluate this threshold, we have performed corresponding experiments and these experiments are described in SI Section S3B. Despite so, Eq.

(S2.13) does predict that when $\left|\vec{B}\right|$ increases, the induced bending moment on the inner beams will also increase and this should make the beams deflect more and open the doors of the drug-dispensing mechanism wider (Fig. S12**a**(ii)). To have a better understanding on the deformation of the robot's inner beams when $\left|\vec{B}\right| > 1.63$ mT, we have therefore characterized the opening of the drug-dispensing mechanism via experimental means in SI Section S3B. Because the deformed inner beams and soft tentacles will also contribute to the net magnetic moment of the soft robot ($\vec{m}_{\text{Disp}}$), the direction of $\vec{m}_{\text{Disp}}$ is expected to change when the soft tentacles and inner beams deform more. As such effects are in principle hard to analyze via theory, we have also characterized how the direction of $\vec{m}_{\text{Disp}}$ varies with respect to different $\left|\vec{B}\right|$ through the experiments. Similar characterizations have also been conducted on the soft robot when it is reprogrammed to its gripping and cutting modes (SI Section S3B).

In their deformed configuration, the three inner beams in the soft robot's main body will contribute a magnetic moment ($\vec{m}_{\text{Main}}$) of:

$$\vec{m}_{\text{Main},\{L\}} = \sum_{i=1}^{3} \iiint \mathbf{R}_z(\gamma_{\text{Inner},i})\vec{M}_{\text{Inner},i,\{M\}}\,\mathrm{d}V,$$

where the magnetization profile of each inner beam is expressed as:

$$\vec{M}_{\text{Inner},i,\{M\}} = M_{\text{Inner}}V_{\text{Inner}} \left\{\delta\left(Z_{\{M\}} - Z_{\text{Inner},\{M\}}\right)\delta\left(Y_{\{M\}} - Y_{\text{Inner},i,\{M\}}\right)\delta\left(X_{\{M\}} - X_{\text{Inner},i,\{M\}}\right)\begin{bmatrix} \sin\left(\frac{2\pi}{3}i - \frac{\pi}{3}\right) \\ -\cos\left(\frac{2\pi}{3}i - \frac{\pi}{3}\right) \\ 0 \end{bmatrix}\right\}$$

and

$$\mathbf{R}_z(\gamma_{\text{Inner},i}) = \begin{pmatrix} \cos\gamma_{\text{Inner},i} & -\sin\gamma_{\text{Inner},i} & 0 \\ \sin\gamma_{\text{Inner},i} & \cos\gamma_{\text{Inner},i} & 0 \\ 0 & 0 & 1 \end{pmatrix}, \qquad (S2.14)$$

where $i$ is a counter for the summation of the three inner beams' magnetic moments (Fig. S8**b-c**). Due to symmetry, $\gamma_{\text{Inner},1}$ and $\gamma_{\text{Inner},2}$ are equal but opposite of one another, while $\gamma_{\text{Inner},3}$ will be zero as the third inner beam will not deform upon actuation. The net magnetic moment of the robot in the drug-dispensing mode, $\vec{m}_{\text{Disp},\{L\}}$, can be computed as:

$$\vec{m}_{\text{Disp},\{L\}} = \vec{m}_{\text{Main},\{L\}} + \iiint \mathbf{R}_x(\gamma_{\text{Tent}})\vec{M}_{\text{Tent(soft)},\{M\}}\,\mathrm{d}V + \iiint \vec{M}_{\text{Tent(rigid)},\{M\}}\,\mathrm{d}V + \iiint \vec{M}_{\text{Rprog},\{M\}}\,\mathrm{d}V + \iiint \vec{M}_{\text{Heat},\{M\}}\,\mathrm{d}V, \qquad (S2.15A)$$

$$\rightarrow \vec{m}_{\text{Disp},\{L\}} = \vec{m}_{\text{Main},\{L\}} + \iiint \mathbf{R}_x(\gamma_{\text{Tent}})\vec{M}_{\text{Tent(soft)},\{M\}}\,\mathrm{d}V + \iiint \vec{M}_{\text{Rprog},\{L\}}\,\mathrm{d}V + \iiint \vec{M}_{\text{Heat},\{L\}}\,\mathrm{d}V. \quad \text{(S2.15B)}$$

On the right-hand side of Eq. (S2.15A), the first to fourth integrals represent the net magnetic moment contribution from the soft robot's deformed tentacles, sixth-DOF enhancement module, reprogrammable module and remote heating component, respectively. Because the magnetization profile of the sixth-DOF enhancement module has a symmetrical configuration, its integral is a null vector. Furthermore, since the reprogrammable module and remote heating component are rigid and non-deformable, their magnetization profiles are identical in the material and local coordinate frames. Equation (S2.15) determines the net magnetic moment of the robot when it is in the drug-dispensing mode. Similar analyses can also be performed to determine net magnetic moment of our robot when it is reprogrammed to the cutting and gripping modes.

In general, the analysis from Eq.s (S2.6-15) and the characterization results in SI Section S3 will be critical for understanding the soft robot's deformation characteristics and net magnetic moment. Such information will be essential for us to execute the soft robot's surgical functions and locomotion.

## B. Actuation in six degrees-of-freedom

Once the deformation of our soft robot is predicted via the theoretical analyses in SI Section S2A or the characterizations in SI Section S3A, we can actuate the robot to execute its six-DOF locomotion. This is true regardless of whether the soft robot is reprogrammed to its locomotion or function modes. To analyze such actuation methods, we define respective local coordinate frames (denoted with a subscript $\{L\}$) for the soft robot when it is reprogrammed to a specific mode (Fig. S9). The local coordinate frames are always assigned in such a way that its Cartesian $Z$-axis is always parallel to the deformed robot's net magnetic moment, e.g., ($\vec{m}_{\text{Loco},\{L\}}$ is parallel to its $Z_{\text{Loco},\{L\}}$-axis). An advantage of having such assignments is that we can easily define the deformed robot's sixth-DOF motion to be the rotation about its Cartesian $Z$-axis. After the local coordinate frames have been assigned, the analysis will be similar regardless of whether the soft robot is reprogrammed to its locomotion or function modes. Hence, here we will primarily perform these analyses with the locomotion mode of the robot as an example and provide essential discussions for the other function modes. To represent the axes of the soft robot in its different operational modes, we add a description to their axis, e.g., $Z_{\text{Loco},\{L\}}$-, $Z_{\text{Disp},\{L\}}$-, $Z_{\text{Cut},\{L\}}$- and $Z_{\text{Grip},\{L\}}$-axes refer to the robot's $Z_{\{L\}}$-axis when it is in the locomotion, drug-dispensing, cutting and gripping mode, respectively (Fig. S9).

Under its locomotion mode, the soft robot's local coordinate frame will be assigned differently when its tentacles deform into different configurations. This is because when the tentacles assume an inverted 'U'-shaped configuration, the soft robot's $\vec{m}_{\text{Loco},\{L\}}$ will be parallel to its $Z_{\{M\}}$-axis, but its net magnetic moment will become anti-parallel to $Z_{\{M\}}$-axis when the tentacles deform into an upright 'U'-shaped configuration (SI Section S2A). Therefore, the deformed robot's local coordinate frame in the latter scenario has a relative 180° rotation about its $Y_{\{M\}}$-axis compared to its local coordinate frame in the former scenario. By understanding how the soft robot's local coordinate frames are assigned, the following wrench analysis will be applicable on our robot regardless of its tentacles' deformed configuration. Based on the assigned local coordinate frame (e.g., Fig. S9**a**(iii)), the net wrench applied on the deformed robot, $\vec{T}_{\{L\}}$ (magnetic torque) and $\vec{F}_{\{L\}}$ (magnetic force), with respect to the applied $\vec{B}_{\{L\}}$ and $\vec{B}_{\mathrm{Grad},\{L\}}$ can be expressed as (Fig. S9**e**) (*6*):

$$\begin{pmatrix}\vec{T}\\ \vec{F}\end{pmatrix}_{\{L\}} = \begin{pmatrix} \iiint \vec{M}_{\mathrm{Loco},\{L\}} \times \vec{B}_{\{L\}}\,\mathrm{d}V + \iiint \vec{r}_{\{L\}} \times \left( \begin{bmatrix} \dfrac{\partial \vec{B}_{\{L\}}}{\partial x_{\{L\}}} & \dfrac{\partial \vec{B}_{\{L\}}}{\partial y_{\{L\}}} & \dfrac{\partial \vec{B}_{\{L\}}}{\partial z_{\{L\}}} \end{bmatrix}^{\mathrm{T}} \vec{M}_{\mathrm{Loco},\{L\}} \right) \mathrm{d}V \\ \iiint \left(\vec{M}_{\mathrm{Loco},\{L\}} \cdot \nabla\right) \vec{B}_{\{L\}}\,\mathrm{d}V \end{pmatrix}$$

$$= \mathbf{D}_{\mathrm{Loco}} \begin{pmatrix}\vec{B}\\ \vec{B}_{\mathrm{Grad}}\end{pmatrix}_{\{L\}}, \qquad \text{(S2.16A)}$$

where the magnetization profile of the deformed robot is:

$$\vec{M}_{\mathrm{Loco},\{L\}} = \mathbf{R}_x(\gamma_{\mathrm{Tent}})\vec{M}_{\mathrm{Tent(soft)},\{M\}} + \vec{M}_{\mathrm{Tent(rigid)},\{M\}} + \vec{M}_{\mathrm{Inner},\{M\}},$$

$$\rightarrow \vec{M}_{\mathrm{Loco},\{L\}} = \mathbf{R}_x(\gamma_{\mathrm{Tent}})\vec{M}_{\mathrm{Tent(soft)},\{M\}} + \vec{M}_{\mathrm{Tent(rigid)},\{L\}} + \vec{M}_{\mathrm{Inner},\{L\}}, \qquad \text{(S2.16B)}$$

and $\vec{r}_{\{L\}}$ is the displacement vector from the soft robot's center of mass to a point of interest and $\mathbf{D}_{\mathrm{Loco}}$ is known as the $6 \times 8$ design matrix for the locomotion mode:

$$\mathbf{D}_{\mathrm{Loco}} = \begin{pmatrix} 0 & -|\vec{m}_{\mathrm{Loco}}| & 0 & 0 & d_2 & 0 & 0 & 0 \\ |\vec{m}_{\mathrm{Loco}}| & 0 & 0 & d_6 & 0 & 0 & 0 & 0 \\ 0 & 0 & 0 & 0 & 0 & 0 & 0 & d_{15} \\ 0 & 0 & 0 & |\vec{m}_{\mathrm{Loco}}| & 0 & 0 & 0 & 0 \\ 0 & 0 & 0 & 0 & |\vec{m}_{\mathrm{Loco}}| & 0 & 0 & 0 \\ 0 & 0 & 0 & 0 & 0 & |\vec{m}_{\mathrm{Loco}}| & 0 & 0 \end{pmatrix}, \qquad \text{(S2.17A)}$$

where

$$d_2 = \iiint (r_y M_y - r_z M_z)\,\mathrm{d}V,\ d_6 = \iiint (r_z M_z - r_x M_x)\,\mathrm{d}V,\ d_{15} = \iiint (r_x M_x - r_y M_y)\,\mathrm{d}V, \qquad \text{(S2.17B)}$$

and $r_x$, $r_y$ and $r_z$ are the Cartesian components of $\vec{r}$. Likewise, $M_x$, $M_y$ and $M_z$ are the Cartesian components of $\vec{M}_{\mathrm{Loco},\{L\}}$. As the geometries of the soft robot will change as it undergoes different amounts of deformation, the absolute values of $d_2$, $d_6$, and $d_{15}$ will also change accordingly. Because $d_{15}$ is non-zero, this implies that the rank of $\mathbf{D}_{\mathrm{Loco}}$ is six and such a full-ranked matrix can endow the soft robot with six-DOF motions (*7, 9*). Although it is easy to express the wrench applied on the soft robot based on its local coordinate frame, it will be challenging to control the robot's orientation with this frame (*7*, *9*). Hence, Eq. (S2.16A) is reanalyzed according to the soft robot's intermediate coordinate frame via the following mapping (Fig. S9) (*7*, *9*):

$$\begin{pmatrix} \vec{T} \\ \vec{F} \end{pmatrix}_{\{I\}} = \mathbf{C}_{\mathrm{Loco}}(\theta) \begin{pmatrix} \vec{B} \\ \vec{B}_{\mathrm{Grad}} \end{pmatrix}_{\{I\}},$$

where

$$\mathbf{C}_{\mathrm{Loco}}(\theta) = \begin{pmatrix} \mathbf{R}_z(\theta) & \mathbf{0}_{3\times3} \\ \mathbf{0}_{3\times3} & \mathbf{R}_z(\theta) \end{pmatrix} \mathbf{D}_{\mathrm{Loco}} \mathbf{A},$$

$$\mathbf{R}_z(\theta) = \begin{pmatrix} \cos\theta & -\sin\theta & 0 \\ \sin\theta & \cos\theta & 0 \\ 0 & 0 & 1 \end{pmatrix},$$

$$\mathbf{A} = \begin{bmatrix} \cos\theta & \sin\theta & 0 & 0 & 0 & 0 & 0 & 0 \\ -\sin\theta & \cos\theta & 0 & 0 & 0 & 0 & 0 & 0 \\ 0 & 0 & 1 & 0 & 0 & 0 & 0 & 0 \\ 0 & 0 & 0 & \cos\theta & \sin\theta & 0 & 0 & 0 \\ 0 & 0 & 0 & -\sin\theta & \cos\theta & 0 & 0 & 0 \\ 0 & 0 & 0 & 0 & 0 & 1 & 0 & 0 \\ 0 & 0 & 0 & 0 & 0 & -\sin^2\theta & \cos 2\theta & -\sin 2\theta \\ 0 & 0 & 0 & 0 & 0 & 0.5\sin 2\theta & \sin 2\theta & \cos 2\theta \end{bmatrix}, \tag{S2.18}$$

and $\theta$ represents the desired sixth-DOF angular displacement of the soft robot. The matrix, $\mathbf{C}_{\mathrm{Loco}}(\theta)$, is known as the control matrix and it is also a full rank matrix because $\mathbf{R}_z$, $\mathbf{A}$, and $\mathbf{D}_{\mathrm{Loco}}$ are all full rank matrices (*7*). Vectors expressed in the intermediate coordinate frame are indicated with a subscript $\{I\}$.

Equation (S2.18) can be solved in such a way that the soft robot's desired orientation can become a minimum potential energy configuration (*7*). Using such an actuation strategy, the soft robot will continuously experience restoring torques until it rotates to the desired orientation (Fig. S9**e**). It will also be possible to apply desired magnetic forces on the soft robot with this actuation strategy. To implement this actuation strategy, we assign $\vec{T}_{\{I\}}$ to be a null vector so that the soft robot can remain in its rotational equilibrium state at the desired $\theta$. Equation (S2.18) can then be solved as:

$$\begin{pmatrix} \vec{B} \\ \vec{B}_{\mathrm{Grad}} \end{pmatrix}_{\{I\}} = \mathbf{C}_{\mathrm{Loco}}{}^{\mathrm{T}} \left[ \mathbf{C}_{\mathrm{Loco}} \mathbf{C}_{\mathrm{Loco}}{}^{\mathrm{T}} \right]^{-1} \begin{pmatrix} \vec{0} \\ \vec{F} \end{pmatrix}_{\{I\}} + k_1 \begin{pmatrix} 0 \\ 0 \\ 1 \\ 0 \\ 0 \\ 0 \\ 0 \\ 0 \end{pmatrix}_{\{I\}} + k_2 \begin{pmatrix} 0 \\ 0 \\ 0 \\ 0 \\ 0 \\ 0 \\ 1 \\ -\tan 2\theta \end{pmatrix}_{\{I\}}. \tag{S2.19}$$

Equation (S2.19) is the general solution of the actuating magnetic signals for the soft robot to achieve six-DOF motions. This solution includes a particular solution that is obtained via pseudo-inverse of $\mathbf{C}_{\mathrm{Loco}}$ and this solution is represented by the first right-hand component. The remaining right-hand components in Eq. (S2.19) represent the homogeneous solutions, which are formed by the null space vectors of $\mathbf{C}_{\text{Loco}}$. The variables, $k_1$ and $k_2$, are the scale factors of their respective null space vectors.

The particular solution dictates that the soft robot will achieve rotational equilibrium at the desired orientation while $\vec{F}_{\{I\}}$ can still be applied on the robot. The first null space vector in Eq. (S2.19), $(0\ 0\ 1\ 0\ 0\ 0\ 0\ 0)^T_{\{I\}}$, increases $|\vec{B}|$ so that the restoring torque that aligns the soft robot's $\vec{m}_{\text{Loco}}$ to its desired direction can become stronger. However, the first null space vector cannot generate the restoring torque about the sixth-DOF axis of the soft robot (*15-17*) (i.e., the rotation about $\vec{m}_{\text{Loco}}$). Instead, the sixth-DOF restoring torque of the soft robot can be generated by the second null space vector (*7*). By tuning $\vec{B}_{\text{Grad},\{L\}}$ via the second null space vector, we can therefore command the soft robot's sixth-DOF angular displacement to self-align to the desired $\theta$ (*7*, *9*). To maximize the restoring sixth-DOF torque that is applied on the soft robot, we can maximize $k_2$ based on the maximum capacity of our electromagnetic coil system (*7*). Although it is also ideal to maximize the value of $k_1$ so that the remaining two axes of restoring torques can be maximized too, its magnitude will also dictate the curvature of the robot's deformed tentacles. This is because the value of $k_1$ is equal to $|\vec{B}|$. Hence, $k_1$ is always tuned in such a way that we can achieve the desired curvature for the tentacles so that the soft robot can successfully execute its locomotion. Once $k_1$ and $k_2$ are specified, the required actuating magnetic fields are expressed according to the global coordinate frame of our electromagnetic coil system via the following mapping (Fig. S9):

$$\begin{pmatrix}\vec{B}\\ \vec{B}_{\text{Grad}}\end{pmatrix}_{\{G\}} = \begin{bmatrix}\mathbf{R}_x(\alpha)\mathbf{R}_y(\beta) & \mathbf{0}\\ \mathbf{0} & \mathbf{A}_2(\alpha,\beta)\end{bmatrix}\begin{pmatrix}\vec{B}\\ \vec{B}_{\text{Grad}}\end{pmatrix}_{\{I\}},$$

where

$$\mathbf{A}_2(\alpha,\beta) = \begin{bmatrix} \cos\alpha\cos 2\beta & \sin\alpha\sin\beta & \cos\alpha\sin 2\beta & \frac{1}{2}\cos\alpha\sin 2\beta & \sin\alpha\cos\beta \\ \frac{1}{2}\cos 2\alpha\sin 2\beta & \cos 2\alpha\cos\beta & -\frac{1}{2}\sin 2\alpha\, cos2\beta & \frac{1}{2}\sin 2\alpha\,(1+\sin^2\beta) & -\cos 2\alpha\sin\beta \\ -\cos^2\alpha\sin 2\beta & \sin 2\alpha\cos\beta & \cos^2\alpha\cos 2\beta & \sin^2\alpha - \cos^2\alpha\sin^2\beta & -\sin 2\alpha\sin\beta \\ -\sin^2\alpha\sin 2\beta & -\sin 2\alpha\cos\beta & \sin^2\alpha\cos 2\beta & \cos^2\alpha - \sin^2\alpha\sin^2\beta & \sin 2\alpha\sin\beta \\ -\sin\alpha\cos 2\beta & \cos\alpha\sin\beta & -\sin\alpha\sin 2\beta & -\frac{1}{2}\sin\alpha\sin 2\beta & \cos\alpha\cos\beta \end{bmatrix},$$

$$\mathbf{R}_x(\alpha) = \begin{pmatrix}1 & 0 & 0\\ 0 & \cos\alpha & -\sin\alpha\\ 0 & \sin\alpha & \cos\alpha\end{pmatrix},$$

$$\mathbf{R}_y(\beta) = \begin{pmatrix}\cos\beta & 0 & \sin\beta\\ 0 & 1 & 0\\ -\sin\beta & 0 & \cos\beta\end{pmatrix}, \qquad \text{(S2.20)}$$

where $\alpha$ and $\beta$ represent the desired angular displacements about their respective axes. Vectors expressed in the global coordinate frame are indicated with a subscript $\{G\}$. By controlling $\alpha$ and $\beta$, we can therefore control the direction of $\vec{B}$ and dictate the orientation of $\vec{m}_{\text{Loco}}$. Equation (S2.20) concludes our six-DOF actuation method, and we can use such methods to fully control the soft robot's orientation and applied desired magnetic forces on it. We can also command our soft robot to follow angular trajectories by first discretizing the trajectories into a series of desired angular displacements, and subsequently make all these orientations into a minimum potential energy configuration. To execute the rolling and two-anchor crawling locomotion of the soft robot, we can also adjust $k_1$ to vary temporally so that the tentacles can produce the necessary time-varying shapes.

Similar actuation strategies can be implemented so that our soft robot can execute its locomotion while it is reprogrammed to different function modes. These can be accomplished by replacing $\mathbf{D}_{\text{Loco}}$ in Eq. (S2.18) with the respective **D** matrices to compute the corresponding **C** matrices of the soft robot when it is reprogrammed to its function modes. The soft robot will have the following $\mathbf{D}_{\text{Disp}}$ matrix when it is in its drug-dispensing mode:

$$\mathbf{D}_{\text{Disp}} = \begin{pmatrix} 0 & -|\vec{m}_{\text{Disp}}| & 0 & 0 & d_2 & 0 & 0 & 0 \\ |\vec{m}_{\text{Disp}}| & 0 & 0 & d_6 & 0 & 0 & 0 & 0 \\ 0 & 0 & 0 & 0 & 0 & 0 & 0 & d_{15} \\ 0 & 0 & 0 & |\vec{m}_{\text{Disp}}| & 0 & 0 & 0 & 0 \\ 0 & 0 & 0 & 0 & |\vec{m}_{\text{Disp}}| & 0 & 0 & 0 \\ 0 & 0 & 0 & 0 & 0 & |\vec{m}_{\text{Disp}}| & 0 & 0 \end{pmatrix}. \quad \text{(S2.21)}$$

Likewise, the soft robot will have the following **D** matrices when it is in its cutting ($\mathbf{D}_{\text{Cut}}$) and gripping/storage modes ($\mathbf{D}_{\text{Grip}}$):

$$\mathbf{D}_{\text{Cut}} = \begin{pmatrix} 0 & -|\vec{m}_{\text{Cut}}| & 0 & 0 & d_2 & 0 & 0 & d_5 \\ |\vec{m}_{\text{Cut}}| & 0 & 0 & d_6 & 0 & d_8 & d_9 & 0 \\ 0 & 0 & 0 & 0 & d_{12} & 0 & 0 & d_{15} \\ 0 & 0 & 0 & |\vec{m}_{\text{Cut}}| & 0 & 0 & 0 & 0 \\ 0 & 0 & 0 & 0 & |\vec{m}_{\text{Cut}}| & 0 & 0 & 0 \\ 0 & 0 & 0 & 0 & 0 & |\vec{m}_{\text{Cut}}| & 0 & 0 \end{pmatrix}, \quad \text{(S2.22)}$$

and

$$\mathbf{D}_{\text{Grip}} = \begin{pmatrix} 0 & -|\vec{m}_{\text{Grip}}| & 0 & 0 & d_2 & 0 & 0 & d_5 \\ |\vec{m}_{\text{Grip}}| & 0 & 0 & d_6 & 0 & d_8 & d_9 & 0 \\ 0 & 0 & 0 & 0 & d_{12} & 0 & 0 & d_{15} \\ 0 & 0 & 0 & |\vec{m}_{\text{Grip}}| & 0 & 0 & 0 & 0 \\ 0 & 0 & 0 & 0 & |\vec{m}_{\text{Grip}}| & 0 & 0 & 0 \\ 0 & 0 & 0 & 0 & 0 & |\vec{m}_{\text{Grip}}| & 0 & 0 \end{pmatrix},$$

where

$$d_5 = \iiint -r_z M_x \,\mathrm{d}V, d_8 = \iiint (-r_z M_x - r_x M_z) \,\mathrm{d}V, d_9 = \iiint -r_z M_x \,\mathrm{d}V, d_{12} = \iiint r_x M_z \,\mathrm{d}V. \quad \text{(S2.23)}$$

Because $\mathbf{D}_{\text{Disp}}$, $\mathbf{D}_{\text{Cut}}$ and $\mathbf{D}_{\text{Grip}}$ are all full-ranked matrices, we can therefore conclude that our soft robot has six-DOF in its function modes too. By substituting the respective control matrices, $\mathbf{C}_{\text{Disp}}$, $\mathbf{C}_{\text{Cut}}$ and $\mathbf{C}_{\text{Grip}}$, into Eq. (S2.19), we can therefore execute Eq.s (S2.19-20) to solve for the required actuating fields of our soft robot. The magnitude of the actuating fields can likewise be tuned via the scale factors, $k_1$ and $k_2$. By controlling the magnitude of $k_1$, we can also tune the deformation of the soft robot's tentacles and inner beams such that it can execute its desired locomotion and surgical functionalities in the respective function modes. Figure S9**b-d** shows all the assigned coordinate frames when the soft robot is in its respective function modes. To simplify our discussions, we have assigned these coordinate frames with the assumption that the robot's tentacles are not deformed.

## SI Section S3. Experimental characterization for the deformation of beams

In this section, we will report the experiments pertaining to the deformation of the soft robot when the robot is reprogrammed to its locomotion (Section S3A) and function modes (Section S3B).

### A. Deformation in locomotion mode

When our robot is in its locomotion mode, only its soft tentacles will deform upon actuation by $\vec{B}$ (SI Section S2A). To characterize the deformation of the soft robot in its locomotion mode, we invert the orientation of the robot and observe its deformation by applying a $\vec{B}$ along the $Z_{\text{Loco},\{L\}}$-axis (Fig. S10). For these experiments, the magnitude of $\vec{B}$ is varied from 5 mT to 30 mT at intervals of 5 mT and we also actuate the soft robot with the largest $|\vec{B}|$ (34 mT), which our electromagnetic coil system can produce. Because the soft robot has been inverted in these experiments, its tentacles have deformed to inverted 'U'-shaped configurations that have varying curvatures (Fig. S10). Theoretically, steeper curvatures should be observed when $|\vec{B}|$ is increased (Eq. (S2.9)). To validate this hypothesis, we have also conducted corresponding simulations by solving Eq. (S2.9) and the simulated predictions are compared to the experimental data by superimposing them together. In general, the simulation predictions agree with the experimental data, but the experimental deformations are observed to be always smaller (Fig. S10). The experimental deflections of the soft tentacles are relatively smaller than the simulations in an ideal environment potentially due to them deforming against gravity. Another potential source of error is that our soft robot may have residual strain energy during the demolding step of our fabrication process and such residual strain energy may also cause our soft robot to deflect less in these experiments (*12*). Despite having some deviations between the simulations and experiments, we remark that the deformation of the soft tentacles can still be controlled well such that the soft robot can always execute its rolling and two-anchor crawling locomotion successfully. A noteworthy point is that the soft robot's $\vec{m}_{\text{Loco}}$, which is generated by the deformed soft tentacles, will always be parallel to the applied $\vec{B}$ since $\vec{M}_{\text{Tent(soft)},\{M\}}$ has a symmetrical configuration. Hence, the soft robot will not rotate when the magnitude of $\vec{B}$ changes, allowing the motion of our soft robot to be more deterministic and thus easy to control.

B. Deformation in function modes

When the soft robot is reprogrammed to a specific function mode, our analysis in SI Section S2A predicts that the soft tentacles and inner beams will deform more when $|\vec{B}|$ is increased. As these components deform, the direction of the soft robot's net magnetic moment will change accordingly too. To characterize the relationship between the direction of the soft robot's net magnetic moment and the applied $|\vec{B}|$, we introduce the concept of deviation angles for our experiments. Each function mode of the soft robot will have two respective deviation angles. When the tentacles of the soft robot are deformed into an inverted 'U'-shaped configuration, the robot has deviation angles of $\xi_{\text{Disp},1}$, $\xi_{\text{Cut},1}$ and $\xi_{\text{Grip},1}$ when it is reprogrammed to its drug-dispensing, cutting and gripping/storage modes, respectively. Conversely, when soft robot's tentacles assume an upright 'U'-shaped configuration, the robot's deviation angles are $\xi_{\text{Disp},2}$, $\xi_{\text{Cut},2}$ and $\xi_{\text{Grip},2}$ for its respective function modes.

In theory, the mathematical relationship between the soft robot's locomotion and function modes' magnetic moments can be expressed as (Fig.s 1**b**(ii), S8**b**(ii), **c**(ii) and SI Section S2A):

<u>Inverted 'U'-shaped tentacles</u>

$$\hat{\vec{m}}_{\text{Disp},\{L\}} = \mathbf{R}_z\left(\varphi_{\text{Disp}} - \frac{\pi}{2}\right)\mathbf{R}_x(-\xi_{\text{Disp},1})\hat{\vec{m}}_{\text{Loco},\{L\}},$$
$$\hat{\vec{m}}_{\text{Cut},\{L\},} = \mathbf{R}_z\left(\varphi_{\text{Cut}} - \frac{\pi}{2}\right)\mathbf{R}_x(-\xi_{\text{Cut},1})\hat{\vec{m}}_{\text{Loco},\{L\}}$$
$$\hat{\vec{m}}_{\text{Grip},\{L\}} = \mathbf{R}_z\left(\varphi_{\text{Grip}} - \frac{\pi}{2}\right)\mathbf{R}_x(-\xi_{\text{Grip},1})\hat{\vec{m}}_{\text{Loco},\{L\}}, \qquad \text{(S3.1A)}$$

<u>Upright 'U'-shaped tentacles</u>

$$\hat{\vec{m}}_{\text{Disp},\{L\}} = \mathbf{R}_z\left(-\varphi_{\text{Disp}} + \frac{3\pi}{2}\right)\mathbf{R}_x(\xi_{\text{Disp},2})\hat{\vec{m}}_{\text{Loco},\{L\}},$$
$$\hat{\vec{m}}_{\text{Cut},\{L\}} = \mathbf{R}_z\left(-\varphi_{\text{Cut}} + \frac{3\pi}{2}\right)\mathbf{R}_x(\xi_{\text{Cut},2})\hat{\vec{m}}_{\text{Loco},\{L\}},$$
$$\hat{\vec{m}}_{\text{Grip},\{L\}} = \mathbf{R}_z\left(-\varphi_{\text{Grip}} + \frac{3\pi}{2}\right)\mathbf{R}_x(\xi_{\text{Grip},2})\hat{\vec{m}}_{\text{Loco},\{L\}}, \qquad \text{(S3.1B)}$$

where all the net magnetic moments in Eq. (S3.1) are expressed as unit vectors.

To experimentally characterize the deviation angles, we first reprogram our soft robot to a corresponding function mode. An upright $\vec{B}$ is then applied on the soft robot and this deformed robot is be expected to rotate until its net magnetic moment aligns with $\vec{B}$. Based on the deformed robot's orientation, we will then gradually rotate $\vec{B}$ about the robot's $X_{\{L\}}$-axis until the robot resumes an upright orientation. The corresponding deviation angle of the soft robot will be equal to the required amount of rotation for $\vec{B}$ (Fig. S8**b**(ii), **c**(ii)). The obtained experimental data of these experiments are summarized in Fig. S11. In these experiments, $|\vec{B}|$ is varied from 5 mT to 25 mT at intervals of 5 mT and each data point is repeated with five trials (Fig. S11). The deviation angles are different when the soft robot is reprogrammed to different function modes because its inner beams will deform differently in such scenarios (Fig. S11). The characterization in Fig. S11 will be critical for controlling the soft robot's locomotion when it is reprogrammed to its function mode.

In addition, we also characterize the extension of the soft robot's cutting tool as well as the width of its drug chamber's and grippers' opening when the robot is actuated in its respective function modes. According to our theory in SI Section S2A, the soft robot's mechanical functions can be actuated by deforming the inner soft beams in its main body. These inner beams are predicted to deflect more when the applied $\vec{B}$ along $\vec{m}_{\mathrm{Rprog}}$ is increased, i.e., $\vec{B}_{\mathrm{Func}}$ (Fig. S8 and SI Section S2A). This is because when $\vec{B}_{\mathrm{Func}}$ is increased, the free ends of these inner beams will be able to produce larger magnetic torques that will bend the beams more (Eq. S2.13). The degree of actuation for each function is quantified differently: for the drug-dispensing mode, we measure the shortest distance between the doors of its drug chamber, $a_{\mathrm{Disp}}$ (Fig. S12**a**); for the cutting mode, we measure the extension of the cutting tool, $a_{\mathrm{Cut}}$ (Fig. S12**b**); and for the gripping/storage mode, the shortest distance between the grippers ($a_{\mathrm{Grip}}$) is measured (Fig. S12**c**). To make these measurements, the soft robot is first reprogrammed to its respective function mode before being constrained to an experimental jig that prevents it from rotating. Subsequently, we apply $\vec{B}_{\mathrm{Func}}$ to the robot, and $|\vec{B}_{\mathrm{Func}}|$ is varied from 0 to 30 mT at intervals of 5 mT. We also measure $a_{\mathrm{Disp}}$, $a_{\mathrm{Cut}}$, and $a_{\mathrm{Grip}}$ when the soft robot is actuated by the largest $\vec{B}_{\mathrm{Func}}$ (34 mT) of our electromagnetic coil system. The relationship between these measured parameters and $\vec{B}_{\mathrm{Func}}$ is shown in Fig. S12, where each data point is evaluated with five trials.

In its drug-dispensing mode, the soft robot's $a_{\mathrm{Disp}}$ generally has a positive correlation with $|\vec{B}_{\mathrm{Func}}|$ and it is observed the drug-dispensing mechanism achieves an opening ($a_{\mathrm{Disp}}$) of 1.2 mm when $|\vec{B}_{\mathrm{Func}}|$ is operating at its maximum capacity of 34 mT (Fig. S12**a**). By realizing such large openings, our soft robot can therefore completely empty all the drugs in its chamber, including a large drug that has a volume of 0.41 mm × 0.7 mm × 0.8 mm (equivalent to the maximum storage capacity of the drug chamber). To activate this surgical function, $|\vec{B}_{\mathrm{Func}}|$ must be greater than or equal to 4 mT (Fig. S12**a**(ii)). This is an advantageous feature because it can allow us to effectively decouple our soft robot's locomotion and drug-dispensing function. Specifically, we can command the soft robot to execute its locomotion without activating its function by restricting $|\vec{B}_{\mathrm{Func}}|$ to have an upper bound. For instance, when the soft robot is carrying a simulated drug of 0.24 mm × 0.4 mm × 0.4 mm (made of polyvinyl

chloride), we can constrain $|\vec{B}_{\text{Func}}|$ to be 3 mT or smaller so that the soft robot can execute its desired locomotion without dispensing the encased drugs. By increasing $|\vec{B}_{\text{Func}}|$ beyond 3 mT, the soft robot can also release its drugs upon command.

When the soft robot is reprogrammed to its cutting mode, its $a_{\text{Cut}}$ has a positive correlation with $|\vec{B}_{\text{Func}}|$ (Fig. S12**b**). The retractable cutting tool can be extended ($a_{\text{Cut}}$) by 464 μm when $|\vec{B}_{\text{Func}}|$ is operating at its maximum capacity of 34 mT. Similar to the drug-dispensing mode, $a_{\text{Cut}}$ will only become greater than zero when $|\vec{B}_{\text{Func}}| \geq 5$ mT (Fig. S12**b**(ii)). Otherwise, the inner beams will not be able to successfully extend the cutting tool out of the soft robot's main body. Due to this beneficial feature, we can therefore use a strong $|\vec{B}_{\text{Func}}|$ to fully extend the cutting tool of the soft robot when the robot is commanded to make deeper incisions. Concurrently, we can also limit $|\vec{B}_{\text{Func}}| < 5$ mT if we wish to move the robot while ensuring that its cutting tool is always safely stored within the body.

When the soft robot is reprogrammed to its gripping/storage function, its $a_{\text{Grip}}$ also has a positive correlation with $|\vec{B}_{\text{Func}}|$ (Fig. S12**c**). The grippers can be extended ($a_{\text{Grip}}$) by 709 μm when $|\vec{B}_{\text{Func}}|$ is operating at its maximum capacity of 34 mT. When higher $|\vec{B}_{\text{Func}}|$ are applied, the grippers can open wider, and this will potentially allow them to encompass and thus grasp larger objects. Having a large $a_{\text{Grip}}$ can also allow the soft robot's grippers to easily release the objects that they are currently grasping. In contrast, the opening distance of the grippers can be reduced by lowering $|\vec{B}_{\text{Func}}|$ and this can allow the soft robot to exert larger gripping forces on the objects that they are currently gripping so that these objects can be held more firmly. Similar to the other two function modes, we also observe that $a_{\text{Grip}}$ is zero when $|\vec{B}_{\text{Func}}| < 7$ mT (Fig. S12**c**(ii)). This implies that the gripping/storage function will only be activated when $|\vec{B}_{\text{Func}}| \geq 7$ mT. By default, the grippers are in the closed position, i.e., $a_{\text{Grip}} = 0$, and thus they will be able to hold on to an object or store an object securely while they are actuated by fields of $|\vec{B}_{\text{Func}}| < 7$ mT that allow the soft robot to generate its locomotion.

## SI Section S4. Locomotion

In this section, we report the experiments pertaining to the soft robot performing six-DOF, rigid-bodied motions (Section S4A), rolling (Section S4B) and two-anchor crawling (Section S4C).

### A. Six-DOF motions

In this subsection, we describe how the soft robot can be commanded to perform six-DOF rigid body motions. When the soft robot is in its locomotion mode, its tentacles can deform into an upright or inverted 'U'-shaped configuration upon actuation by a $\vec{B}$ along its $Z_{\text{Loco},\{L\}}$ (Fig. 2**a**-**b**). In this deformed configuration, the soft robot possesses an effective $\vec{m}_{\text{Loco}}$ that is necessary to perform rigid body motions (SI Section S2). By exploiting the phenomenon in which the soft robot's $\vec{m}_{\text{Loco}}$ tends to align with the external $\vec{B}$, we can rotate $\vec{B}$ to allow the robot to rotate about its $X_{\mathrm{Loco},\{L\}}$- and $Y_{\mathrm{Loco},\{L\}}$-axes, respectively. The frequency of the rotating $\vec{B}$ is 0.141 Hz while its magnitude is held constant at 15 mT throughout these experiments (Fig. 2**a** and SI Video S2). To rotate the robot about its $Z_{\mathrm{Loco},\{L\}}$-axis (i.e., the sixth-DOF axis), we can control the second null space component in Eq. (S2.19) such that $\theta$ is linearly increased from 0° to 360° within 3 s (Fig. 2**b** and SI Video S2). To characterize the achievable angular velocities about the $X_{\mathrm{Loco},\{L\}}$- and $Y_{\mathrm{Loco},\{L\}}$-axes of the soft robot, we applied a rotating $\vec{B}$ of 20 mT about the respective axes, varying the frequency of $\vec{B}$ ($f_{\vec{B}}$) from 0.5 Hz to 3.5 Hz at intervals of 0.5 Hz and measured the corresponding angular frequency of the robot ($f_{\mathrm{Robot}}$) (Fig. S13**a**(i)-(ii)). In these experiments, the robot's tentacles possess the inverted 'U'-shaped configuration and $k_2$ in Eq. (S2.19) is held constant at 0.4 so that the robot will not rotate about its six-DOF axis uncontrollably. The achievable angular velocities about the $Z_{\mathrm{Loco},\{L\}}$-axis of the robot are characterized by first applying a $\vec{B}$ of 15 mT such that its tentacles assume the upright 'U'-shaped configuration. The variable $k_2$ in Eq. (S2.19) is then held constant at 0.4 but the variable $\theta$ in the second null space vector of Eq. (S2.19) is linearly increased from 0° to 360° at varying frequencies ($f_{\vec{B}_{\mathrm{Grad}}}$) of 0.10 Hz to 0.30 Hz at intervals of 0.05 Hz while the corresponding $f_{\mathrm{Robot}}$ was measured. Upon receiving such actuation signals, the robot will rotate about its $Z_{\mathrm{Loco},\{L\}}$-axis (Fig. S13**a**(iii)). For these experiments, we command the robot's tentacles to assume an upright 'U'-shaped configuration because this can reduce its friction with the ground as it rotates. From our characterization experiments, we report that our soft robot is able to achieve angular velocities of 16.5 rad/s, 16.1 rad/s and 1.56 rad/s about the robot's $X_{\mathrm{Loco},\{L\}}$-, $Y_{\mathrm{Loco},\{L\}}$-, and $Z_{\mathrm{Loco},\{L\}}$-axes respectively (Fig. S13**a**). Using the pseudo-inverse component in Eq. (S2.19), we can apply appropriate magnetic spatial gradients of 0.210 T/m, 0.217 T/m, and 0.310 T/m to translate the robot along its $X_{\mathrm{Loco},\{L\}}$-, $Y_{\mathrm{Loco},\{L\}}$-, and $Z_{\mathrm{Loco},\{L\}}$-axes, respectively (Fig. 2**c** and SI Video S2). Under such actuation signals, our robot can translate at speeds of 0.126-0.911 mm/s (Fig. 2**c** and SI Video S2). Due to the limitations of our electromagnetic coil system, we have to adhesively attach a Styrofoam cube to the top of our soft robot to increase its overall buoyancy for these experiments (Fig. 2**c** and SI Video S2). In theory, our soft robot will be able to translate against gravity if our electromagnetic coil system can output 4.39 T/m and such electromagnetic coil systems are already available in the literature (*19*). The experiments in Fig. 2**a-c** and SI Video S2 imply that our soft robot possesses six-DOF in its locomotion mode.

Similar experiments have been conducted on our soft robot when it is in its function modes (Fig.s S13**b**, S14). We report that the soft robot is able to perform six-DOF rigid body motions when it is in the function modes. As an example, we have demonstrated full six-DOF motions of the robot when it is reprogrammed to its drug-dispensing mode (Fig. S14).

B. Rolling

When the robot is in its locomotion mode, it can perform various forms of rolling. The rolling gaits can be executed by first applying a $\vec{B}$ of 15 mT to deform the robot's soft tentacles into an inverted 'U'-shaped configuration so that the robot can possess an effective $\vec{m}_{\mathrm{Loco}}$ that is parallel $Z_{\mathrm{Loco},\{L\}}$-axis (Fig.s 2**a**, **d**-**f**, S7**b**(i)). By rotating $\vec{B}$ about the robot's $X_{\mathrm{Loco},\{L\}}$-axis with a constant magnitude of 15 mT, the robot is able to roll along its length (Fig. 2**a**(i), **d**(i)). Similarly, by rotating $\vec{B}$ about the robot's $Y_{\mathrm{Loco},\{L\}}$-axis at 15 mT, we can command the soft robot to roll along its width (Fig. 2**a**(ii), **d**(ii)). To steer the rolling direction, we can control the sixth-DOF angular displacement ($\theta$) about the robot's $\vec{m}_{\mathrm{Loco}}$ via the second null space component in Eq. (S2.19). The scale factor, $k_2$, for this null space vector is held constant at 0.4 for these experiments.

In theory, the soft robot's rolling speed ($v_{\mathrm{Roll}}$) can be computed via the product of its rolling circumference and frequency ($f_{\mathrm{Roll}}$) under no-slip conditions. To validate this hypothesis, we measure the soft robot's rolling speed as it rolls along the length and width in air and in an oil reservoir. During the experiments, the magnitude of $\vec{B}$ is specified to be 20 mT while $f_{\mathrm{Roll}}$ is varied from 0.1 to 1 Hz. The experimental results are plotted in Fig. S15 and each data point is measured with five trials. In general, the results indicate that $v_{\mathrm{Roll}}$ does have a linear relationship with $f_{\mathrm{Roll}}$, and the measured rolling speed along the soft robot's length is between 1.02 mm/s and 6.75 mm/s while the measured rolling speed along its width is between 0.995 mm/s and 6.21 mm/s in oil. For the experiments conducted in air, the measured rolling speed along the soft robot's length is between 1.29 mm/s and 12.7 mm/s while the measured rolling speed along its width is between 0.964 mm/s and 10.8 mm/s. We remark that the soft robot can also perform its rolling locomotion when it is in its function mode (Fig. 3**a** and SI Video S5).

C. Two-anchor crawling

To execute the two-anchor crawling locomotion, the robot's soft tentacles are first deformed into their inverted 'U'-shaped configuration via applying a $|\vec{B}|$ of 22 mT along the positive $Z_{\mathrm{Loco},\{L\}}$-direction (Fig. S16). In this configuration, the free ends of the tentacles will serve as the anchor points for the robot (Fig. S16). The two-anchor crawling locomotion can be initiated by first lifting the leading free end of the tentacles via rotating $\vec{B}$ (32° anticlockwise) about the robot's $X_{\mathrm{Loco},\{L\}}$-axis (Fig. S16**a**(ii)-(iii)). Subsequently, $|\vec{B}|$ is reduced to 8 mT to decrease the curvature of the tentacles (Fig. S16**a**(iv)). By rotating $\vec{B}$ clockwise about the robot's $X_{\mathrm{Loco},\{L\}}$-axis (32°), the leading free end of the tentacles extends forward and resumes contact with the substrate while the trailing free end is lifted (Fig. S16**a**(v)). The magnitude of $\vec{B}$ is increased back to 22 mT which causes the curvature of the legs to increase and therefore retract the trailing end. Lastly, $\vec{B}$ is returned to its original orientation until both anchor points are back in contact with the substrate (Fig. S16**a**(vi)). This sequence constitutes one cycle of two-anchor crawling for the soft robot and such cycles can be executed repeatedly upon command. Similar to the rolling locomotion, we can steer the soft robot's crawling direction by controlling the magnetic spatial gradients based on the second null space vector of Eq. (2.19). The scale factor, $k_2$, of this null space vector is specified to be 0.4 for these experiments. Additionally, we can also control the stride length of the robot in every cycle by tuning the magnitude of $\vec{B}$ (Fig. S16**a**).

The speed of the two-anchor crawling locomotion ($v_{\text{Crawl}}$) can be controlled through the frequency of each cycle ($f_{\text{Crawl}}$). Assuming no-slip conditions, the ideal $v_{\text{Crawl}}$ should be the product of $f_{\text{Crawl}}$ and the soft robot's stride length per cycle. We measured $v_{\text{Crawl}}$ as we varied $f_{\text{Crawl}}$ and the results are plotted in Fig. S16**b** and they are compared to the ideal $v_{\text{Crawl}}$, which has a stride length of 0.742 mm in our experiments. In general, our theoretical predictions agree with the experimental data because increasing $f_{\text{Crawl}}$ does result in higher $v_{\text{Crawl}}$. The deviations between the measured and ideal $v_{\text{Crawl}}$ may be attributed to slipping, which becomes more apparent at higher values of $f_{\text{Crawl}}$. For these experiments (Fig. S16**b**), $f_{\text{Crawl}}$ is varied from 0.5 to 2.5 Hz at an interval of 0.5 Hz and each trial is repeated five times. The experiments are conducted in air and in paraffin oil. The two-anchor crawling speed is measured to be between 0.4 mm/s to 1.06 mm/s in oil and between 0.322 mm/s to 0.78 mm/s in air. Unlike the gaits presented in SI Sections S4A-B, the two-anchor crawling locomotion is exclusive to the locomotion mode as it depends heavily on the deformation of the soft tentacles to be executed.

## SI Section S5. Surgical functionalities

In this section, we provide the experimental details for demonstrating the soft robot's surgical functionalities. The experiments pertaining to the robot's drug-dispensing, cutting, gripping and storage, and remote heating functions are discussed in Sections S5A-D, respectively. For these experiments, the soft robot has already been reprogrammed to its respective function mode prior to the start of the experiments.

### A. Drug-dispensing mode

The experiments shown in Fig. 3**a** and SI Video S5 are conducted by first commanding the soft robot to roll along its length via rotating $\vec{B}$ about the $X_{\text{Disp},\{L\}}$-axis with a constant magnitude of 10 mT. The rolling robot can also be steered via the magnetic control signals based on the second null space vector of Eq. (S2.19) to control the robot's sixth-DOF angular displacement so that it can produce a 90° sharp turn to reach the targeted location indicated by the red circle. Upon reaching the targeted location, the soft robot is commanded to rotate about its $X_{\text{Disp},\{L\}}$-axis so that it can angle its body to a favorable orientation to dispense its encased drug. This can be achieved by rotating $\vec{B}$ about the robot's $X_{\text{Disp},\{L\}}$-axis with a magnitude of 5 mT. Subsequently, by increasing the magnitude of $\vec{B}$ to 29 mT, the doors of the drug chamber can be activated open to release the drug at the target location. The simulated drug that we use here is a piece of polyvinyl chloride (PVC) colored in blue and it has a volume of 0.24 mm × 0.4 mm × 0.4 mm (0.0384 $mm^3$). After the drug is dispensed successfully, the soft robot rolls away from the target location. The demonstration in Fig. 3**a** and SI Video S5 show that the robot is able to successfully transport and dispense the drug in its function mode while ensuring that the drug will not be dispensed accidentally during this process. It is noteworthy that the maximum volume of drug, which our robot can dispense is 0.41 mm × 0.7 mm × 0.8 mm (0.230 $mm^3$) and this volume is equivalent to the maximum storage capacity of our drug chamber (Fig. S17**a**). Since only 0.115 $mm^3$ of drugs are generally required for targeted drug delivery treatments (*20*), this implies that the amount of drugs that our soft robot can carry and transport is twofold of what is required for practical biomedical applications. If the soft robot is required to deliver liquid drugs, we can potentially encapsulate these fluids within a biodegradable, solid capsule so that our robot can dispense it successfully too. Such biodegradable capsules have been used by Zhang et al. (*21*).

B. Cutting mode

Based on the experimental data in Fig. S12**b**(ii), we have conducted cutting experiments in air (Fig. 3**b** and SI Video S6). Unlike other experiments, the actuating magnetic fields in this experiment are generated by a pair of permanent magnets that are arranged in a Helmholtz configuration. During the experiment, a magnetic field of 29.2 mT is applied on the soft robot so that it can be actuated to extend its cutting tool out of the main body. Subsequently, the robot is made to translate towards a biological structure made of gelatin via applying a magnetic spatial gradient of 1.5 T/m. The magnetic gradient-pulling force applied on the robot (53.4 μN) allows its cutting tool to generate sufficient pressure on the gelatin structure to make a successful penetration (Fig. 3**b**(iii) and SI Video S6). By reversing the direction of the magnetic spatial gradients as well as reducing the magnitude of the applied $\vec{B}$, a magnetic force is applied on the soft robot such that it can pull away from the gelatin structure while its cutting tool is simultaneously retracted.

The utilized 10 wt% gelatin structure has similar fracture toughness to those of human tissues (*22–24*). We analyzed the cutting rate in SI Video S6 to be approximately 2 mm per minute, at which the fracture toughness of 10 wt% gelatin is estimated to be 2.5 N/m (*22*). Such fracture toughness is similar to those of the tissues in the human brain and liver (1-10 N/m) (*23, 24*). As a result, this experiment suggests that our soft robot can indeed have the potential to perform successful incision during surgery. The gelatin structure is fabricated by mixing fish gelatin powder (Bloom 200) with 95°C water. Since pristine gelatin is transparent, we add red dye to the gelatin so that we can observe the gelatin structure more clearly in the experiments. The gelatin is then allowed to set at 6.3°C for 2 h before it is cut into the desired shape and deployed for the experiment in Fig. 3**b** and SI Video S6.

C. Gripping/storage mode

For the experiment in Fig. 3**c** and SI Video S7, the robot starts within gripping distance from an object, which is created by PVC colored with red ink. We first command the robot to pick up the object via increasing $\vec{B}$ to $25$ mT and rotating the field about the robot's $X_{\text{Grip},\{L\}}$-axis such that the grippers can encompass the object before decreasing $\vec{B}$ to $10$ mT to grab the object tightly. While the object is gripped, the soft robot can execute a spin-walking locomotion to move towards the targeted location (green circle). The spin-walking locomotion is executed by rotating the soft robot about its $X_{\text{Grip},\{L\}}$-axis and $Y_{\text{Grip},\{L\}}$-axis repeatedly to alternate the anchoring point of the robot. Although spin-walking is a slower gait than rolling, this locomotion allows the soft robot to both maintain an upright orientation during locomotion while also enabling the robot to be positioned more accurately. We also remark that the spin-walking locomotion can be utilized by the soft robot in any of its three function modes and it is not specific to just the gripping mode. At the targeted location, we increase the magnitude of $\vec{B}$ to $25$ mT so that the soft robot can release the object and place it at this targeted location before moving away.

In the experiment shown in Fig. 3**d** and SI Video S7, the soft robot is initially in the gripping mode. It is then commanded to pick up the object (PVC colored red with ink) by increasing $|\vec{B}|$ to $25$ mT so that the opening of its grippers can encompass the object. The magnitude of $\vec{B}$ is then decreased to $10$ mT so that the grippers of the soft robot can grab the object more firmly. Subsequently, the soft robot is commanded to rotate until its grippers are pointing against gravity. This can be achieved via rotating $\vec{B}$ about the soft robot's $X_{\text{Grip},\{L\}}$-axis while maintaining its magnitude at $10$ mT. Once the soft robot has reached its desired orientation, the magnitude of $\vec{B}$ is increased to $29$ mT so that the grippers can release the object and allow it to fall into the robot's storage chamber via gravity. By decreasing $|\vec{B}|$ to $10$ mT, and the soft robot can reduce the opening of its grippers before it returns to its upright position via rotating $\vec{B}$ about its $X_{\text{Grip},\{L\}}$-axis. The reprogrammable module is then demagnetized so that the soft robot can securely store the object while it executes the two-anchor crawling gait in the locomotion mode. The size of the stored object is $0.4$ mm $\times$ $0.4$ mm $\times$ $0.4$ mm (volume of $0.064$ $mm^3$) and this is $1.78$-fold of what is required for deoxyribonucleic acid (DNA) extraction and polymerase chain reaction (PCR) amplification. Such biopsy applications generally require $0.0359$ $mm^3$ volume of samples (*25*).

D. Remote heating function

The experiment illustrated in Fig. 3**e** and SI Video S8 is conducted in air with the robot reprogrammed to its locomotion mode. A layer of thermochromic dye (Narika P70-2637) is painted on the remote heating component of the soft robot for this experiment. The attached dye is designed to be blue when its temperature is below 40°C, and pink when its temperature is 40°C or higher. At the start of the experiment, the thermochromic dye is blue because the temperature of the soft robot and the ambient temperature is 26.3°C (Fig. 3**e** and SI Video S8). By applying an alternating magnetic field of 9.34 mT and 75.4 kHz for one minute, the remote heating component of the robot triggers the thermochromic dye to change its color to pink after 35 s, indicating that that the temperature of the dye has already increased to at least 40°C (Fig. 3**e** and SI Video S8). It is also observed that after the remote heating component is allowed to cool for 30 s, the thermochromic dye changes its color back to blue. The soft robot remains stationary during the remote heating experiment, and this implies that we can effectively decouple the remote heating function and locomotion of the soft robot. We observe similar outcomes when this experiment is repeated while the soft robot is in its function modes (Fig. S17**b**). Based on the experimental results, and specific heat capacity (661 J/(kg·K)) and mass (5.26 mg) of the remote heating component, it is estimated that our soft robot can produce heat at a mean rate of 1.36 mW since the remote heating component can raise its temperature to 40°C from 26.3°C within 35 s. Because our computations have ignored the effects of heat transfer from the soft robot to the surroundings, the actual heat generation of our robot should be higher than 1.36 mW. This is theoretically sufficient for our robot to perform a wide range of hyperthermia treatments, which involve raising tissue temperatures to a minimum of 40°C, as a method to kill malignant tumor cells or in conjunction with radiotherapy or chemotherapy (*26–28*). While our remote heating is conducted with alternating magnetic fields of 75.4 kHz, it may be possible to increase the frequency of these magnetic fields to 1 MHz in the future. This can potentially increase the heating power and make hyperthermia treatments more effective.

**SI Section S6. Experiments conducted on biological phantoms**

In this section, we will provide more information details for the experiments shown in Fig. 4**a** and SI Video S9 (Section S6A), as well as those in Fig. 4**b** and SI Video S10 (Section S6B). The utilized biological phantoms in these experiments are made of PDMS that is colored in red, and these phantoms are created via molding techniques.

A. Locomotion and four reprogrammable surgical functionalities

In Fig. 4**a** and SI Video S9, we show that our soft robot can be reprogrammed to execute a combination of its surgical functionalities on a biological phantom. To create a synthetic tumor in the environment, we attach a 0.064 $mm^3$ cube of red-colored polyvinyl chloride (PVC) to the phantom's substrate via a piece of molded butter. Because our electromagnetic coil system in Fig. S6**a** is unable to generate sufficient magnetic spatial gradients for our soft robot to cut through 10 wt% gelatin, butter is used as a medium between the substrate and the synthetic tumor in this proof-of-concept demonstration as this material is softer and easier to cut. A blue PVC cuboid of 0.0384 $mm^3$ is also placed in the soft robot's main body to represent the drug, which is going to be delivered at the targeted location (marked with a black dotted-line circle in Fig. 4**a**(i)). The soft robot begins the experiment in its locomotion mode. By rotating $\vec{B}$ of 15 mT about the robot's $X_{\mathrm{Loco},\{L\}}$-axis, the soft robot can roll along its length to the targeted location for drug delivery (Fig. 4**a**(ii)). Upon reaching the location, the soft robot is reprogrammed to its drug-dispensing mode. Because the soft robot's $\vec{m}_{\mathrm{Disp}}$ will tend to align with $\vec{B}$, we can control the direction of $\vec{B}$ to allow the robot to adopt an orientation favorable to dispense its encased drug. Subsequently, the magnitude of $\vec{B}$ is increased to 25 mT so that the soft robot can create an opening to release the encased drug (Fig. 4**a**(iii)). After the drug is dispensed, the soft robot is reprogrammed to its locomotion mode via the demagnetizing field specified in Eq. (S1.3). Under its locomotion mode, the soft robot executes its two-anchor crawling locomotion to move towards the synthetic tumor (Fig. 4**a**(iv)). At close proximity to the tumor, the soft robot is then reprogrammed to its cutting mode. By applying a $|\vec{B}|$ of 20 mT, the cutting tool is extended. The soft robot is then commanded to perform a rotation that allows it to slice though the butter, detaching the synthetic tumor from the substrate (Fig. 4**a**(v)). The soft robot is then reprogrammed to its locomotion mode via the demagnetized field specified in Eq. (S1.3). To grip and eventually encase the excised tumor, the soft robot is reprogrammed to its gripping mode. In this configuration, we control the soft robot's orientation via varying the direction of $\vec{B}$ so that the robot can adopt a favorable orientation to pick up the excised tumor. By increasing $|\vec{B}|$ to 26 mT, the grippers of the soft robot can be extended out of the robot's main body. Once the grippers are in place to grasp the excised tumor, $|\vec{B}|$ is reduced to 10 mT so that they can grab the tumor and pick it up (Fig. 4**a**(vi)). The soft robot is then commanded to tilt back such that it can exploit gravity to allow the excised tumor to fall into its storage chamber (Fig. 4**a**(vii)). By reprogramming the soft robot back to its locomotion mode, the robot can execute the two-anchor crawling locomotion to exit the phantom while ensuring that the excised tumor can be stored safely within its main body (Fig. 4**a**(viii)).

B. Locomotion and remote heating

To exhibit our soft robot's remote heating functionality on another biological phantom, we place a layer of the thermochromic dye (Narika P70-2637) at a targeted location of the phantom. Similar to the experiments in Fig. 3**e** and SI Video S8, the thermochromic dye is designed to be blue when its temperature is below 40°C, and pink when its temperature is 40°C or higher. The experiment in Fig. 4**b** and SI Video S10 is conducted in air, and the dye is blue at the start of the experiment as the initial ambient temperature is 25°C. In its locomotion mode, we rotate $\vec{B}$ of 15 mT about the soft robot's $X_{\mathrm{Loco},\{L\}}$-axis such that the robot will roll towards the targeted locomotion and eventually allow its remote heating component to establish contact with the thermochromic dye (Fig. 4**b**(ii)-(iii) and SI Video S10). The soft robot is then subjected to an alternating magnetic field of 9.34 mT and 75.4 kHz for 15 minutes before the dye changes its color to pink, indicating that its temperature is raised to at least 40°C (Fig. 4**b**(iii)-(iv) and SI Video S10).

**SI Section S7. Additional discussion**

In this section, we will discuss about the safety aspects of our actuating magnetic fields when the proposed soft robot is deployed for its potential medical treatments (Section S7A). A scaling analysis of our soft robot will be discussed in Section S7B too.

A. Safety aspects of the applied magnetic fields

In general, if patients are exposed to magnetic fields that are of too high frequency and magnitude, they may experience cardiac fibrillation due to nerve stimulation and their body temperature may be elevated to dangerous levels. To examine if the actuating magnetic fields of our soft robot are safe for medical treatments, we have first classified these actuating signals into four categories: (I) for locomotion; (II) for activating the surgical functionalities; (III) for reprogramming the soft robot to different operating modes; (IV) and for heating remotely. The actuating signals of these four categories are denoted with their roman numerals in the subscript and we will determine if these actuating signals will theoretically cause nerve stimulation and overheating for the patients by evaluating their temporal rate of change. In general, it is ideal to minimize the actuating magnetic fields' temporal rate of change because such fields will be safer for the patients.

Of the magnetic fields that are used for the soft robot's locomotion, those that command the robot to rotate have the highest angular frequency ($f_{\mathrm{Rotate}} = 3$ Hz) and greatest magnitude of $B_{\mathrm{Rotate}}$ (15 mT). Therefore, such magnetic fields have the highest temporal rate of change ($\frac{\mathrm{d}|\vec{B}|}{\mathrm{d}t}$) for category I and they can be expressed as:

$$|\vec{B}_{\mathrm{I}}| = B_{\mathrm{Rotate}} \cos(2\pi f_{\mathrm{Rotate}} t)\,, \tag{S7.1}$$

where $t$ represents time. For category II, the desired magnetic fields can be represented as a step response along a specified direction. Because our electromagnetic coil system generates its magnetic fields with a finite transient response, the actual magnetic fields that are output by this system can be expressed as:

$$|\vec{B}_{\mathrm{II}}| = B_{\mathrm{Act}} \left(1 - e^{-\frac{R_{\mathrm{EC}}}{L_{\mathrm{EC}}} t}\right). \tag{S7.2}$$

The variables $L_{\mathrm{EC}}$ and $R_{\mathrm{EC}}$ correspondingly represent the electrical inductance and resistance of our electromagnetic coil system, while $B_{\mathrm{Act}}$ represents the magnitude of the step response. The corresponding values of $B_{\mathrm{Act}}$, $L_{\mathrm{EC}}$ and $R_{\mathrm{EC}}$ are 34 mT, $6.3 \times 10^{-4}$ H and 0.79 Ω.

The reprogramming magnetic fields in category III can be sub-divided into its magnetizing fields and demagnetizing fields. It is possible to magnetize the soft robot via either a step or a ramp input. By accounting for the transient effects of the step input, the actual magnetizing fields can be computed as

$$\left|\vec{B}_{\mathrm{IIIa,step}}\right| = B_{\mathrm{Mag}}\left(1 - e^{-\frac{R_{\mathrm{RC}}}{L_{\mathrm{RC}}}t}\right), \tag{S7.3}$$

where $L_{\mathrm{RC}}$ and $R_{\mathrm{RC}}$ represents the inductance and resistance of the reprogramming coil, respectively. The variable, $B_{\mathrm{Mag}}$, represents the magnitude of the step function in Eq. (S7.3). The values of $B_{\mathrm{Mag}}$, $L_{\mathrm{RC}}$ and $R_{\mathrm{RC}}$ in Eq. (S7.3) are 60 mT, $1.9 \times 10^{-4}$ H and 2.5 Ω, respectively.

Alternatively, the soft robot can be magnetized via a ramp input:

$$\left|\vec{B}_{\mathrm{IIIa,ramp}}\right| = Kt, \tag{S7.4}$$

where $K$ represents the gradient of the ramp and it has a value of 12 T/s. At $t = 5$ ms, $|\vec{B}_{\mathrm{IIIa,ramp}}|$ will reach 60 mT. While the demagnetizing field has been specified in Eq. (S1.3), here we restate the demagnetizing field for the convenience of the reader:

$$\left|\vec{B}_{\mathrm{IIIb}}\right| = \left(B_{\mathrm{Demag}} - K_{\mathrm{Demag}} f_{\mathrm{Demag}} t\right)\cos\left(2\pi f_{\mathrm{Demag}} t\right), \tag{S7.5}$$

where the maximum amplitude of the demagnetizing field, $B_{\mathrm{Demag}}$, is 65 mT and the decaying magnitude, $K_{\mathrm{Demag}}$, is 2 mT per period. For our experiments, the frequency of the demagnetizing field, $f_{\mathrm{Demag}}$, is 45 Hz. Lastly, the magnetic field of category IV can be expressed as:

$$|\vec{B}_{\mathrm{IV}}| = B_{\mathrm{Heat}} \sin(2\pi f_{\mathrm{Heat}} t), \tag{S7.6}$$

where $B_{\mathrm{Heat}}$ and $f_{\mathrm{Heat}}$ represent the amplitude and frequency of the alternating magnetic field that is required for executing the soft robot's remote heating function. The values of $B_{\mathrm{Heat}}$ and $f_{\mathrm{Heat}}$ in Eq. (S7.6) are 9.34 mT and 75.4 Hz, respectively.

To prevent nerve stimulation for the patients, it is recommended that the temporal rate of change of the magnetic field is below the following upper bound threshold ($\frac{\mathrm{d}|\vec{B}|}{\mathrm{d}t}_{\mathrm{Max}}$) (*29*):

$$\frac{\mathrm{d}|\vec{B}|}{\mathrm{d}t}_{\mathrm{Max}} = 54\left(1 + \frac{0.138}{\eta}\right), \tag{S7.7}$$

where $\eta$ represents the duration of a monotonic increasing or decreasing gradient and this variable is expressed in the units of milliseconds. The criterion in Eq. (S7.7) is based on the requirements specified in general MRI procedures (*29*).

Using Eq. (S7.7), we calculate the respective $\frac{\mathrm{d}|\vec{B}|}{\mathrm{d}t}_{\mathrm{Max}}$ for each category of the actuating fields. The value of $\eta$ is based on a quarter of their period for the harmonic fields. For the step fields, the value of $\eta$ is selected based on the 2% settling time of output response. Specifically, the value of $\eta$ is 3.19 ms and 83.3 ms for categories I and II, respectively; for category IIIa, the value of $\eta$ is 0.307 ms and 5 ms for the step input and ramp input, respectively; for category IIIb, the value of $\eta$ is 5.55 ms; for category IV, $\eta$ is $3.30 \times 10^{-3}$ ms. We compute the highest $\frac{\mathrm{d}|\vec{B}|}{\mathrm{d}t}$ of each actuating signal in Eq.s (S7.1-6) and denote them as $\frac{\mathrm{d}|\vec{B}|}{\mathrm{d}t}_{\mathrm{Highest}}$. The values of each $\frac{\mathrm{d}|\vec{B}|}{\mathrm{d}t}_{\mathrm{Highest}}$ are compared against their respective $\frac{\mathrm{d}|\vec{B}|}{\mathrm{d}t}_{\mathrm{Max}}$ in Table S3. Except for the step magnetizing field in Eq. (S7.3), our actuating magnetic fields are theoretically within the safety limits of not inducing nerve stimulation. Although we have used the step magnetizing fields for most of our experiments, we report that our proposed soft robot can be also reprogrammed successfully with the safer ramp magnetizing field that is specified in Eq. (S7.4). Using the ramp magnetizing fields, our soft robot can be reprogrammed to its locomotion mode and all of its function modes successfully without producing any undesirable motions. These outcomes are therefore similar to those of the step magnetizing fields. We have used the step magnetizing fields in Eq. (S7.3) for most of our experiments in this first study because they are relatively easier to implement. There is, however, no loss in generality to apply the safer ramp magnetizing fields in Eq. (S7.4) when our soft robot is deployed for potential treatments in the future.

To prevent overheating for the human body, it is recommended to keep the frequencies ($f$) and amplitudes of the magnetic fields ($|\vec{H}|$) within the range 0.05-1.2 MHz and 0-15 kA/m, respectively (*30–32*). In general, the heating power of the magnetic fields can be quantified using the product of $|\vec{H}|$ and $f$, and the magnetic fields are considered safe if $|\vec{H}|f < 9.46 \times 10^{9}$ A/(m·s) (*32*). Using this threshold as a guideline, we have calculated the highest $|\vec{H}|f$ (denoted as $|\vec{H}|f_{\mathrm{Highest}}$) for all the categories of our magnetic fields (Table S3). Since all of our actuating fields have $|\vec{H}|f_{\mathrm{Highest}}$ of $5.57 \times 10^{8}$ A/(m·s) or lower, they are therefore within the safety threshold, suggesting that our actuating magnetic fields will not overheat the human body. It is noteworthy that the induced magnetic field $\vec{B}$ is equal to $\mu\vec{H}$ where $\mu$ represents the magnetic permeability of biological tissues ($\mu = 4\pi \times 10^{-7}$ H/m) (*33*).

## B. Scaling analysis

Here we discuss the effects of our soft robot when the main body of our robot is shrunk from 2.5 mm to 2.07 mm and the length of its soft tentacles are scaled down from 4.4 mm to 2.5 mm (Fig. S18). In the subsequent analysis, we will determine the required changes for our scaled-down robot such that it can theoretically still accomplish all the necessary surgical functionalities and locomotion while using the same types of $\vec{B}$ that are applied on our current robot. To facilitate this discussion, we have made several assumptions: (i) the materials of the soft robot remain the same; (ii) the rigid components of the soft robot will be scaled down isotropically, (iii) while the parameters of its soft components and attached cutting tool are allowed to be adjusted accordingly as long as such adjustments are within manufacturing constraints.

In general, the magnetic torque ($|\vec{T}|$) and force ($|\vec{F}|$) that our soft robot can generate are linearly proportional to the volume of its magnetized components:

$$|\vec{T}| \propto {l_\mathrm{c}}^3, \qquad \text{(S7.8A)}$$

$$|\vec{F}| \propto {l_\mathrm{c}}^3, \qquad \text{(S7.8B)}$$

where $l_\mathrm{c}$ is the soft robot's characteristic length. To facilitate our discussion, we introduce a scale factor, $\lambda = \frac{l_\mathrm{c,n}}{l_\mathrm{c,o}}$, where the additional subscripts 'o' and 'n' denote the $l_\mathrm{c}$ of the current and scaled down soft robot, respectively. Similar notations will also be applied to other parameters of our soft robot. Since the volume of our soft robot's magnetized rigid components will be scaled down by a factor of ${\lambda_\mathrm{Body}}^3$ (i.e., ${\lambda_\mathrm{Body}}^3 = \left(\frac{2.07\ \mathrm{mm}}{2.50\ \mathrm{mm}}\right)^3 = 0.568$), Eq. (S7.8) implies that its $|\vec{T}|$ and $|\vec{F}|$ will only be $\sim 56.8\%$ of its original value, i.e., $|\vec{T}|_\mathrm{n} = {\lambda_\mathrm{Body}}^3 |\vec{T}|_\mathrm{o}$ and $|\vec{F}|_\mathrm{n} = {\lambda_\mathrm{Body}}^3 |\vec{F}|_\mathrm{o}$. To ensure that the scaled-down robot is still able to produce sufficient deformation to execute its surgical functions, the stiffness of its soft passive components must therefore also be reduced to $\sim 56.8\%$ of its original value. In general, the soft passive components of our robot can be modelled as uniform or non-uniform beams and their stiffness can be reduced by varying their beam thickness. To analyze the scaling effects on the deformation of the inner beams that are in our soft robot's main body, we first expand and arrange Eq. (S2.13) to establish the following relationship:

$$\gamma_\mathrm{Intip} = \frac{12|\vec{T}|l_\mathrm{Inner}}{Ebh^3}, \qquad \text{(S7.9)}$$

where $E$ is the Young's modulus of the inner beam, while $b$ and $h$ represent the beam's width and thickness, respectively. Equation (S7.9) is valid for both the current and scaled-down robots. Specifically, Eq. (S7.9) can be used to describe the current soft robot's deformation as:

$$\gamma_\mathrm{Intip,o} = \frac{12|\vec{T}|_\mathrm{o} l_\mathrm{Inner,o}}{E b_\mathrm{o} {h_\mathrm{o}}^3}. \qquad \text{(S7.10)}$$

Likewise, Eq. (S7.9) can also be applied on the scaled-down robot:

$$\gamma_{\text{Intip,n}} = \frac{12\left|\vec{T}\right|_{\text{n}} l_{\text{Inner,n}}}{E b_{\text{n}} {h_{\text{n}}}^3}. \tag{S7.11}$$

By dividing Eq. (S7.11) over Eq. (S7.10), the following scaling relationship can be established:

$$\frac{\gamma_{\text{Intip,n}}}{\gamma_{\text{Intip,o}}} = \left(\frac{\left|\vec{T}\right|_{\text{n}}}{\left|\vec{T}\right|_{\text{o}}}\right)\left(\frac{l_{\text{Inner,n}}}{l_{\text{Inner,o}}}\right)\left(\frac{b_{\text{o}}}{b_{\text{n}}}\right)\left(\frac{{h_{\text{o}}}^3}{{h_{\text{n}}}^3}\right). \tag{S7.12}$$

For the scaled-down robot to achieve the same level of deformation as those of our current robot, $\gamma_{\text{Intip,n}}$ must be equal to $\gamma_{\text{Intip,o}}$. Based on this requirement, Eq. (S7.12) can be rearranged into the following form if we also assume that $\frac{b_{\text{o}}}{b_{\text{n}}}$ is equal to $\frac{1}{\lambda_{\text{Body}}}$:

$${h_{\text{n}}}^3 = \left({\lambda_{\text{Body}}}^3\right)\left(\lambda_{\text{Body}}\right)\left(\frac{1}{\lambda_{\text{Body}}}\right)\left({h_{\text{o}}}^3\right)$$

$$\rightarrow h_{\text{n}} = \lambda_{\text{Body}} h_{\text{o}}. \tag{S7.13}$$

Equation (S7.13) states that the scaled-down robot can generate deformations that are similar to those of our current robot if its beam thickness in the main body can be reduced by a factor of $\lambda_{\text{Body}}$, i.e., $\lambda_{\text{Body}} = 0.828$. Because all of our current robot's beam thickness in the main body are in the range of 20-60 μm, it will be possible to use photolithography and 3D-printing methods to manufacture such thinner beams for the scaled-down robot, i.e., their beam thickness can be fabricated to be in the range of 16.6-49.8 μm. Hence, it is indeed feasible to manufacture a scaled-down robot, which is sufficiently compliant to realize the desired surgical functionalities that are similar to those presented by our current robot.

A similar scaling analysis can be applied on the soft robot's tentacles. Specifically, we can first approximate Eq. (S2.9) as follows:

$$M_{\text{Tent(soft)}}\left|\vec{B}\right| bh \approx E\frac{bh^3}{12}\frac{\Delta(\Delta\gamma_{\text{Tent}})}{{l_{\text{c}}}^2},$$

$$\rightarrow M_{\text{Tent(soft)}}\left|\vec{B}\right| = E\frac{h^2}{12}\frac{\Delta(\Delta\gamma_{\text{Tent}})}{{l_{\text{c}}}^2}, \tag{S7.14}$$

where $E$ is the Young's modulus of the soft tentacles, while $b$ and $h$ represent the soft tentacle's width and thickness, respectively. Equation (S7.14) is valid for our current robot as well as for its scaled-down version, i.e., we can establish the following equations:

$$M_{\text{Tent(soft),n}}\left|\vec{B}_{\text{n}}\right| = E\frac{{h_{\text{n}}}^2}{12}\frac{\Delta\left(\Delta\gamma_{\text{Tent,n}}\right)}{{l_{\text{c,n}}}^2}, \tag{S7.15A}$$

$$M_{\mathrm{Tent(soft),o}}\left|\vec{B}_{\mathrm{o}}\right| = E\frac{{h_{\mathrm{o}}}^{2}}{12}\frac{\Delta\left(\Delta\gamma_{\mathrm{Tent,o}}\right)}{{l_{\mathrm{c,o}}}^{2}}. \quad (S7.15B)$$

By dividing Eq. (S7.15A) over Eq. (S7.15B), the following scaling relationship can be established:

$$1 = \frac{{h_{\mathrm{n}}}^{2}}{{h_{\mathrm{o}}}^{2}}\frac{{l_{\mathrm{c,o}}}^{2}}{{l_{\mathrm{c,n}}}^{2}}. \quad (S7.16)$$

The relationship in Eq. (S7.16) is established by assuming that the soft tentacles of the scaled-down robot can achieve the same level of deformation as those of our current robot, i.e., $\Delta\left(\Delta\gamma_{\mathrm{Tent,n}}\right) = \Delta\left(\Delta\gamma_{\mathrm{Tent,o}}\right)$. We also assume that the magnetization magnitude $\left(M_{\mathrm{Tent(soft)}}\right)$ and the strength of the actuating magnetic field $\left(\left|\vec{B}\right|\right)$ are equivalent for the scaled-down robot and those of our current robot. Furthermore, because the soft tentacle's length $(l_{\mathrm{c}})$ is decreased by a factor of $\lambda_{\mathrm{Tent}}$ (i.e., $\lambda_{\mathrm{Tent}} = \frac{2.5\ \mathrm{mm}}{4.4\ \mathrm{mm}} = 0.568$) for the scaled-down robot, Eq. (S7.16) implies that the following relationship between $h_{\mathrm{n}}$ and $h_{\mathrm{o}}$ must hold:

$$h_{\mathrm{n}} = \lambda_{\mathrm{Tent}} h_{\mathrm{o}}. \quad (S7.17)$$

Equation (S7.17) states that the scaled-down robot can generate similar deformations to those of our current robot if the thickness of its soft tentacles can be reduced by a scale factor of $\lambda_{\mathrm{Tent}}$. Because the thickness of our current robot's tentacles is 150 μm, it is possible to use photolithography methods to mold such thinner tentacles for the scaled-down robot, i.e., the thickness of the tentacles can be fabricated as 85.2 μm (*34*, *35*). Using such thinner tentacles, the scaled-down robot can potentially enable identical locomotion as those presented by our current robot.

While the volume of the scaled-down robot is 0.568-fold of our current robot (due to the scaling of ${\lambda_{\mathrm{Body}}}^{3}$), such a robot can theoretically still carry 0.131 mm$^3$ of drugs and such cargo storage is generally sufficient for targeted drug delivery applications, which require 0.115 mm$^3$ of drugs (*20*). Similarly, while the scaled-down robot can theoretically only carry 0.0363 mm$^3$ of excised biological tissues, such volumes of samples are also generally sufficient for biopsy purposes, which require 0.0359 mm$^3$ of tissues (*25*). For the scaled-down robot to perform incision successfully, its cutting tool must also be able to generate the same magnitude of pressure as those of our current robot. Since pressure is linearly proportional to the magnetic force, $\left|\vec{F}\right|$, and inversely proportional to the surface area of the cutting tool, when the smaller robot's $\left|\vec{F}\right|$ is scaled down by a factor of ${\lambda_{\mathrm{Body}}}^{3}$ (0.568), its cutting tool's surface area must also be reduced by ${\lambda_{\mathrm{Body}}}^{3}$ for it to generate the same pressure. As our current cutting tool's tip surface area is estimated to be $1.47 \times 10^{-4}$ mm$^2$, the scaled-down cutting tool should have a surface area of $8.34 \times 10^{-5}$ mm$^2$ to ensure that incision can be executed successfully. In general, there exist microcutting or micromilling tools that have diameters of 5 μm and their estimated area will be $7.85 \times 10^{-5}$ mm$^2$ (*36*), which is smaller than $8.34 \times 10^{-5}$ mm$^2$ and this satisfies our scaling requirement. Therefore, this suggests that it is indeed feasible to equip the scaled-down robot with an appropriate cutting tool, which can still potentially enable it to perform successful incision in a surgery.

To evaluate if the scaled-down robot can also perform effective hyperthermia treatments, we first state that the heat power that can be generated by our current soft robot, $\dot{Q}_{\mathrm{Gen}}$, is linearly proportionally to the volume of its remote heating component:

$$\dot{Q}_{\mathrm{Gen}} \propto {l_{\mathrm{c}}}^{3}. \tag{S7.18}$$

Likewise, we also state that soft robot's heat transfer rate to the surroundings, $\dot{Q}_{\mathrm{Trf}}$, is linearly proportional to the surface area of the remote heating component:

$$\dot{Q}_{\mathrm{Trf}} \propto {l_{\mathrm{c}}}^{2}. \tag{S7.19}$$

By dividing Eq. (S7.18) with Eq. (S7.19), the following scaling relationship for the soft robot's remote heating function can be established:

$$\frac{\dot{Q}_{\mathrm{Gen}}}{\dot{Q}_{\mathrm{Trf}}} \propto l_{\mathrm{c}}. \tag{S7.20}$$

Equation (S7.20) implies that when $l_{\mathrm{c}}$ is reduced, $\dot{Q}_{\mathrm{Trf}}$ will dominate over $\dot{Q}_{\mathrm{Gen}}$. This suggests that the soft robot will be more efficient in transferring heat energy to its surrounding when it becomes smaller. Thus, the scaled-down robot should theoretically be able to deliver hyperthermia treatments that are more effective than those of our current robot.

The analyses from Eq.s (S7.8-20) therefore conclude that it is indeed feasible to create a smaller version of our soft robot, which is able to execute all of its desired locomotion and functionalities successfully. Because our analysis in this sub-section is based on the actuating magnetic fields, which are specified in SI Section S7A to be safe for humans, the scaled-down robot will have great potential to be compatible for future medical treatments too.

**Supporting Tables**

**Table S1. Chemical, mechanical, and magnetic properties for different components of the soft robot.**

| Component | Matrix | Powder | Matrix:Powder (mass ratio) | Cure time [h] | $E$ [kPa] | $H_{ci}$ [mT] | Magnetized $M$ [kA/m] | Demagnetized $M$ [kA/m] |
|---|---|---|---|---|---|---|---|---|
| Remote heating component | PDMS | Fe3O4 | 2:3 | 12 | - | - | 6.52 | 0.766 |
| Main body (non-magnetic) | PDMS | - | - | 0.5 | 570 | - | - | - |
| Main body (magnetic) | PDMS | NdFeB | 1:4 | 0.5 | - | 598 | 108 | - |
| Reprogrammable module | PDMS | AlNiCo | 1:3 | 0.5 | - | - | 7.18 | 1.19 |
| Sixth-DOF enhancement module | PDMS | NdFeB | 1:3 | 0.5 | - | 614 | 88.7 | - |
| Soft tentacles | Ecoflex | NdFeB | 1:3 | 1 | 396 | 93.3 | 37.5 | - |

**Table S2. Physical definitions and numerical values of fixed variables required for describing the magnetization profile of the soft robot.**

| Variable | Physical definition | Numerical value |
|---|---|---|
| **From soft tentacles** | | |
| $Z_{\mathrm{Tent(soft)},\{M\}}$ | Displacement of the tentacles' top edge (along the $Z_{\{M\}}$-axis) | -1.055 mm |
| $t_{\mathrm{Tent(soft)},\{M\}}$ | Thickness of the soft tentacles | 0.15 mm |
| $b_{\mathrm{Tent}}$ | Width of the soft tentacles | 1.5 mm |
| $l_{\mathrm{Tent}}$ | Length of the soft tentacles | 4.4 mm |
| **From sixth-DOF enhancement module** | | |
| $Z_{\mathrm{Tent(rigid)},\{M\}}$ | Displacement of the centroids (along the $Z_{\{M\}}$-axis) | -0.905 mm |
| $Y_{\mathrm{Tent(rigid)},1,\{M\}}$ | Displacement of the left halve centroid (along the $Y_{\{M\}}$-axis) | -0.531 mm |
| $Y_{\mathrm{Tent(rigid)},2,\{M\}}$ | Displacement of the right halve centroid (along the $Y_{\{M\}}$-axis) | 0.531 mm |
| $V_{\mathrm{Six}}$ | Volume of the sixth-DOF enhancement module | 0.491 $mm^3$ |
| **From the main body** | | |
| $Z_{\mathrm{Main},\{M\}}$ | Displacement of the centroids (along the $Z_{\{M\}}$-axis) | 0.175 mm |
| $X_{\mathrm{Inner},1,\{M\}}$ | Displacement of the first inner magnet centroid (along the $X_{\{M\}}$-axis) | -0.52 mm |
| $X_{\mathrm{Inner},2,\{M\}}$ | Displacement of the second inner magnet centroid (along the $X_{\{M\}}$-axis) | 0.52 mm |
| $X_{\mathrm{Inner},3,\{M\}}$ | Displacement of the third inner magnet centroid (along the $X_{\{M\}}$-axis) | 0 mm |
| $Y_{\mathrm{Inner},1,\{M\}}$ | Displacement of the first inner magnet centroid (along the $Y_{\{M\}}$-axis) | 0.3 mm |
| $Y_{\mathrm{Inner},2,\{M\}}$ | Displacement of the second inner magnet centroid (along the $Y_{\{M\}}$-axis) | 0.3 mm |
| $Y_{\mathrm{Inner},3,\{M\}}$ | Displacement of the third inner magnet centroid (along the $Y_{\{M\}}$-axis) | -0.6 mm |
| $V_{\mathrm{Inner}}$ | Volume of an inner magnet | 0.0208 $mm^3$ |
| **From the reprogrammable module** | | |
| $Z_{\mathrm{Rprog},\{M\}}$ | Displacement of the centroid (along the $Z_{\{M\}}$-axis) | -0.59 mm |
| $V_{\mathrm{Rprog}}$ | Volume of the reprogrammable module | 2.6 $mm^3$ |
| **From the remote heating component** | | |
| $Z_{\mathrm{Heat},\{M\}}$ | Displacement of the centroid (along the $Z_{\{M\}}$-axis) | 0.94 mm |
| $V_{\mathrm{Heat}}$ | Volume of the remote heating component | 2.6 $mm^3$ |

**Table S3. Temporal rates of change of magnetic field and $|\vec{H}|f$ values for different categories of our applied magnetic fields. The temporal rate of change of the magnetic field that is above the threshold is marked with an asterisk.**

<table>
<tr><th></th><th></th><th></th><th colspan="2">$\frac{\mathrm{d}|\vec{B}|}{\mathrm{d}t}$ [T/s]</th><th colspan="2">$|\vec{H}|f$ [A/(m·s)]</th></tr>
<tr><th>Category</th><th>Action</th><th>Type</th><th>Maximum allowable</th><th>Highest</th><th>Maximum allowable</th><th>Highest</th></tr>
<tr><td>I</td><td>Actuating a function</td><td>Step input</td><td>56.3</td><td>10.7</td><td rowspan="6">$9.46\times10^{9}$</td><td>1.7</td></tr>
<tr><td>II</td><td>Locomotion</td><td>Harmonic</td><td>54.1</td><td>0.283</td><td>0.0358</td></tr>
<tr><td rowspan="2">IIIa</td><td rowspan="2">Reprogramming – magnetization</td><td>Step input</td><td>78.3</td><td>196*</td><td>31.2</td></tr>
<tr><td>Ramp</td><td>55.5</td><td>12</td><td>9.55</td></tr>
<tr><td>IIIb</td><td>Reprogramming – demagnetization</td><td>Harmonic (45 Hz)</td><td>55.3</td><td>18</td><td>2.33</td></tr>
<tr><td>IV</td><td>Remote heating</td><td>High frequency harmonic</td><td>2290</td><td>$2.7\times10^{-13}$</td><td>557</td></tr>
</table>

## Supporting Figures

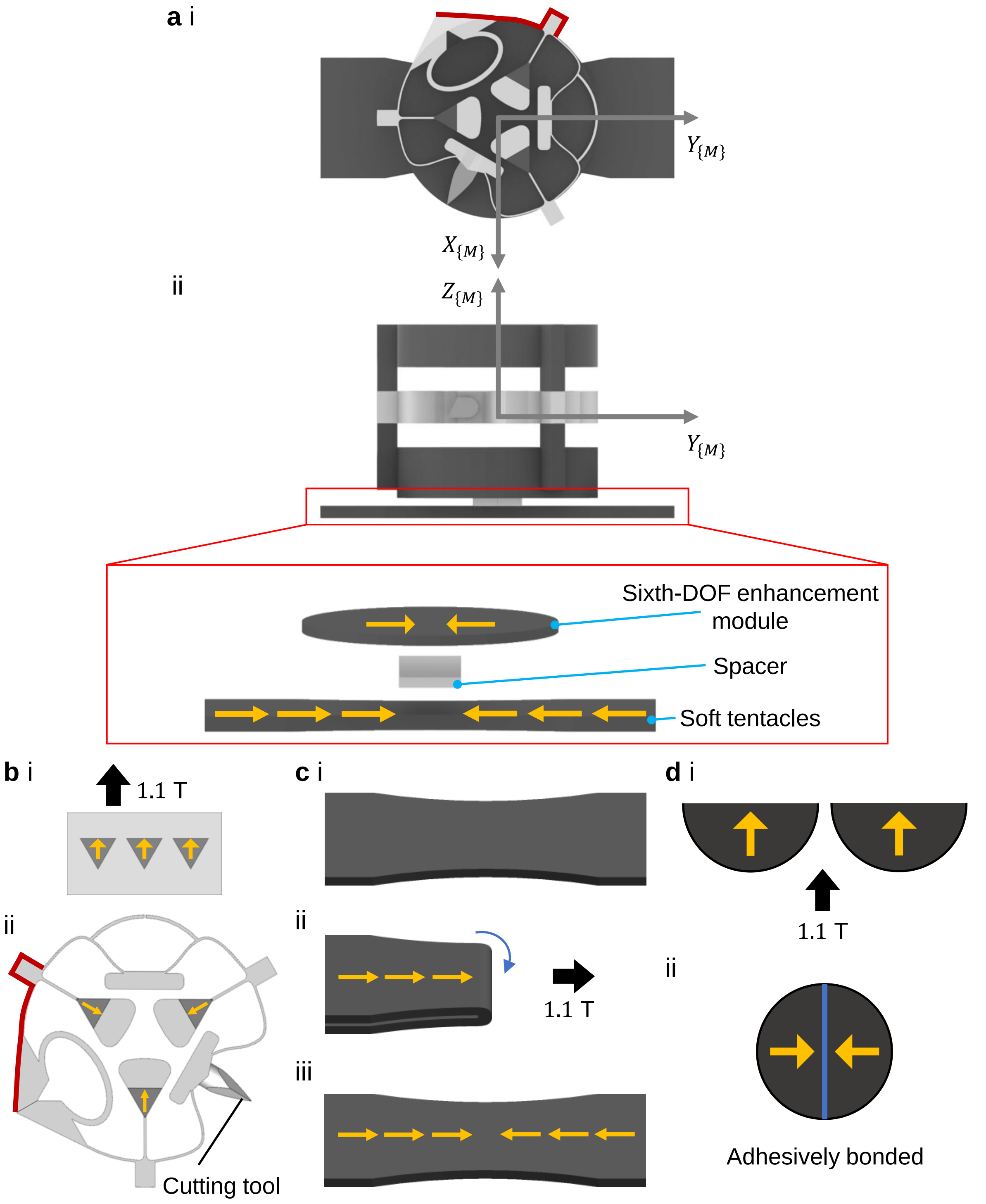

**Fig. S1 | Material coordinate frame and fabrication process of the soft robot. a,** (i)-(ii) show the material coordinate frame of our soft robot from the top and side view, respectively. The axes of the material coordinate frame are denoted by the subscript $\{M\}$. The soft robot's tentacles can be further broken down into several components: a sixth-DOF enhancement module, spacer and the soft part of the tentacles. **b,** The magnetization process for the magnetized components in the soft robot's main body. Once the magnets are molded and magnetized, they are attached along with the cutting tool to the remaining parts of the soft robot's main body. **c,** The magnetization process for the soft tentacles. **d,** The magnetization process for the sixth-DOF enhancement module. The semi-circles are adhesively bonded together with pristine PDMS. The yellow and black arrows represent the permanent magnetic dipole and magnetizing field, respectively.

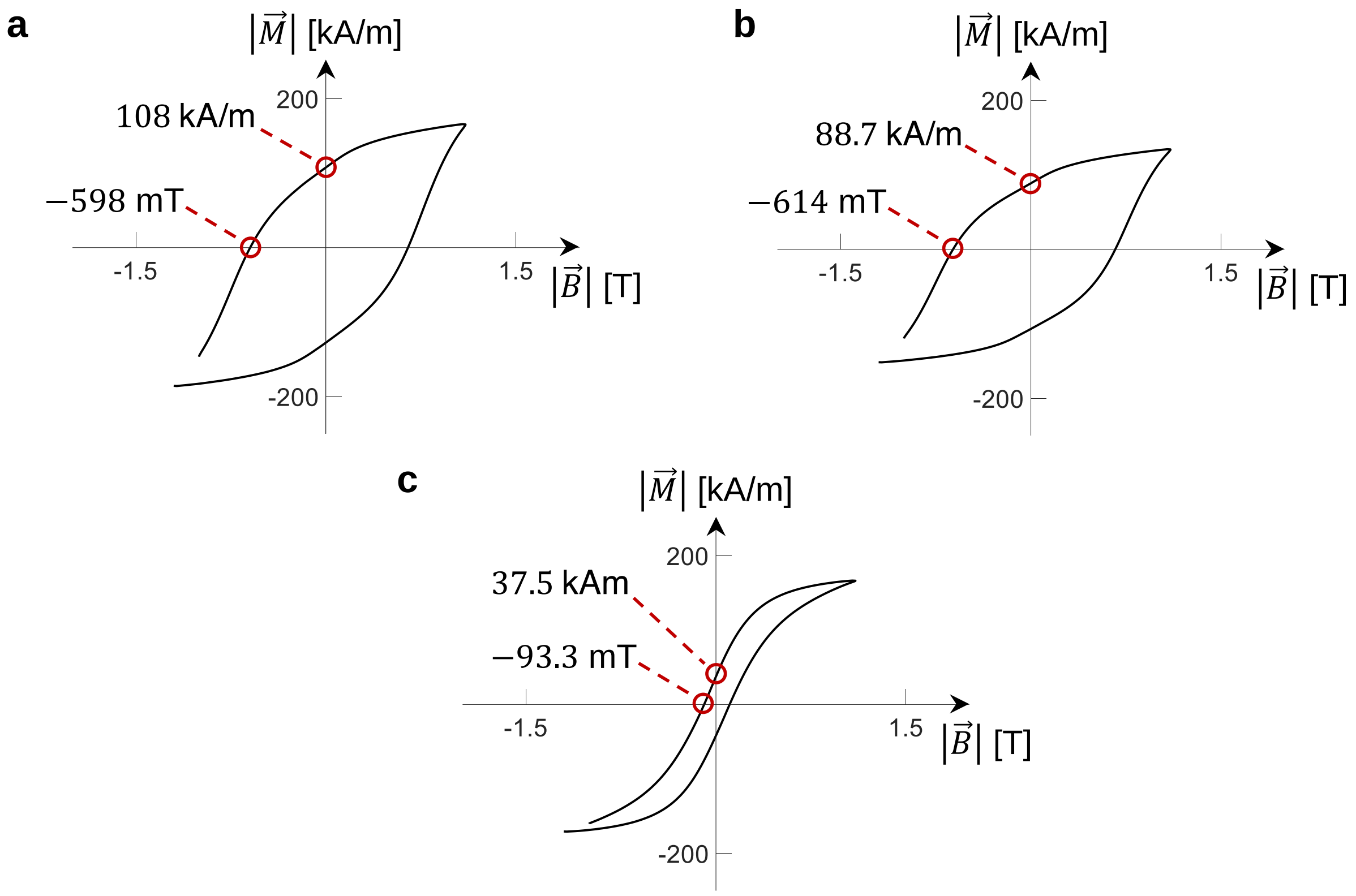

**Fig. S2 | Magnetic hysteresis loops for the components of the soft robot, which have hard magnetic properties. a,** The magnetic hysteresis loop for the material that is used for constructing the inner magnets. **b,** The magnetic hysteresis loop for the material that is used for constructing sixth-DOF enhancement module. **c,** The magnetic hysteresis loop for the material that is used for constructing the soft tentacles. The $Y$- and the $X$-axes intercepts of the graphs indicate the corresponding material's magnetization magnitude ($B_r$) and intrinsic coercivity ($H_{ci}$), respectively.

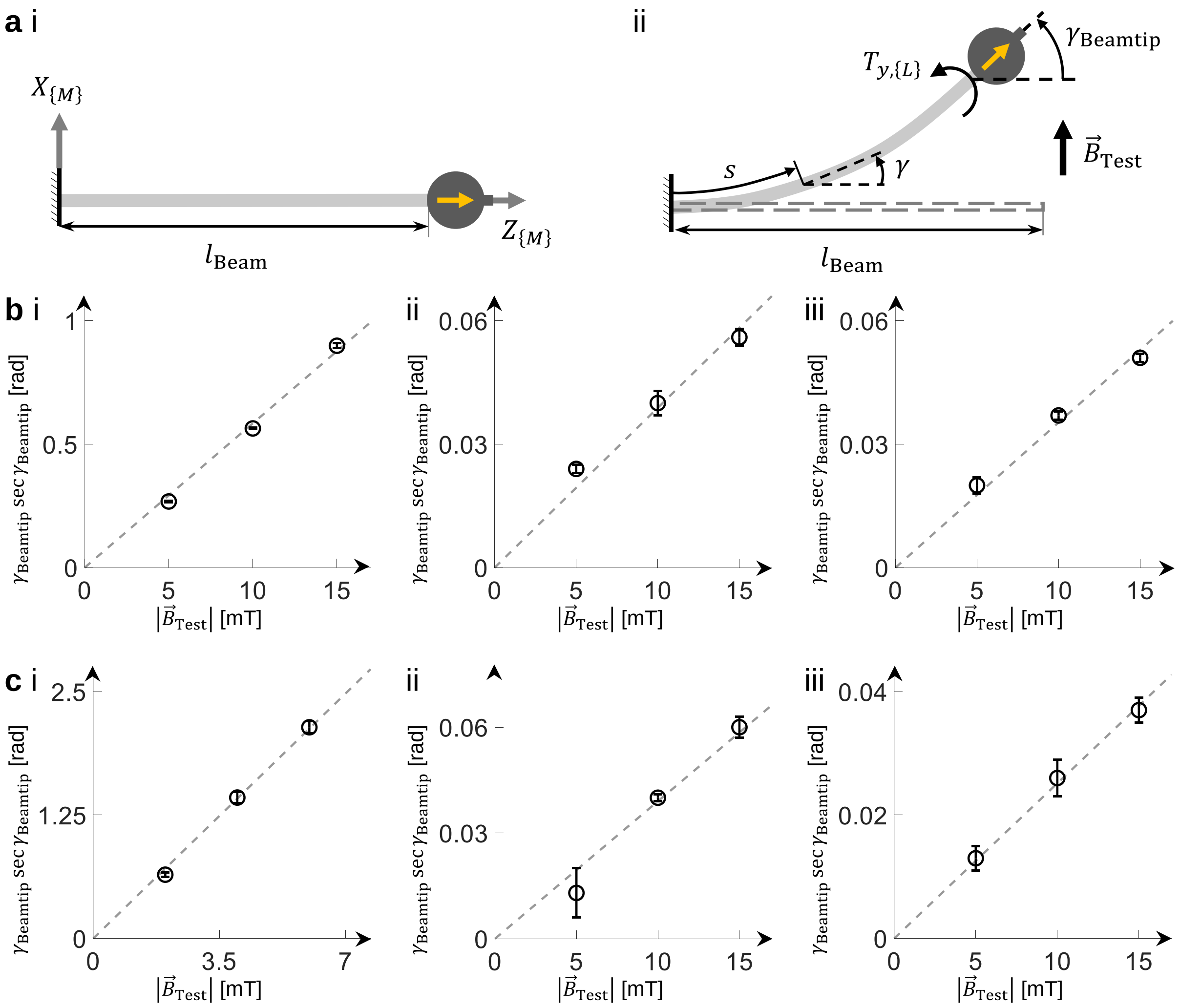

**Fig. S3 | Experimental setup and data for the characterization of $M_{\text{Rprog}}$ and $M_{\text{Heat}}$ (the customized actuation tests). a,** Schematic of the experimental set up. When $|\vec{B}_{\text{Test}}|$ is applied, the magnetized free end of the fixed-free beam will experience a torque of $T_{y,\{L\}}$, which will in turn deform the beam from its configuration in (i) to the one in (ii). The rotary deflection at the free end of the beam is denoted as $\gamma_{\text{Beamtip}}$. **b,** Plotting the samples' $\gamma_{\text{Beamtip}} \sec \gamma_{\text{Beamtip}}$ against $|\vec{B}_{\text{Test}}|$ with the stiffer beam. The experimental data when the free end of the fixed-free beam is attached with the materials, which construct the inner magnets, reprogrammable module (magnetized), remote heating component (magnetized) of the soft robot are shown in (i)-(iii), respectively. The gradient of the best fit line in (i)-(iii) are 0.0585 (A·s$^2$)/g, 0.00389 (A·s$^2$)/g and 0.00353 (A·s$^2$)/g, respectively. **c,** Plotting the samples' $\gamma_{\text{Beam}} \sec \gamma_{\text{Beam}}$ against $|\vec{B}_{\text{Test}}|$ with the more compliant beam. The experimental data when the free end of the fixed-free beam is attached with the materials, which construct the inner magnets, reprogrammable module (demagnetized), remote heating component (demagnetized) of the soft robot are shown in (i)-(iii), respectively. The gradient of the best fit line in (i)-(iii) are 0.354 (A·s$^2$)/g, 0.00390 (A·s$^2$)/g and 0.00251 (A·s$^2$)/g, respectively. All the best fit lines are constrained to pass through the origin and their gradients are used for calculating $M_{\text{Rprog}}$, $M_{\text{Heat}}$, and the flexural rigidity of the beams. Each data point is evaluated with five trials. The mean and standard deviation of the data points are represented by their centroid and error bars, respectively.

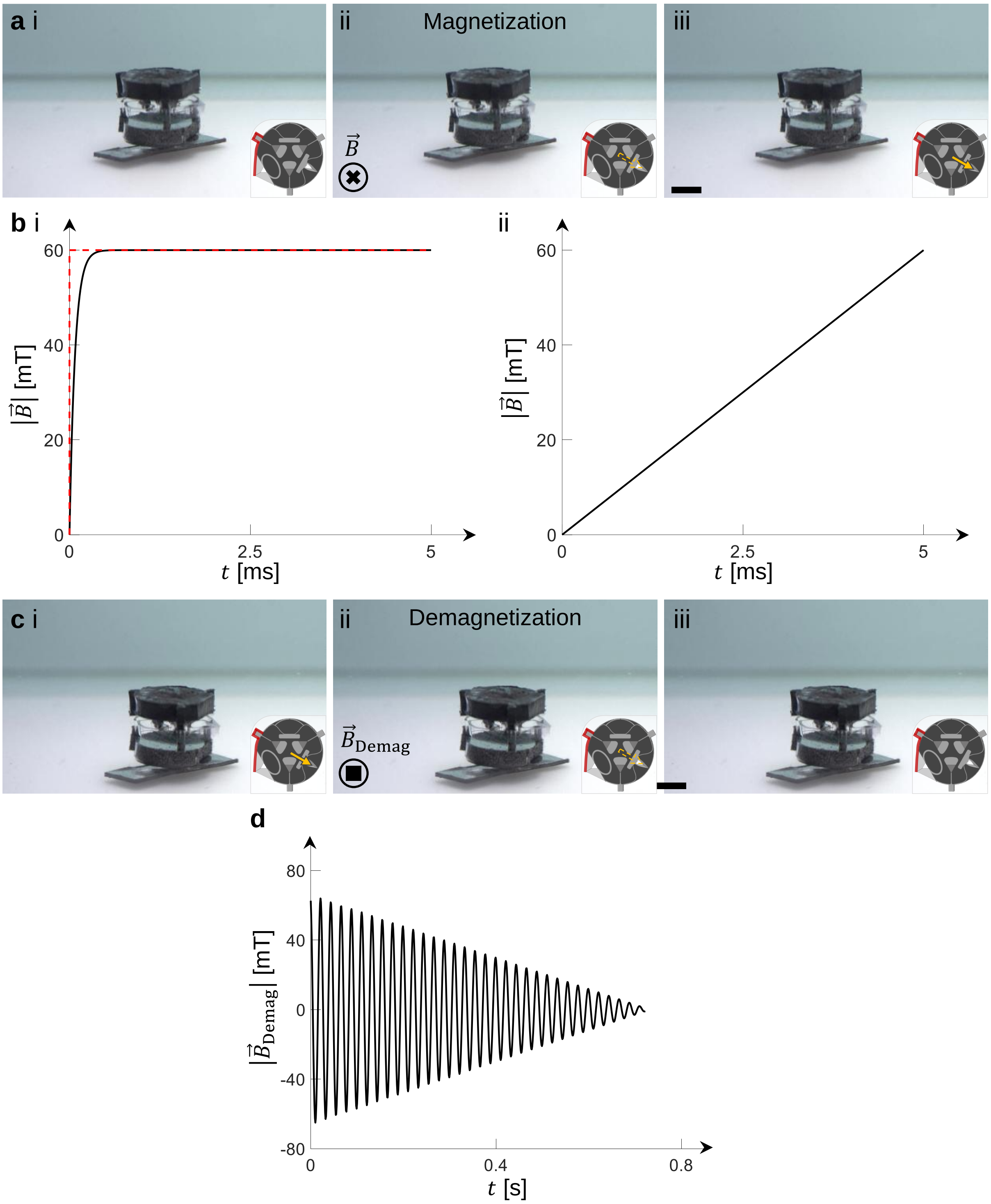

**Fig. S4 | Reprogramming the soft robot via the magnetizing and demagnetizing fields. a,** An illustration for the process of magnetizing the reprogrammable module and remote heating component such that the soft robot can be reprogrammed from its locomotion mode to its cutting mode. **b,** (i) The ideal step magnetizing field of 60 mT is shown as a red dashed line and the actual output field of our reprogramming coil is plotted as a black solid line. (ii) The ramp magnetizing field is plotted as a black solid line. **c,** The reprogrammable module and remote heating component are demagnetized such that the soft robot is reprogrammed from its cutting mode to its locomotion mode. **d,** The demagnetizing field (45 Hz) that can be applied to the soft robot. Scale bars, 1 mm.

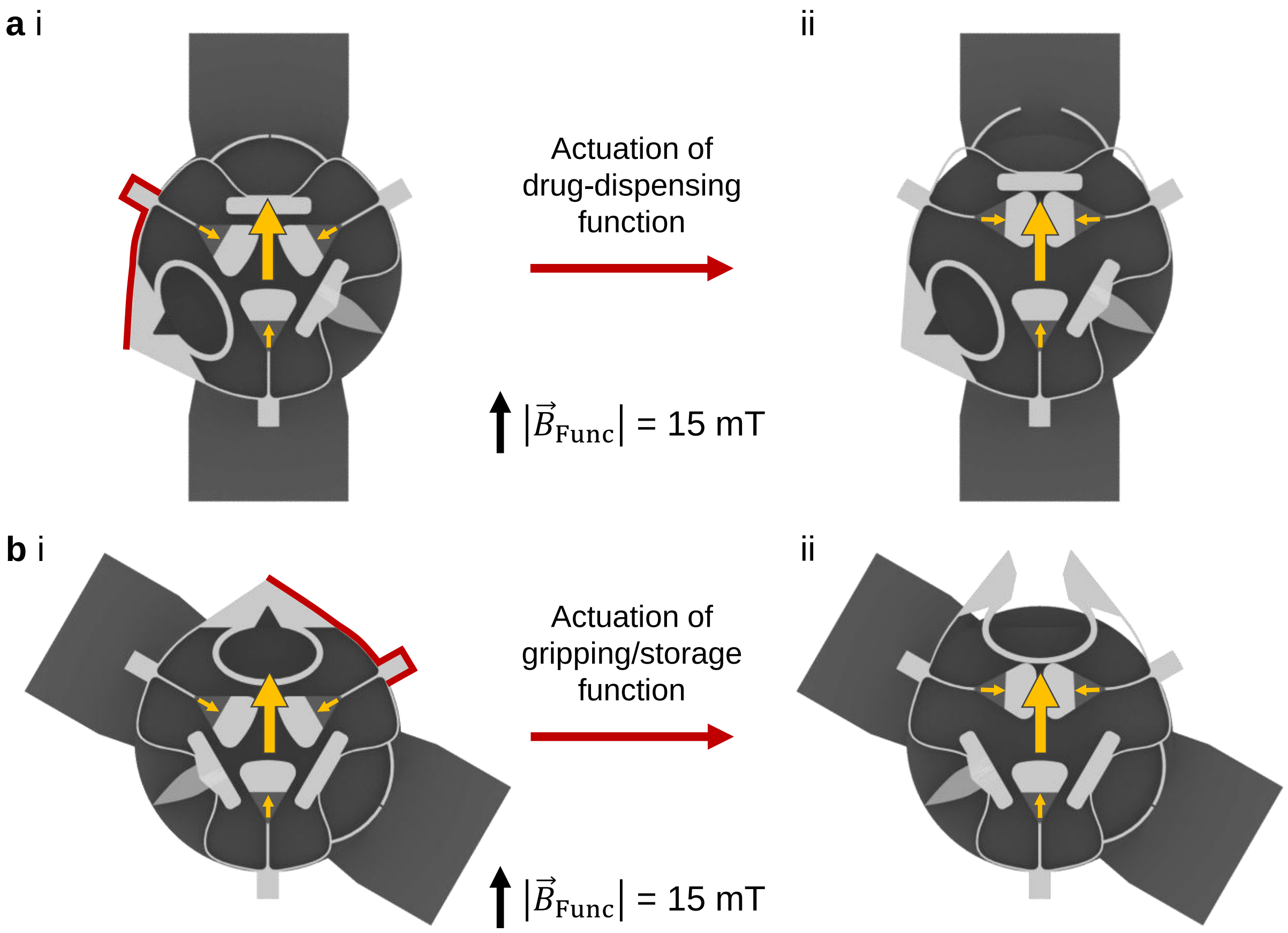

**Fig. S5 | Actuating the drug-dispensing and gripping/storage functions of the soft robot.** When $\vec{B}_{\mathrm{Func}}$ is applied, a pair of inner beams in the soft robot's main body will deform and activate a corresponding function. **a,** shows the drug-dispensing function of the soft robot. **b,** shows the gripping/storage function of the soft robot. The yellow and black arrows represent the magnetic dipole of the soft robot and $\vec{B}_{\mathrm{Func}}$, respectively.

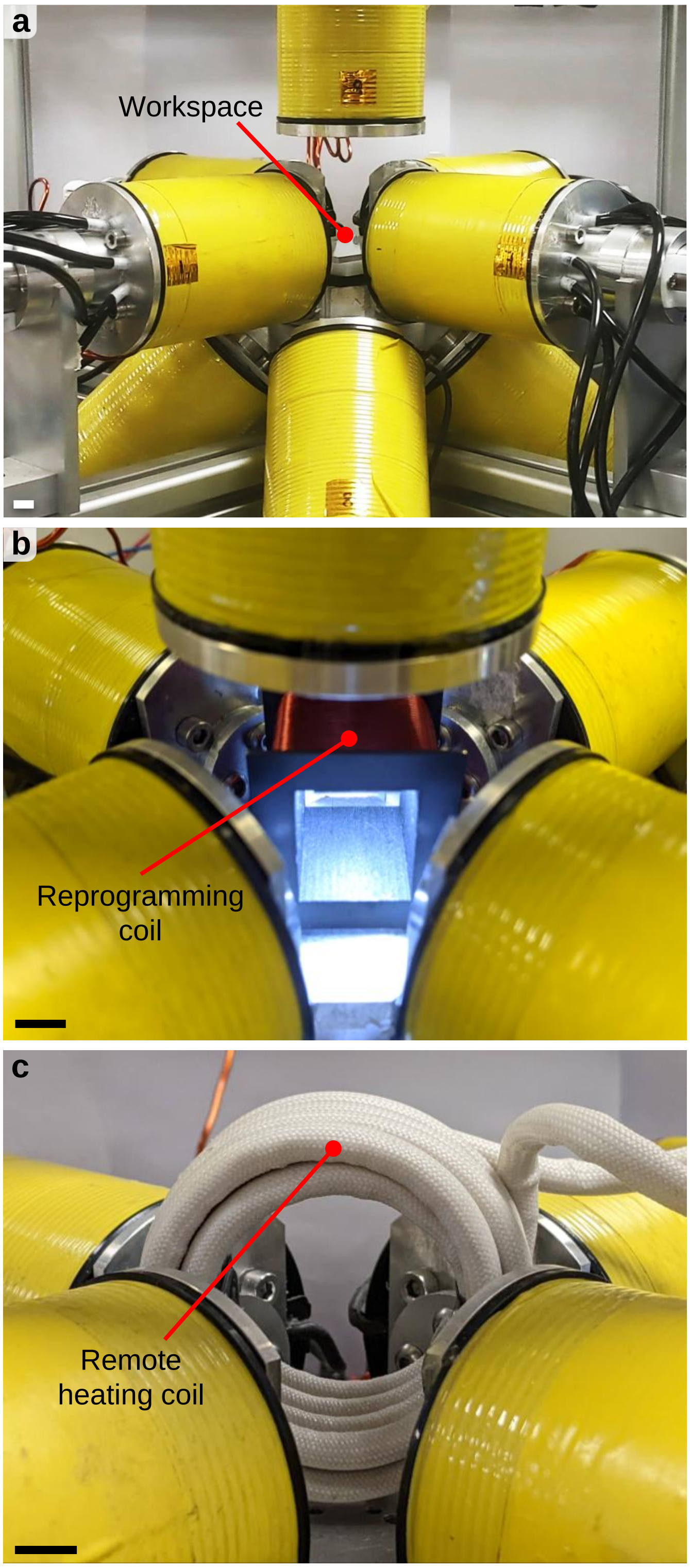

**Fig. S6 | Our electromagnetic coil system. a,** The workspace of our soft robot is located at the center of the electromagnetic coil system. **b,** Integrating the reprogramming coil with the electromagnetic coil system. **c,** Integrating the remote heating coil with the electromagnetic coil system. Due to space constraints, the reprogramming coil and the remote heating coil cannot be placed together within the workspace of our electromagnetic coil system. Scale bars, 1 cm.

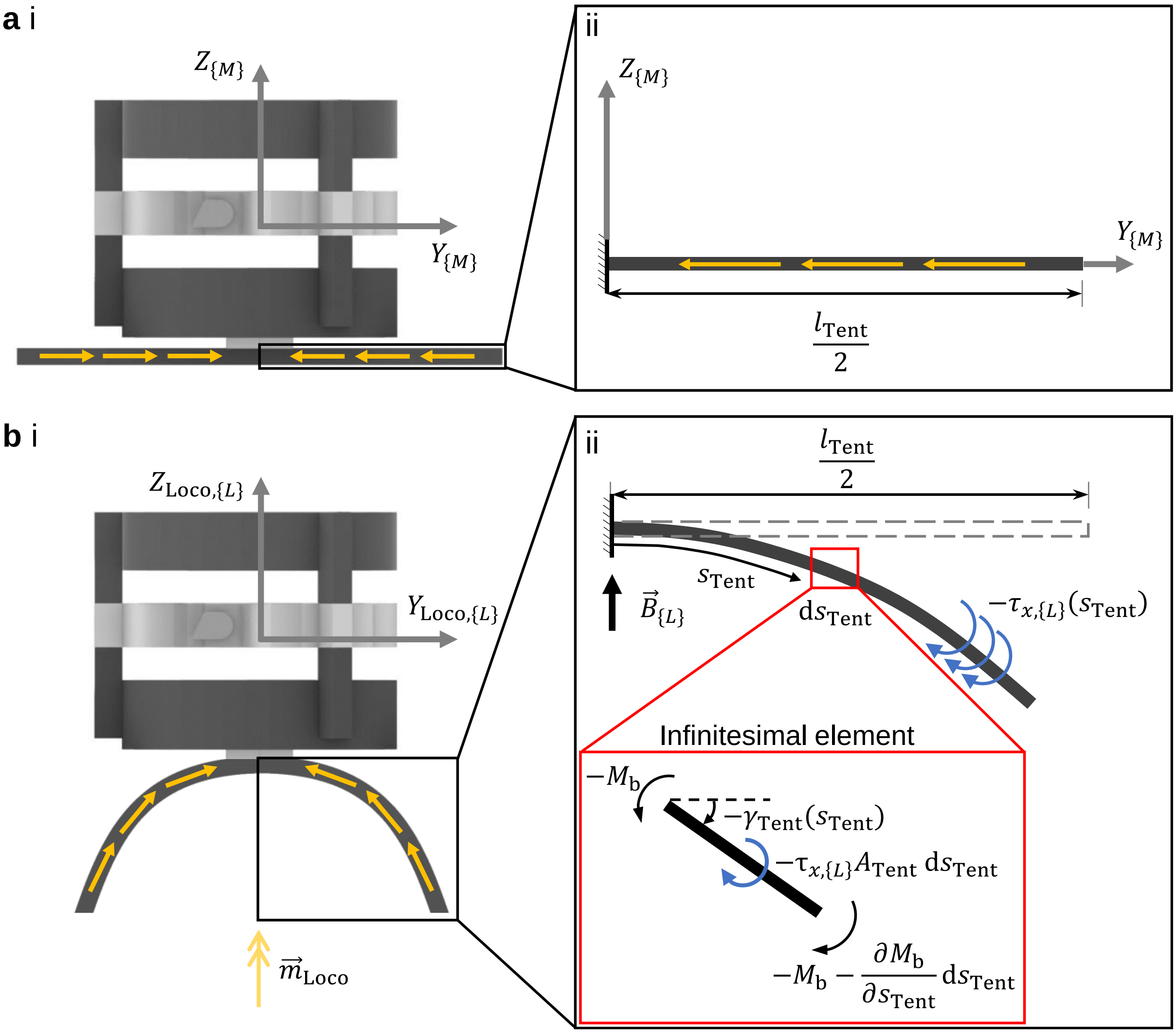

**Fig. S7 | Deformation analysis for the soft robot's tentacles. a,** shows the geometry and magnetization of the robot's tentacles before they deform. The material coordinate frame of the soft tentacles is denoted with the subscript: $\text{Tent(soft)}, \{M\}$. Due to symmetry, we can simplify the analysis by only considering the tentacles' deformation in the right half space. **b,** When $\vec{B}$ is applied, the tentacles will experience a distribution of magnetic torque of $(\tau_{x,\{L\}})$. For the ease of illustration, we only mark out a portion of this distribution of magnetic torque. When the tentacles deform, the soft robot will possess a net magnetic moment of $\vec{m}_{\text{Loco}}$. The local coordinate frame of the deformed soft robot is denoted with the subscript $\{L\}$. The yellow and black arrows represent the magnetic moment of the soft robot and $\vec{B}_{\{L\}}$, respectively.

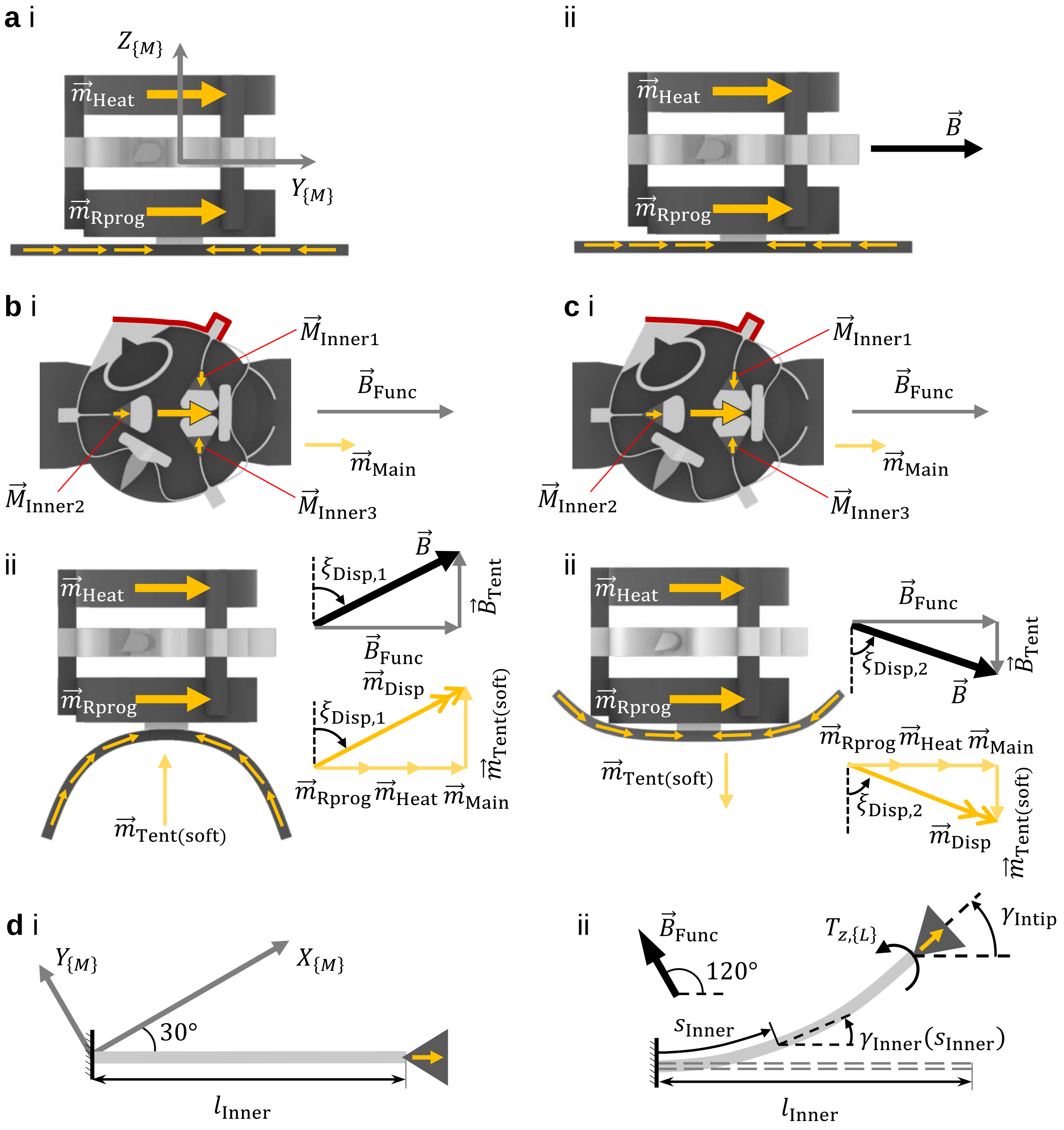

**Fig. S8 | Deformation analysis for the inner beams in the soft robot's main body (function mode). a,** When $\vec{B}_{\mathrm{Func}}$ is perfectly aligned with $\vec{m}_{\mathrm{Rprog}}$ and $\vec{m}_{\mathrm{Heat}}$, only the inner beams of the soft robot will deform. When $\vec{B}_{\mathrm{Func}}$ is not perfectly aligned with $\vec{m}_{\mathrm{Rprog}}$ and $\vec{m}_{\mathrm{Heat}}$, the soft tentacles and inner beams will deform concurrently. **b,** If the soft tentacles are deflected into an inverted 'U'-shaped configuration, the soft robot's net magnetic moment will deviate from the vertical axis (denoted by $\xi_{\mathrm{Disp},1}$). **c,** A similar analysis as those in **b**, except that here we assume that the tentacles of the soft robot are deformed into an upright 'U'-shaped configuration, and the soft robot's net magnetic moment deviates by $\xi_{\mathrm{Disp},2}$. **d,** Deformation analysis for one of the inner beams. By applying $\vec{B}_{\mathrm{Func}}$, the free end of the beam will experience a magnetic torque of $T_{z,\{L\}}$ and undergo a rotary deflection of $\gamma_{\mathrm{Intip}}$. The yellow and black arrows represent the magnetic dipole of the soft robot and $\vec{B}_{\mathrm{Func}}$, respectively. This illustration uses the drug-dispensing mode of the soft robot as an example.

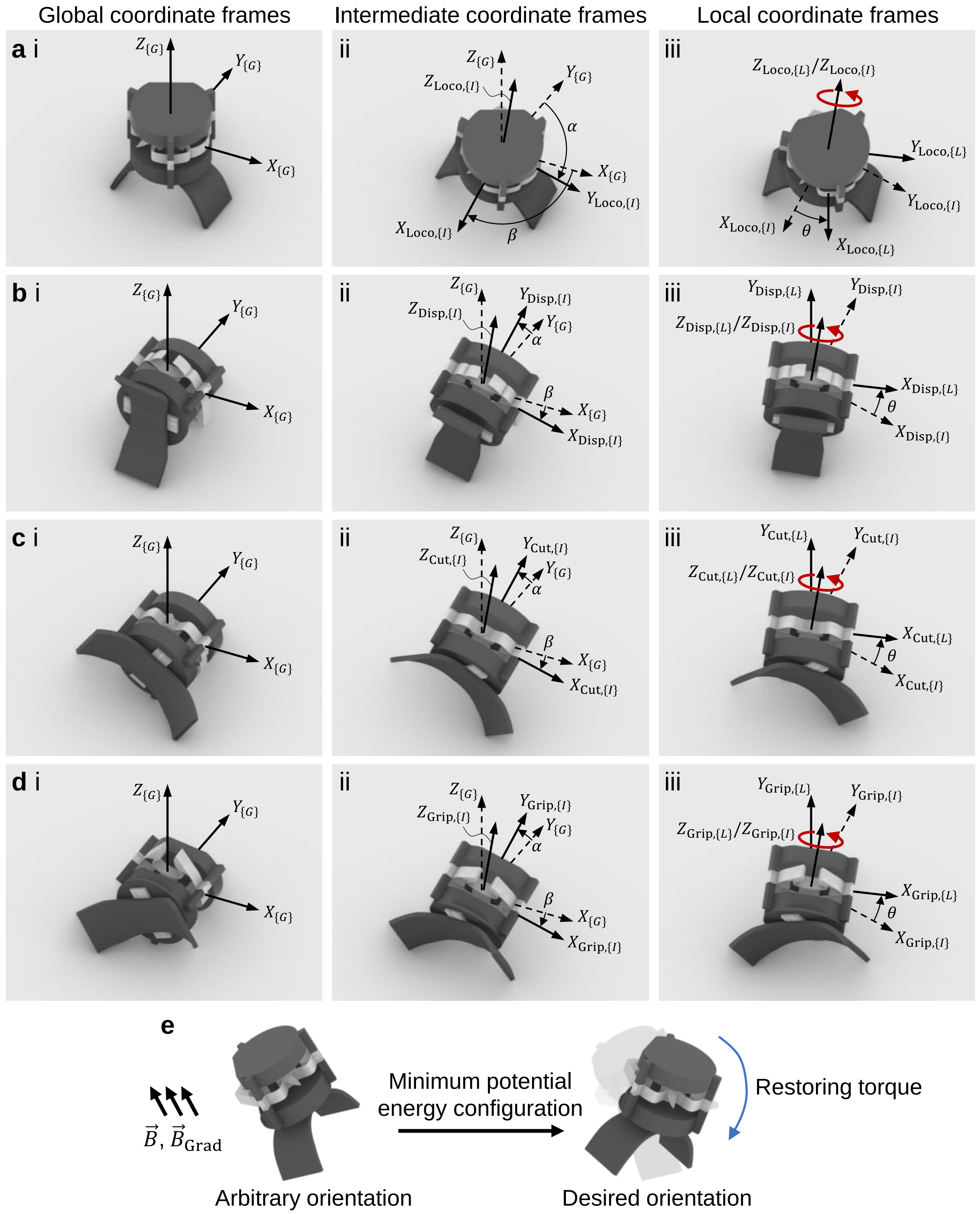

**Fig. S9 | Coordinate frames and control strategy of our soft robot.** The required global, intermediate, and local coordinate frames of the soft robot to describe its locomotion are denoted with the subscripts *G*, *I*, and *L*, respectively. In these illustrations, we assume that the local and global coordinate frames are coincident before the soft robot rotates. The required coordinate frames for describing the soft robot's locomotion when it is reprogrammed to its locomotion, drug-dispensing, cutting and gripping/storage mode are shown in **a-d**, respectively. **e,** The actuation strategy for our soft robot. We determine the required $\vec{B}$ and $\vec{B}_{\mathrm{Grad}}$ such that the desired orientation of our soft robot can become it's a minimum potential energy configuration. Using this actuation strategy, the soft robot will continuously experience three axes of restoring torque until it reaches the desired orientation. Desired magnetic forces can also be applied on the robot.

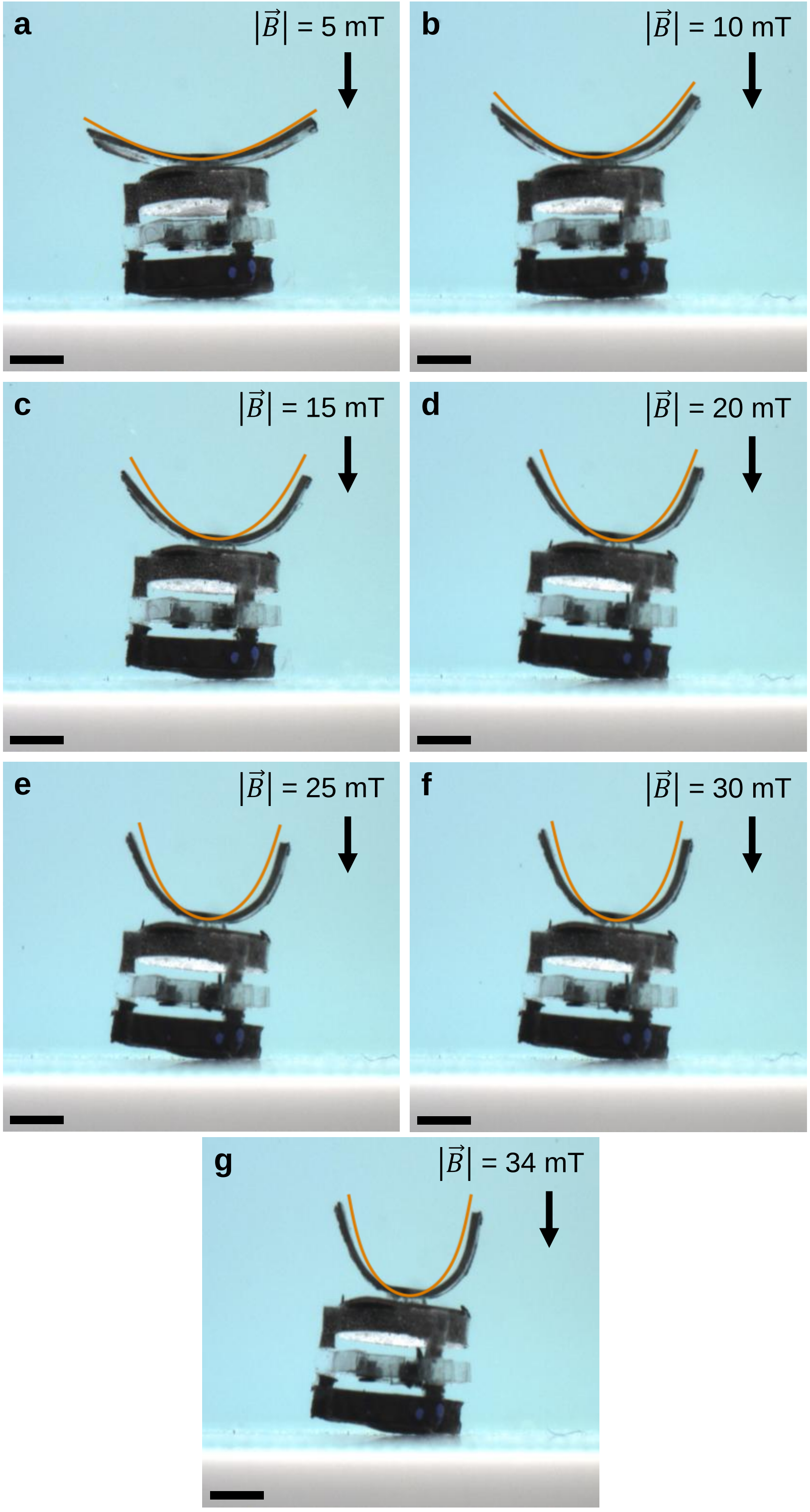

**Fig. S10 | Simulated and experimental deflection of the soft tentacles.** The experimental deflections of the soft tentacles are compared against their simulated predictions (represented by the superimposed orange line). Scale bars, 1 mm.

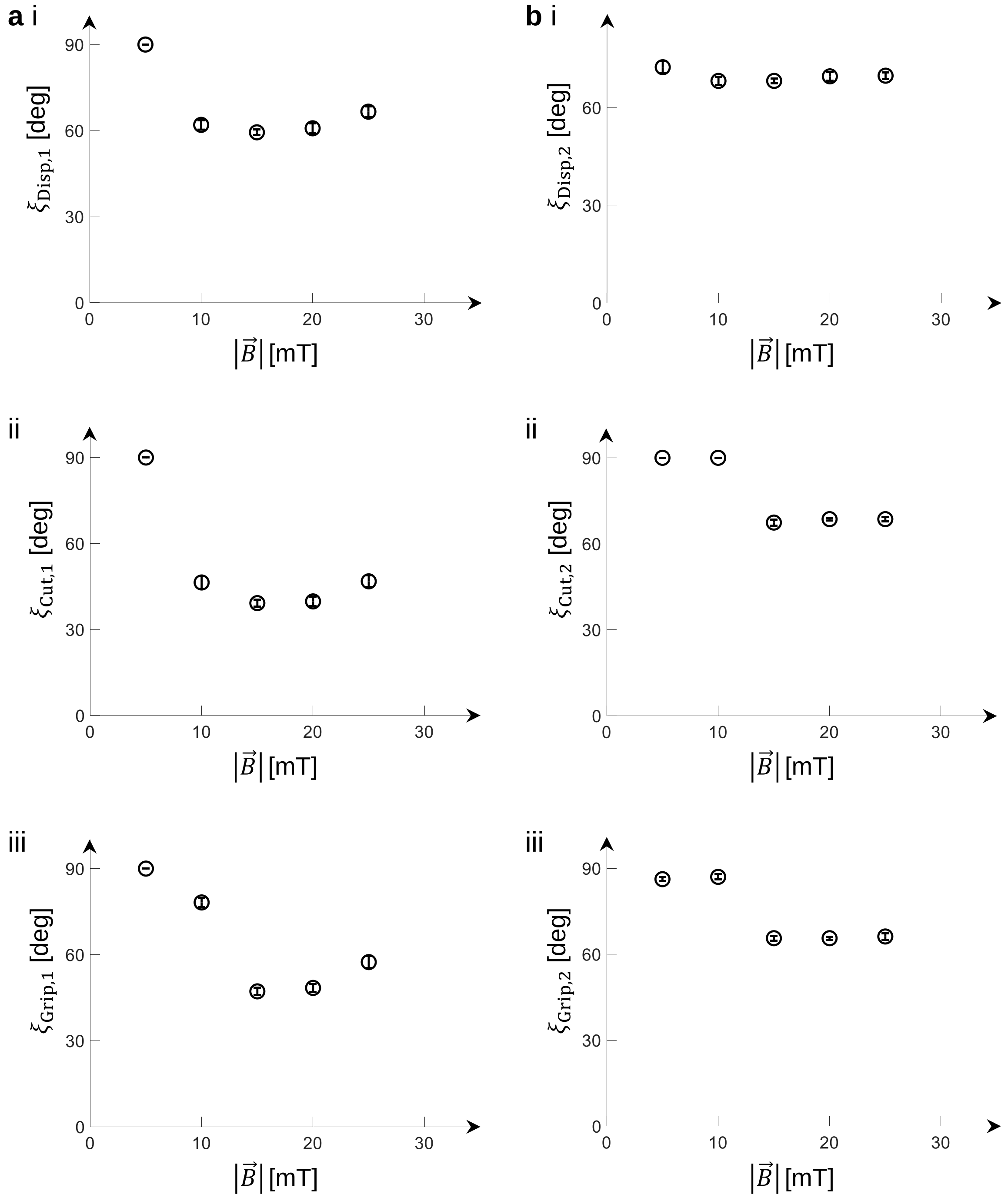

**Fig. S11 | Characterization of the soft robot's deviation angles.** The deviation angles are plotted against the applied $|\vec{B}|$. **a,** When the soft robot's soft tentacles are deformed in an inverted 'U'-shaped configuration, its corresponding deviation angles for different function modes are shown in (i)-(iii). **b,** When the soft robot's soft tentacles are deformed in an upright 'U'-shaped configuration, its corresponding deviation angles for different function modes are shown in (i)-(iii). Each data point is evaluated with five trials. The mean and standard deviation of the data points are represented by their centroid and error bars, respectively.

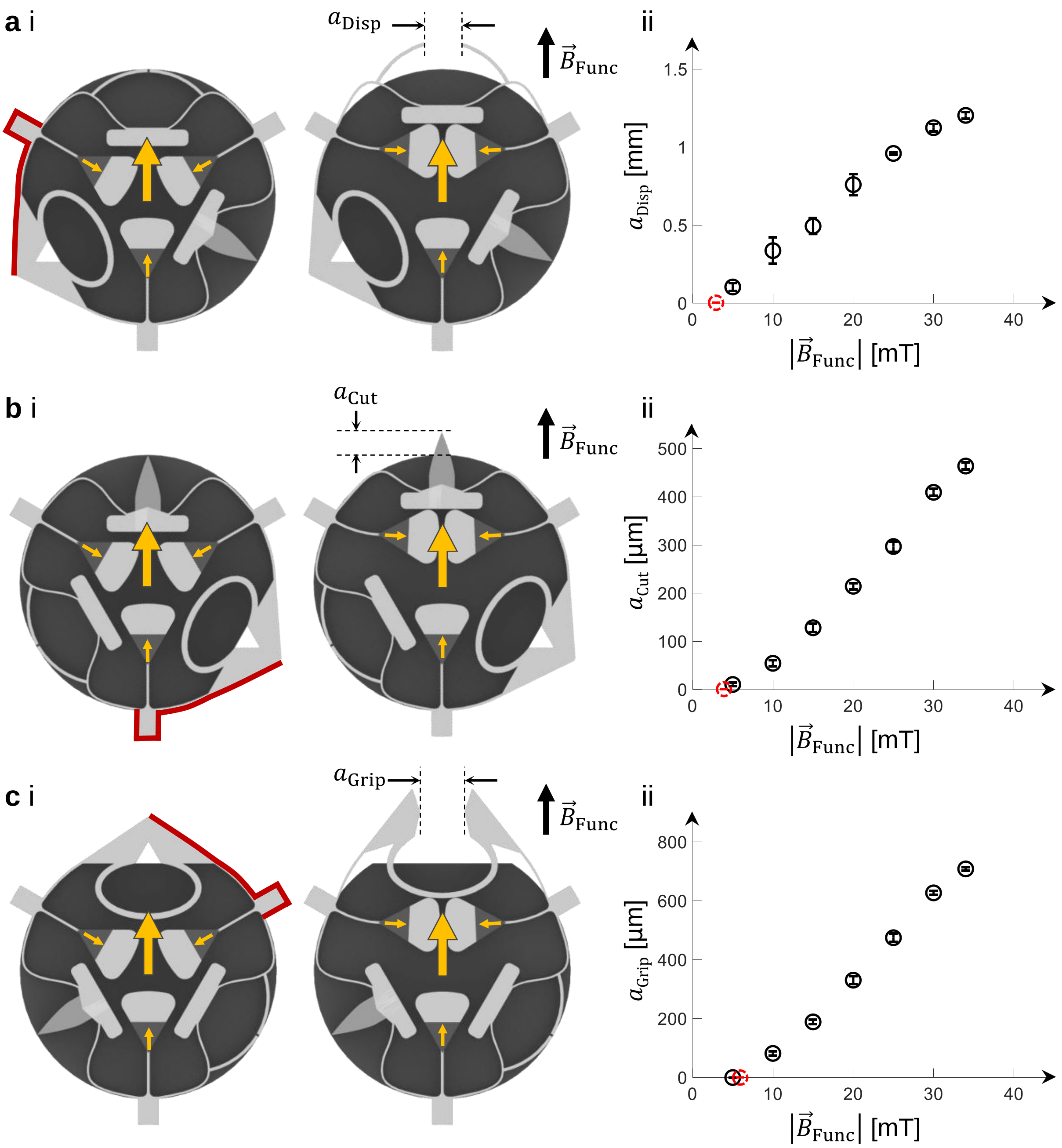

**Fig. S12 | Characterization for each surgical function of the soft robot. a,** The variable, $a_{\mathrm{Disp}}$, defines the shortest distance between the doors of the drug chamber when the soft robot is actuated in its drug-dispensing mode. The relationship between $a_{\mathrm{Disp}}$ and $|\vec{B}_{\mathrm{Func}}|$ is shown in (ii). **b,** The variable, $a_{\mathrm{Cut}}$, defines the extension of the cutting tool when the soft robot is actuated in its cutting mode. The relationship between $a_{\mathrm{Cut}}$ and $|\vec{B}_{\mathrm{Func}}|$ is shown in (ii). **c,** The variable, $a_{\mathrm{Grip}}$, defines the shortest distance between the grippers when the soft robot is actuated in its gripping/storage mode. The relationship between $a_{\mathrm{Grip}}$ and $|\vec{B}_{\mathrm{Func}}|$ is shown in (ii). The red dashed marker in each plot indicates the maximum applicable $|\vec{B}_{\mathrm{Func}}|$ that will not activate a respective surgical function. Specifically, their values are 3 mT, 4 mT, and 6 mT for the drug-dispensing, cutting and gripping/storage functions, respectively. Each data point is evaluated with five trials. The mean and standard deviation of the data points are represented by their centroid and error bars, respectively.

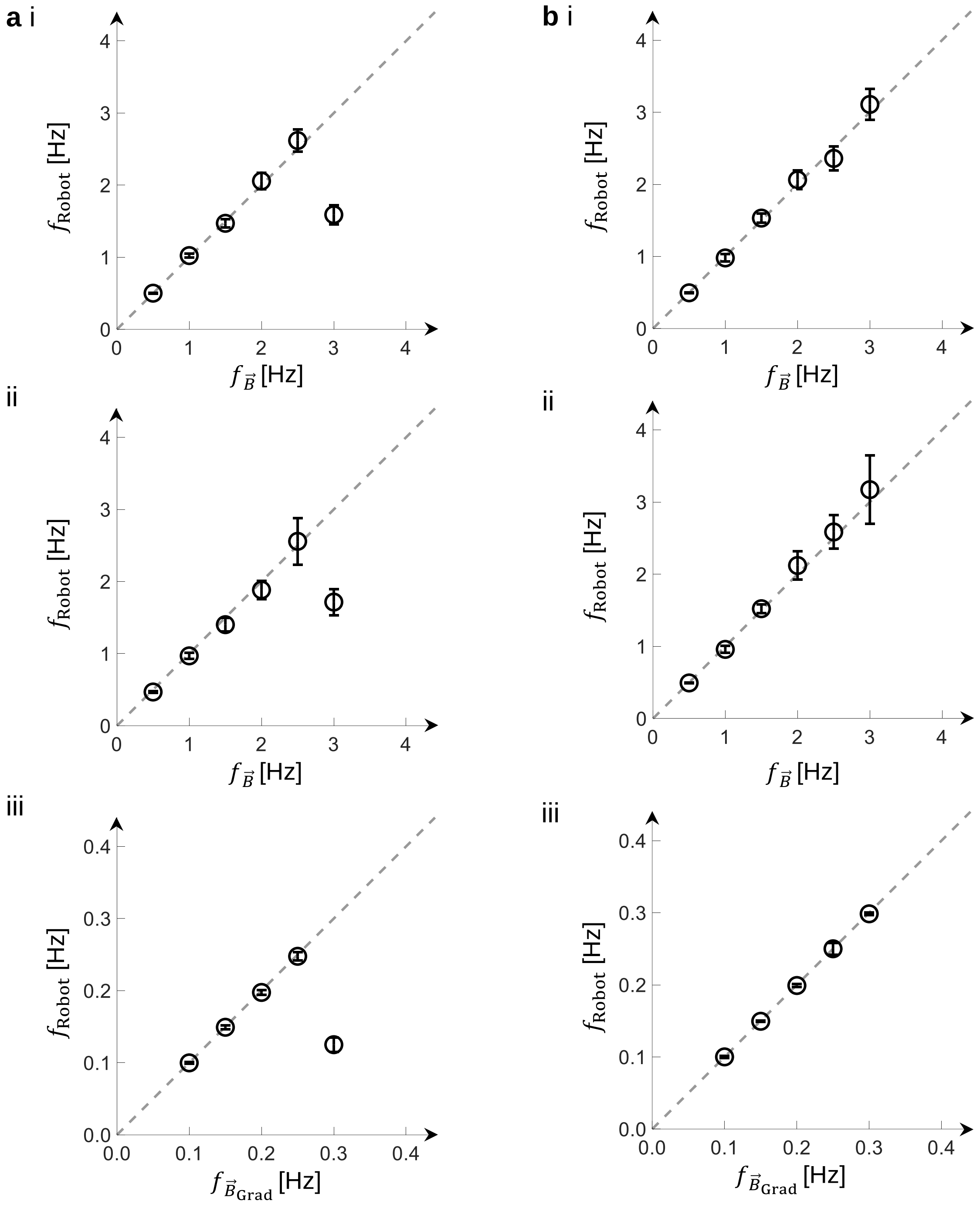

**Fig. S13 | Characterization of the soft robot's angular frequency (in its locomotion and drug-dispensing modes).** The soft robot's angular frequency ($f_{\text{Robot}}$) is plotted against the frequency of the rotating $\vec{B}$ ($f_{\vec{B}}$). The angular frequency of the robot is characterized in **a,** the locomotion mode about its (i) $X_{\text{Loco},\{L\}}$-axis, (ii) $Y_{\text{Loco},\{L\}}$-axis, and (iii) $Z_{\text{Loco},\{L\}}$-axis, as well as in **b,** the drug-dispensing mode along its (i) $X_{\text{Disp},\{L\}}$-axis, (ii) $Y_{\text{Disp},\{L\}}$-axis, and (iii) $Z_{\text{Disp},\{L\}}$-axis. The grey dashed line represents the ideal line where $f_{\text{Robot}} = f_{\vec{B}}$ in **a**(i)-(ii) and **b**(i)-(ii), and where $f_{\text{Robot}} = f_{\vec{B}_{\text{Grad}}}$ in **a**(iii) and **b**(iii). The frequency where the data point deviates from the ideal line represents the step-out frequency of the soft robot for that respective axis. Each data point is evaluated with five trials. The mean and standard deviation of the data points are represented by their centroid and error bars, respectively.

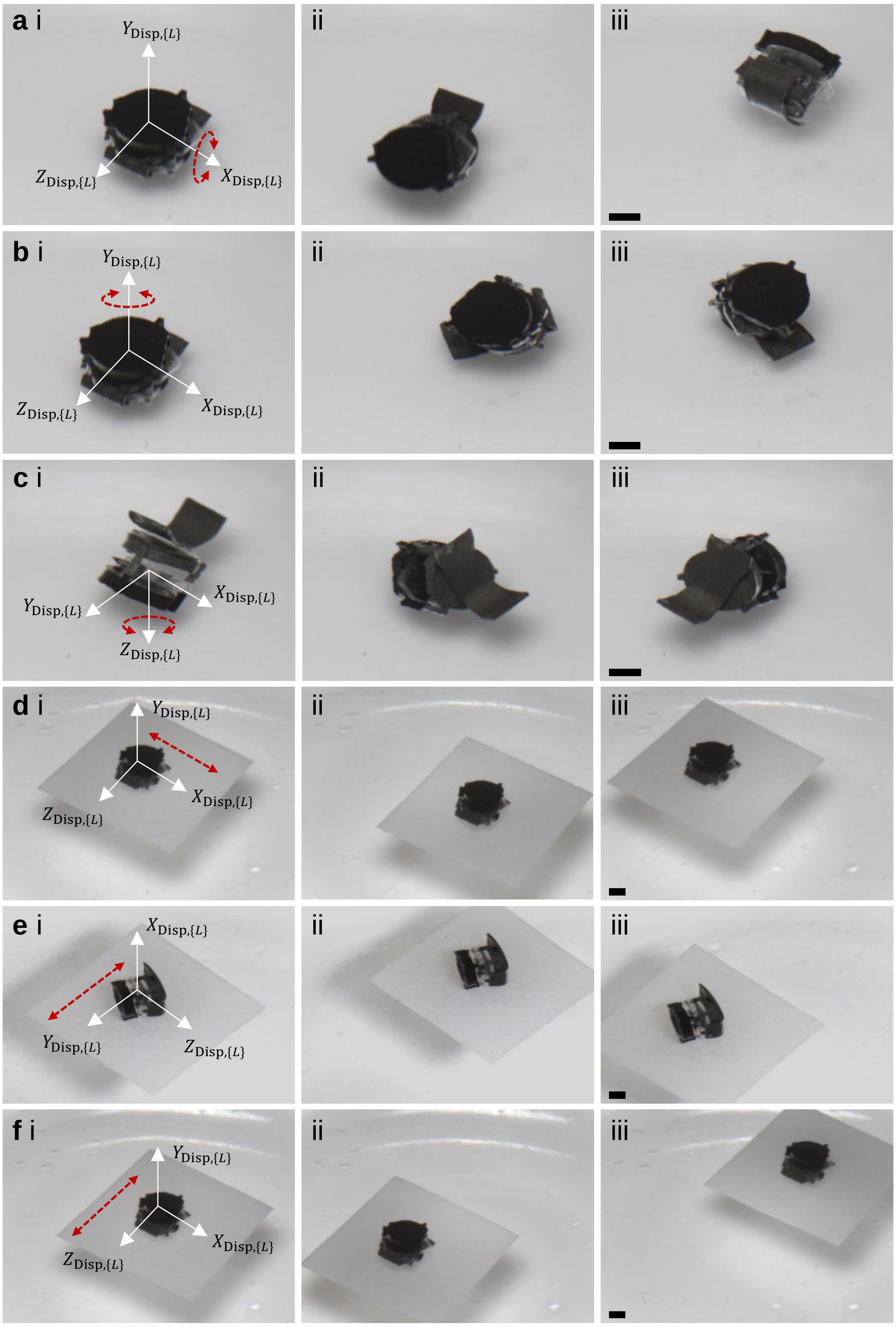

**Fig. S14 | Six degrees-of-freedom rigid body motions of the soft robot in its function mode (drug-dispensing mode).** When reprogrammed to the drug dispensing mode, the soft robot is able to rotate about its **a,** $X_{\mathrm{Disp},\{L\}}$-axis, **b,** $Y_{\mathrm{Disp},\{L\}}$-axis, and **c,** $Z_{\mathrm{Disp},\{L\}}$-axis, as well as translate along its **d,** $X_{\mathrm{Disp},\{L\}}$-axis, **e,** $Y_{\mathrm{Disp},\{L\}}$-axis and **f,** $Z_{\mathrm{Disp},\{L\}}$-axis. The initial position of the robot in **c** and **e** are rotated such that their respective rotation and translation can be shown more clearly. The red dashed arrows indicate the respective rotations about or translations along the robot's local coordinate frame in the drug-dispensing mode. Additionally, the translation experiments are conducted with the robot on top of a piece of paper floated on the oil-water interface to reduce friction. Scale bars, 1 mm.

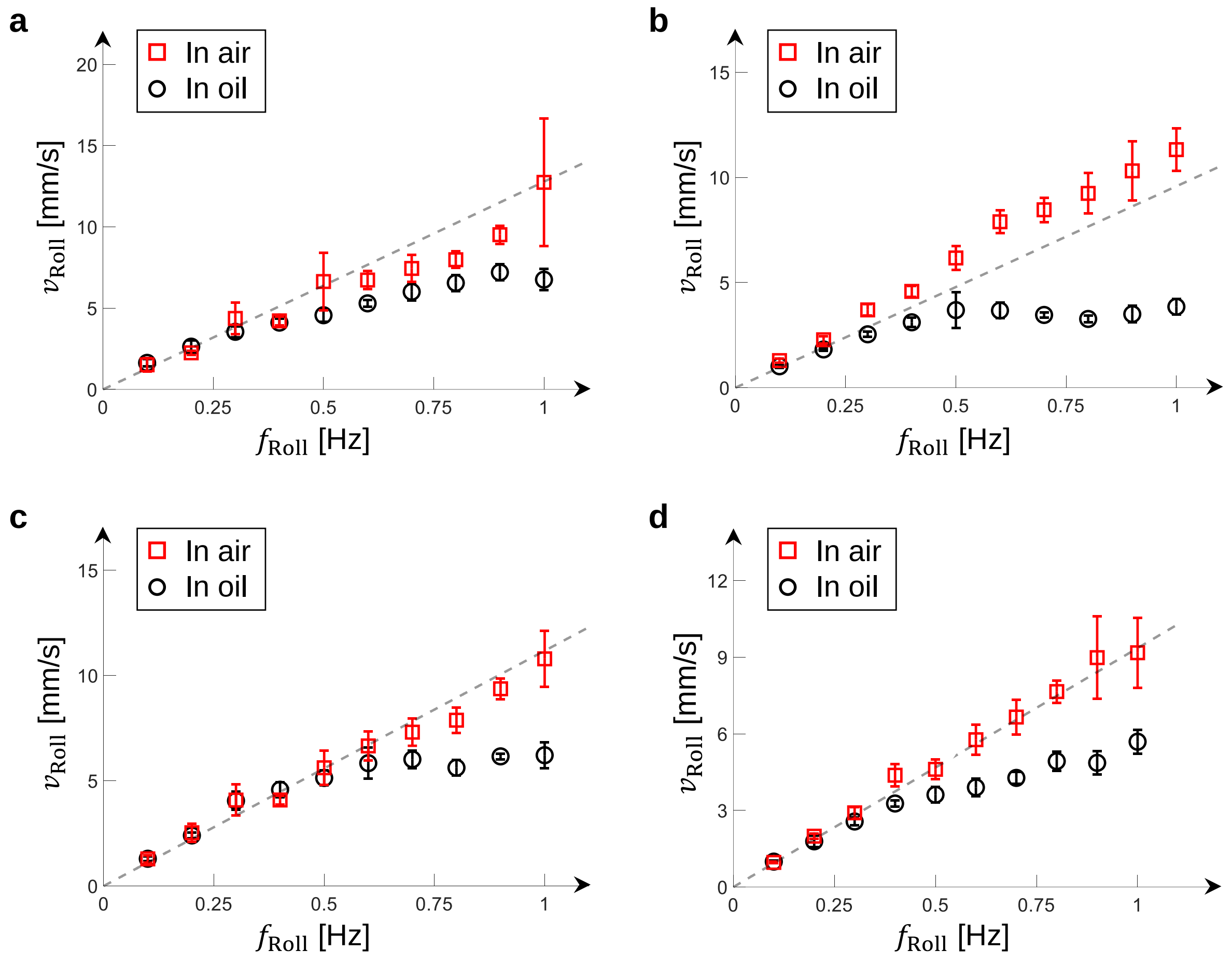

**Fig. S15 | Characterization of the soft robot's rolling velocity.** The soft robot's rolling velocity ($v_{\mathrm{Roll}}$) is plotted against the frequency of the rotating $\vec{B}$ ($f_{\mathrm{Roll}}$). The robot's rolling velocity is characterized in both air and oil environments. **a-b**, The experimental data of the soft robot when it rolls along its length with its soft tentacles deforming to an inverted and upright 'U'-shaped configurations, respectively. **c-d,** The experimental data of the soft robot when it rolls along its width with its soft tentacles deforming to an inverted and upright 'U'-shaped configurations, respectively. The ideal $v_{\mathrm{Roll}}$ is plotted as a grey solid line. The deviations between the measured and ideal rolling velocity may be caused by slipping effects. Each data point is evaluated with five trials. The mean and standard deviation of the data points are represented by their centroid and error bars, respectively.

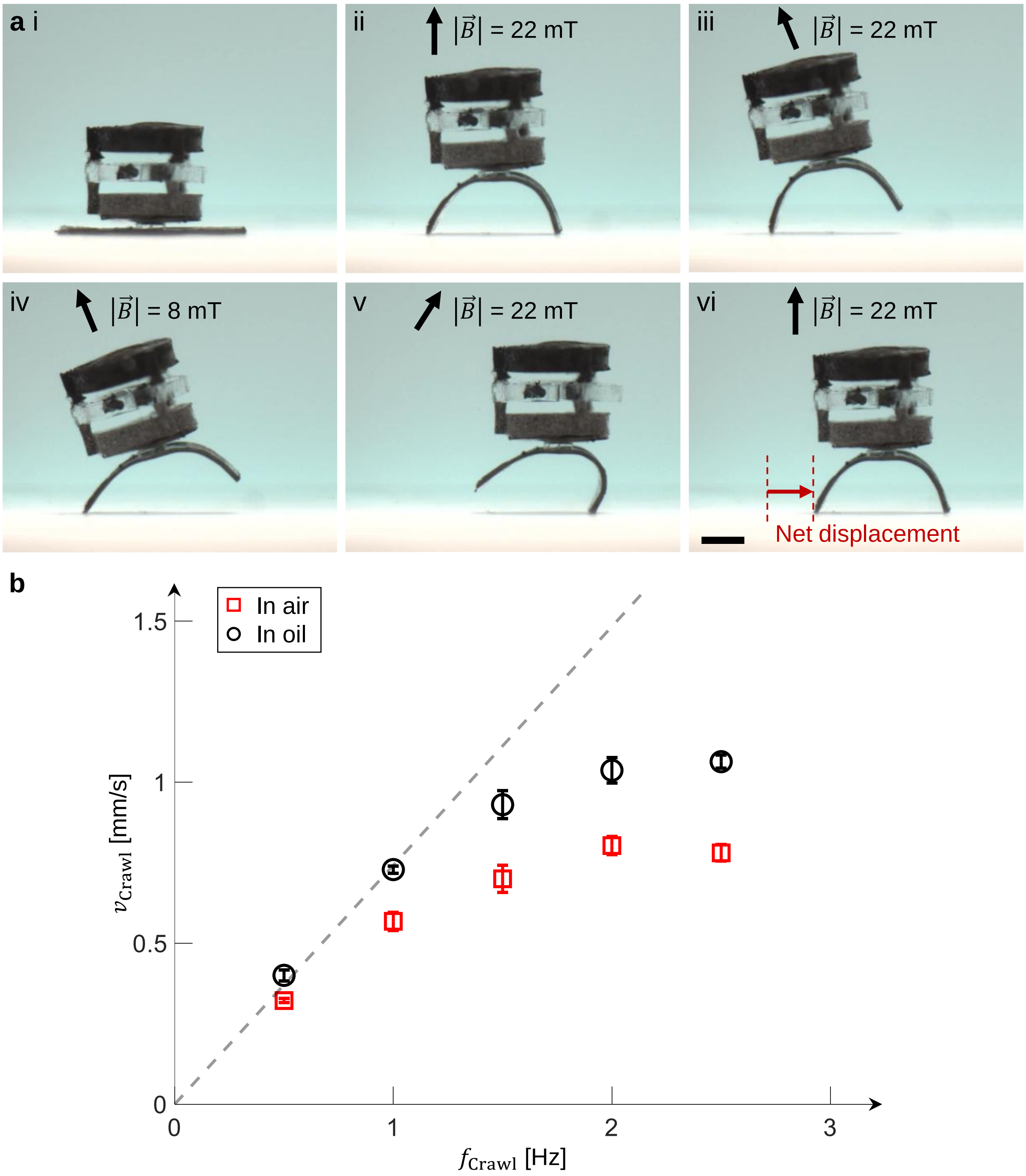

**Fig. S16 | The soft robot's two-anchor crawling gait. a,** A cycle of the two-anchor crawling gait. **b,** The soft robot's two-anchor crawling speed ($v_{\mathrm{Crawl}}$) is plotted against the frequency of the applied $\vec{B}$ ($f_{\mathrm{Crawl}}$) in both air and in oil environments. The ideal crawling velocity is equal to the product of $f_{\mathrm{Crawl}}$ and the soft robot's net displacement in one cycle (0.742 mm) and this ideal velocity is plotted as a grey dashed line in the graph. The deviations between the measured and ideal crawling velocities may be caused by slipping effects. The mean and standard deviation of the data points are represented by their centroid and error bars, respectively. Scale bars, 1 mm.

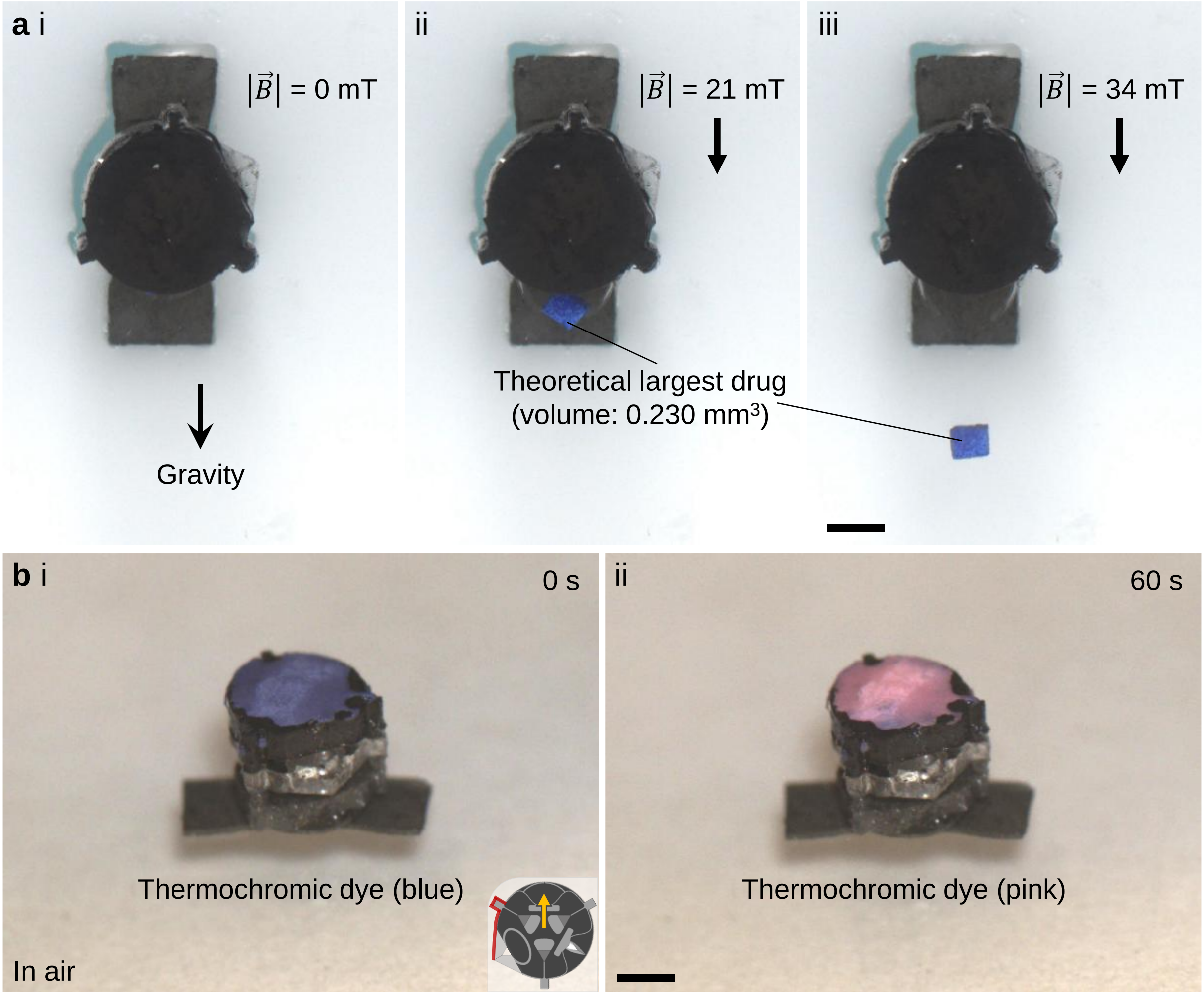

**Fig. S17 | Maximum drug-dispensing capacity and remote heating actuation (drug-dispensing mode). a,** shows the largest theoretical drug (0.230 $mm^3$) that our soft robot can dispense. In this experiment, we have constrained the tentacles of the soft robot such that they do not deform. **b,** The soft robot is commanded to heat a layer of thermochromic dye remotely in air while it is reprogrammed to its drug-dispensing mode. The thermochromic dye is blue when its temperature is below 40°C, and it will change to pink when the temperature is equal to or above 40°C. The remote heating magnetic field was applied for one minute. Scale bars, 1 mm.

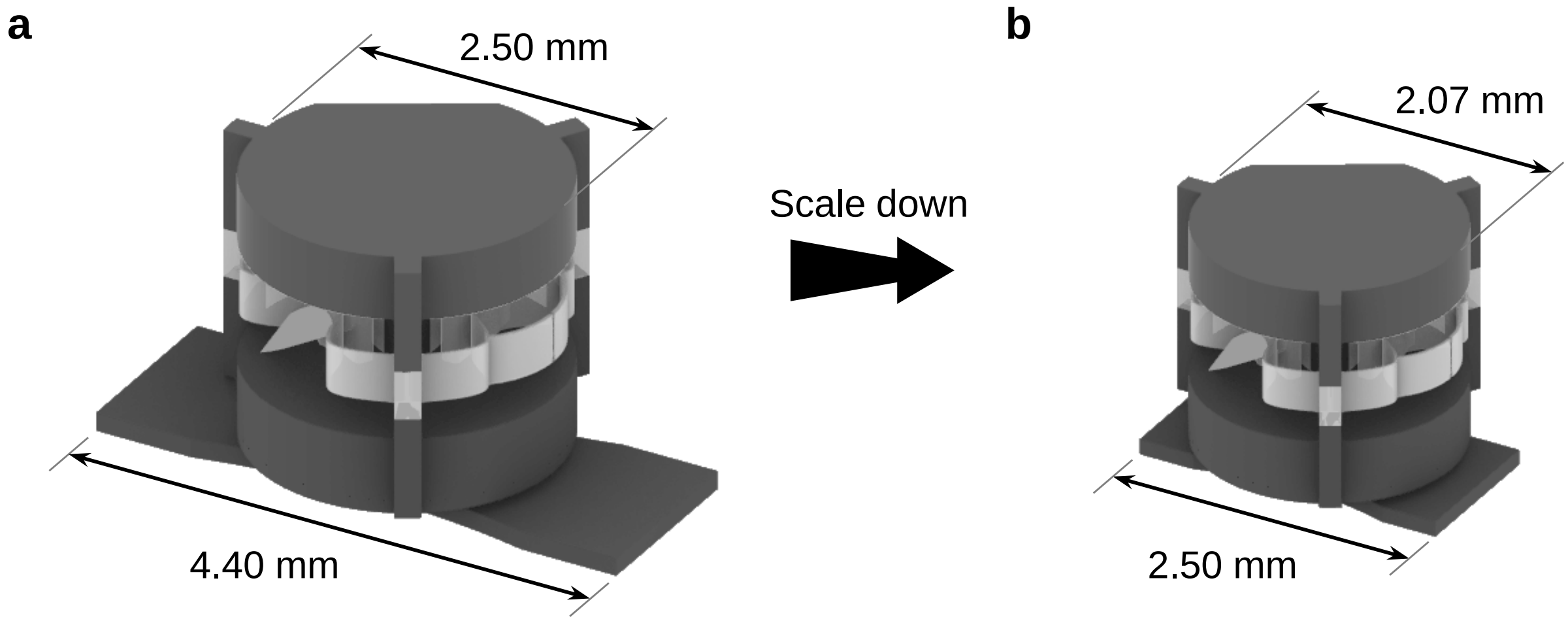

**Fig. S18 | Scaling analysis of our soft robot. a,** The dimensions of our current soft robot. **b,** Dimensions of the scaled-down soft robot. In this analysis, we assume that the main body of the soft robot is scaled down isotropically, while the length of its soft tentacles is scaled down from 4.4 mm to 2.5 mm.

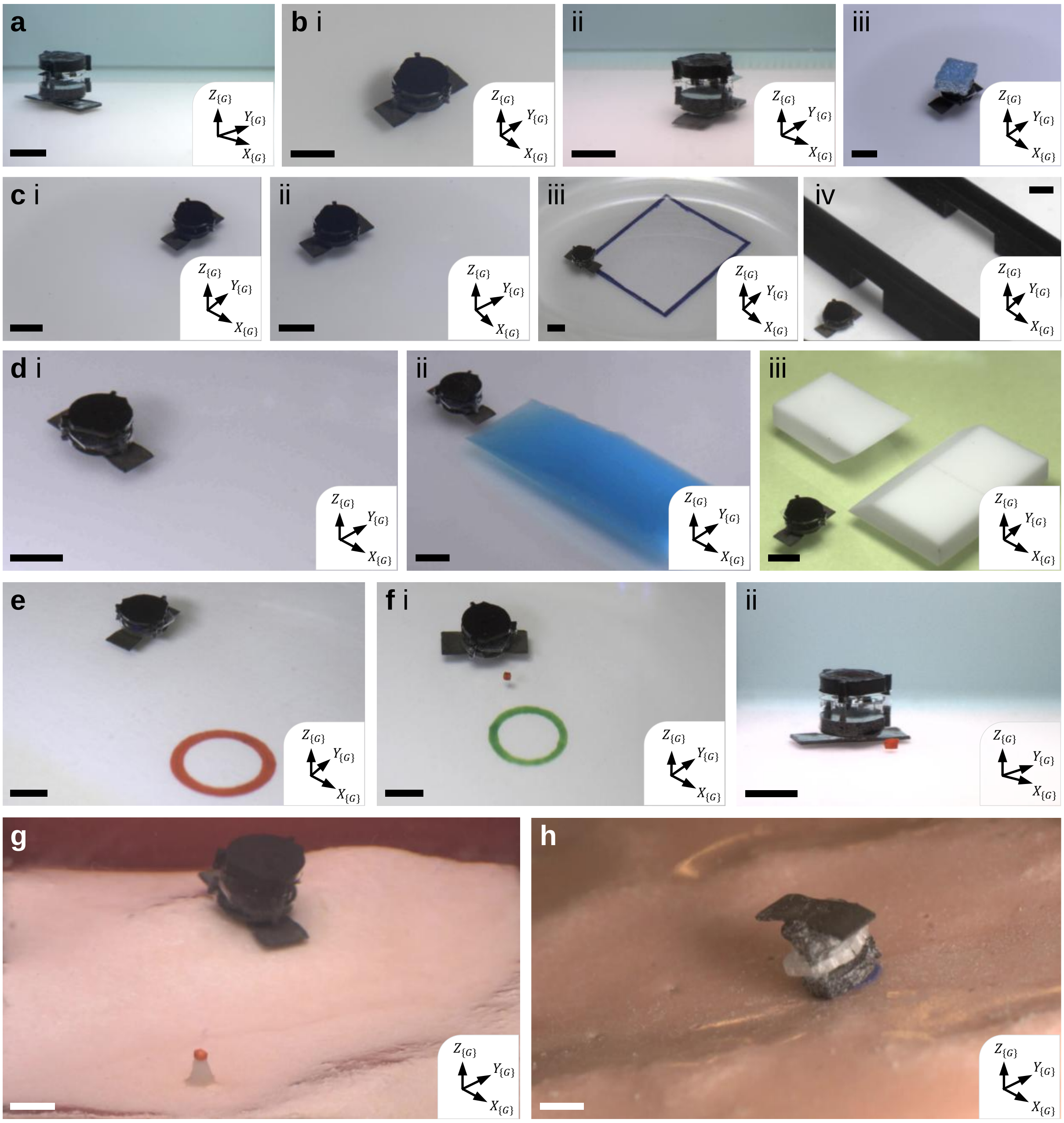

**Fig. S19 | Global reference frames for the experiments shown in SI Videos S1-S5, S7, S9 and S10. a,** SI Video S1**,** demonstrating the soft robot's reprogramming strategies. **b,** SI Video S2, showing (i) rotations about the soft robot's $X_{\text{Loco},\{L\}}$-, $Y_{\text{Loco},\{L\}}$-, and (ii) $Z_{\text{Loco},\{L\}}$-axes as well as (iii) translations along the soft robot's $X_{\text{Loco},\{L\}}$-, $Y_{\text{Loco},\{L\}}$-, and $Z_{\text{Loco},\{L\}}$-axis. **c,** SI Video S3, the robot demonstrates (i) rolling along its length and (ii) width, (iii) steers its rolling locomotion to follow a path, and (iv) rolling across challenging barriers. **d,** SI Video S4, demonstrating the two-anchor crawling locomotion of the robot with (i) steering, (ii) ascending a slope of 15°, and (iii) overcoming a challenging barrier with strict shape constraints. **e,** SI Video S5, showing the robot rolling to dispense a drug at the targeted location. **f,** SI Video S7, demonstrating (i) the pick-and-place function, and (ii) the soft robot picking, storing, and safely transporting a synthetic excised tissue. **g,** SI Video S9, demonstrating the soft robot's ability to perform locomotion and four surgical functions on a biological phantom. **h,** SI Video S10, demonstrating the soft robot's ability to perform locomotion and targeted remote heating on a biological phantom. Scale bars, 2 mm.

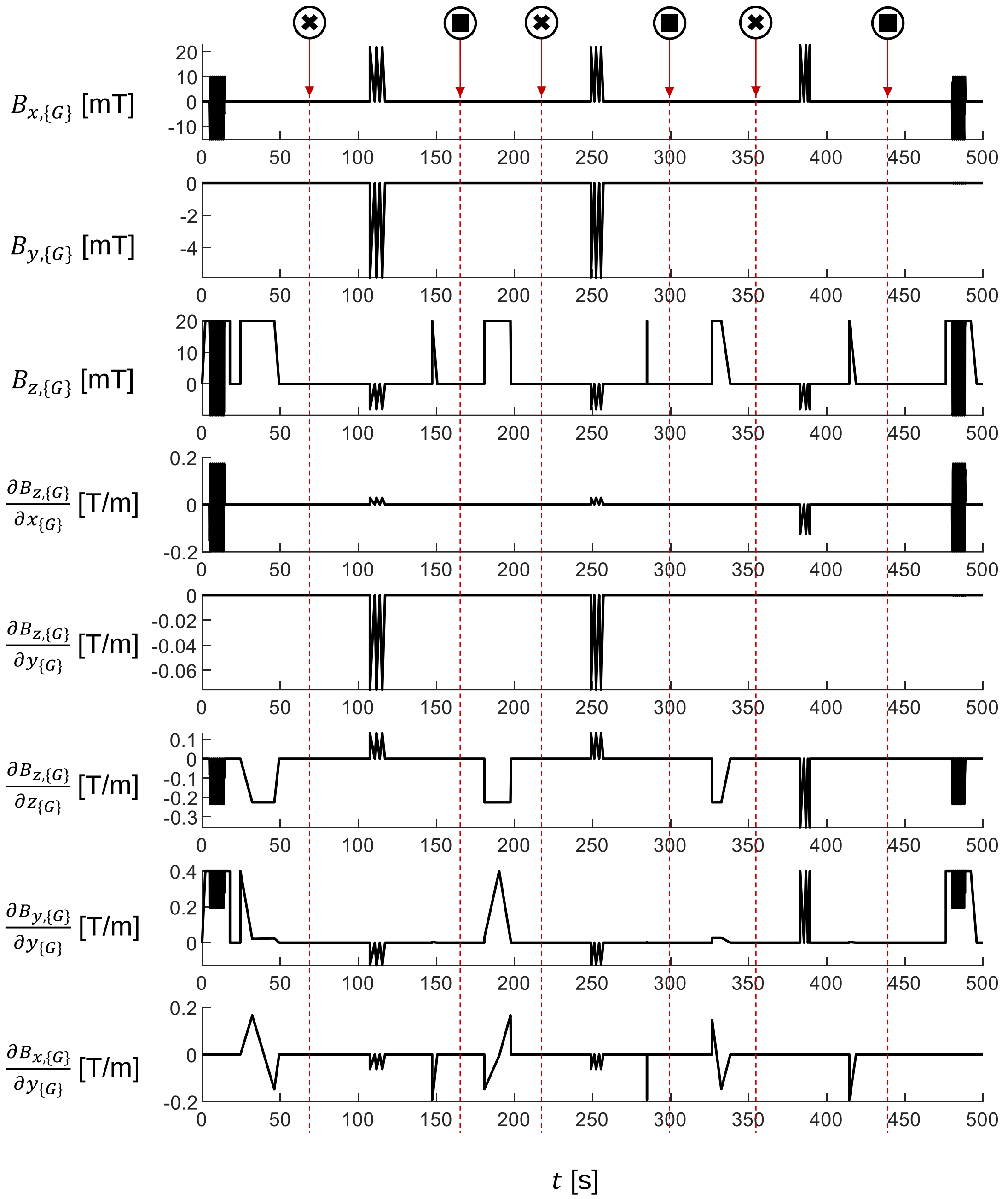

**Fig. S20 | Magnetic actuating signals for the experiment in SI Video S1 (demonstrating the soft robot's reprogramming strategies).** The time instances when the magnetizing and demagnetizing fields are applied are indicated with a cross in a circle and a square in a circle symbols, respectively. Please see Fig. S4 for the magnetizing and demagnetizing fields.

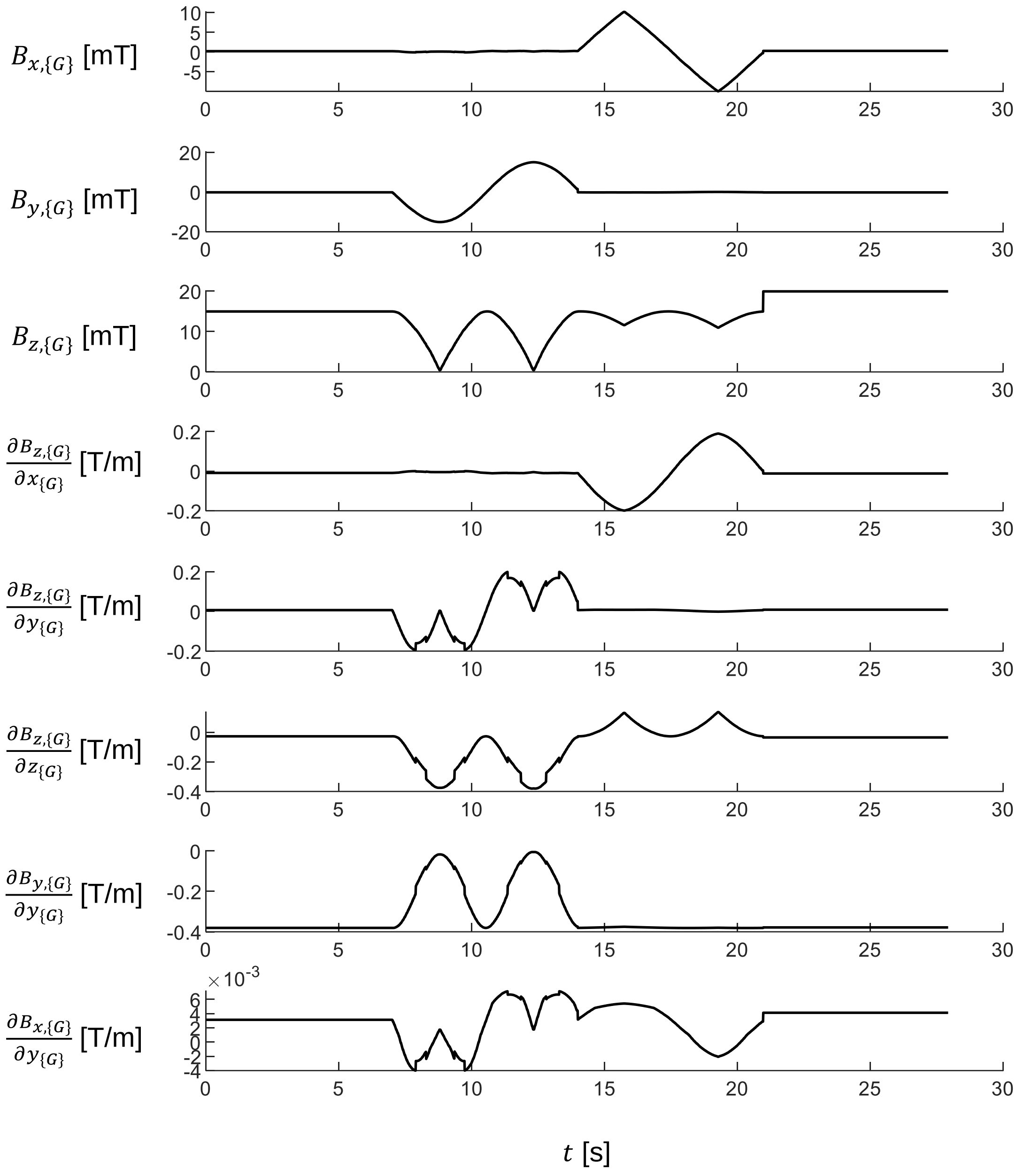

**Fig. S21 | Magnetic actuating signals for the experiment in SI Video S2 (rotating the soft robot about its $X_{\mathrm{Loco},\{L\}}$- and $Y_{\mathrm{Loco},\{L\}}$-axis).**

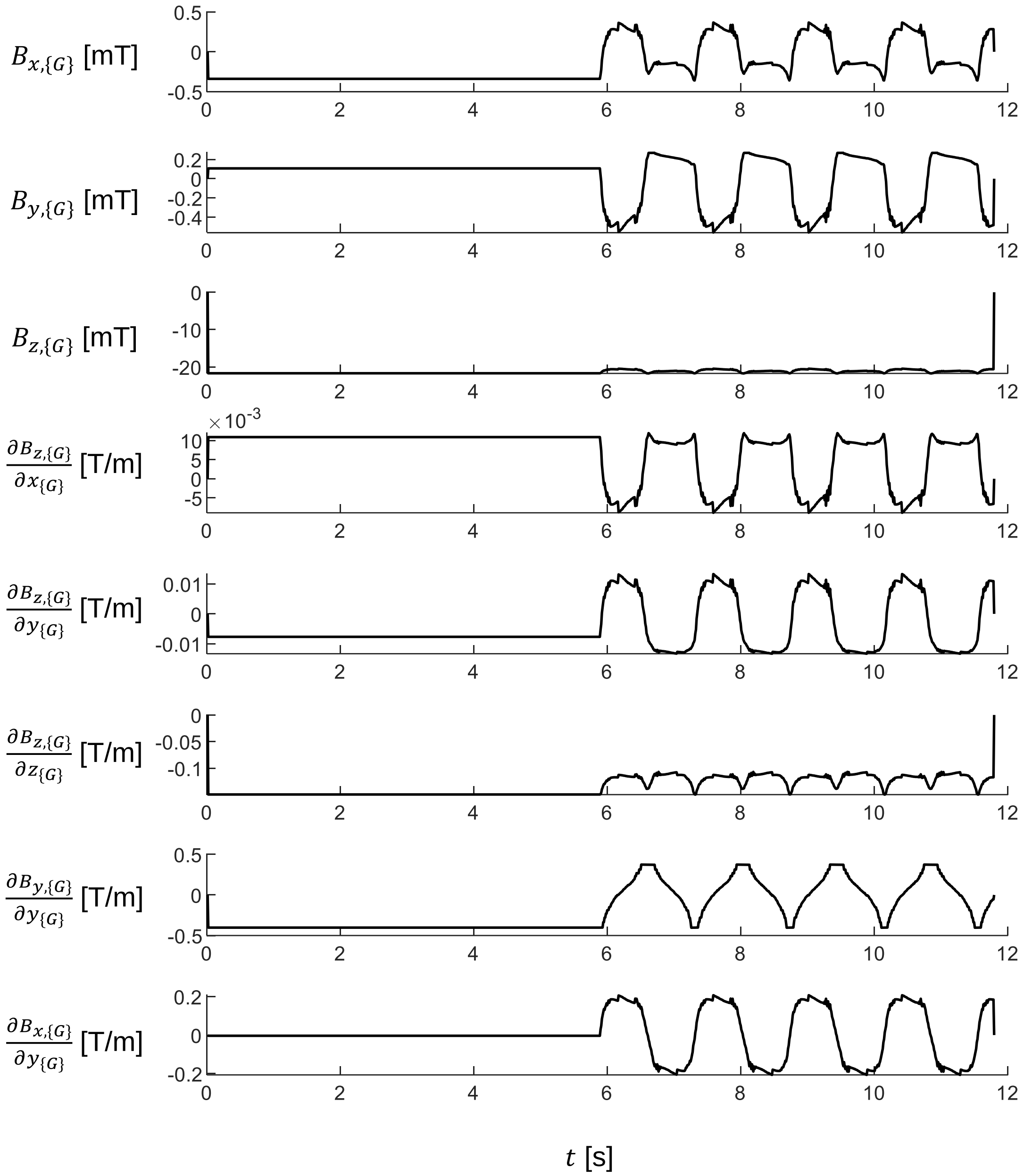

**Fig. S22 | Magnetic actuating signals for the experiment in SI Video S2 (rotating the soft robot about its $Z_{\mathrm{Loco},\{L\}}$-axis).**

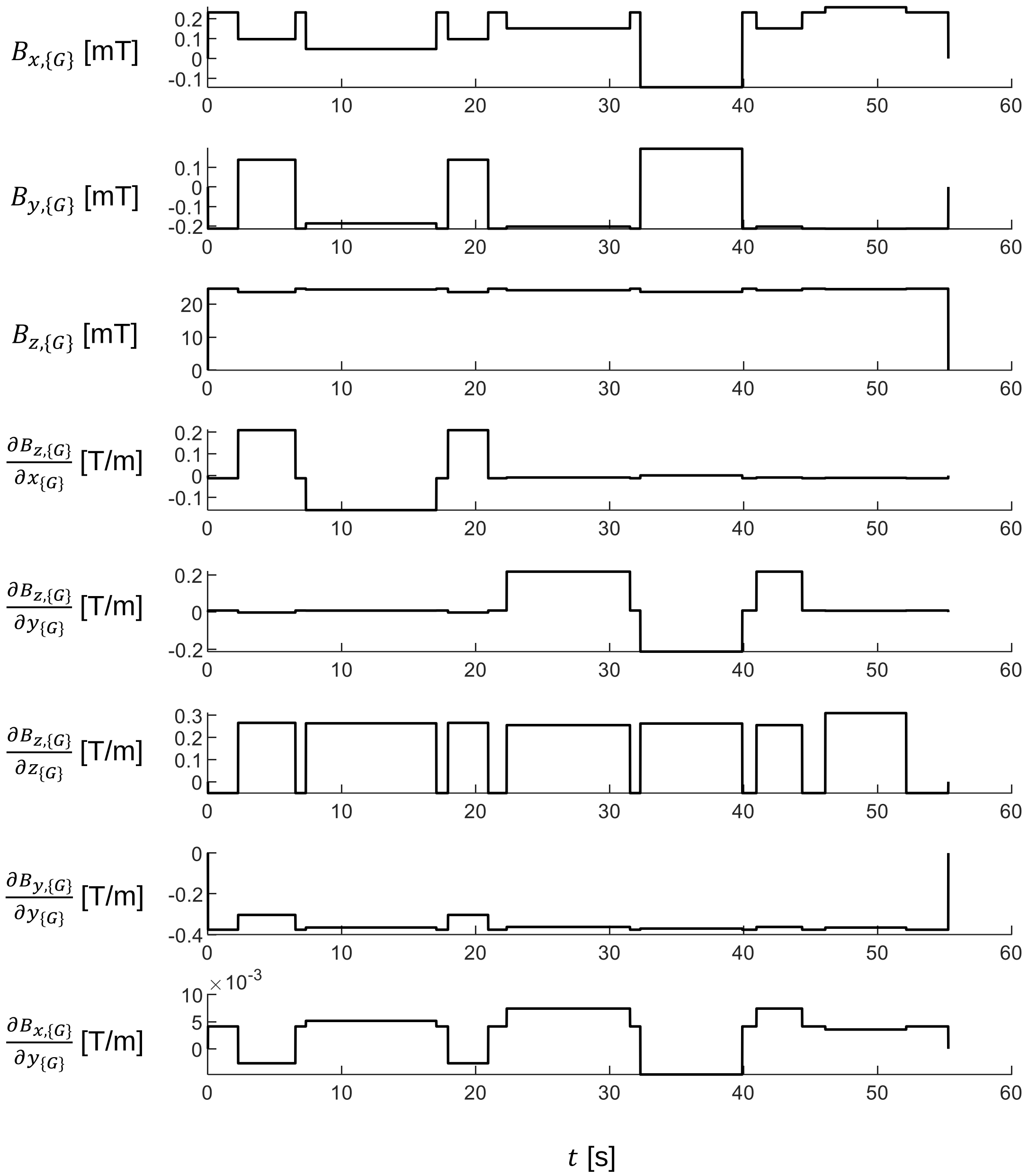

**Fig. S23 | Magnetic actuating signals for the experiment in SI Video S2 (translating the soft robot along its $X_{\mathrm{Loco},\{L\}}$-, $Y_{\mathrm{Loco},\{L\}}$-, and $Z_{\mathrm{Loco},\{L\}}$-axes).**

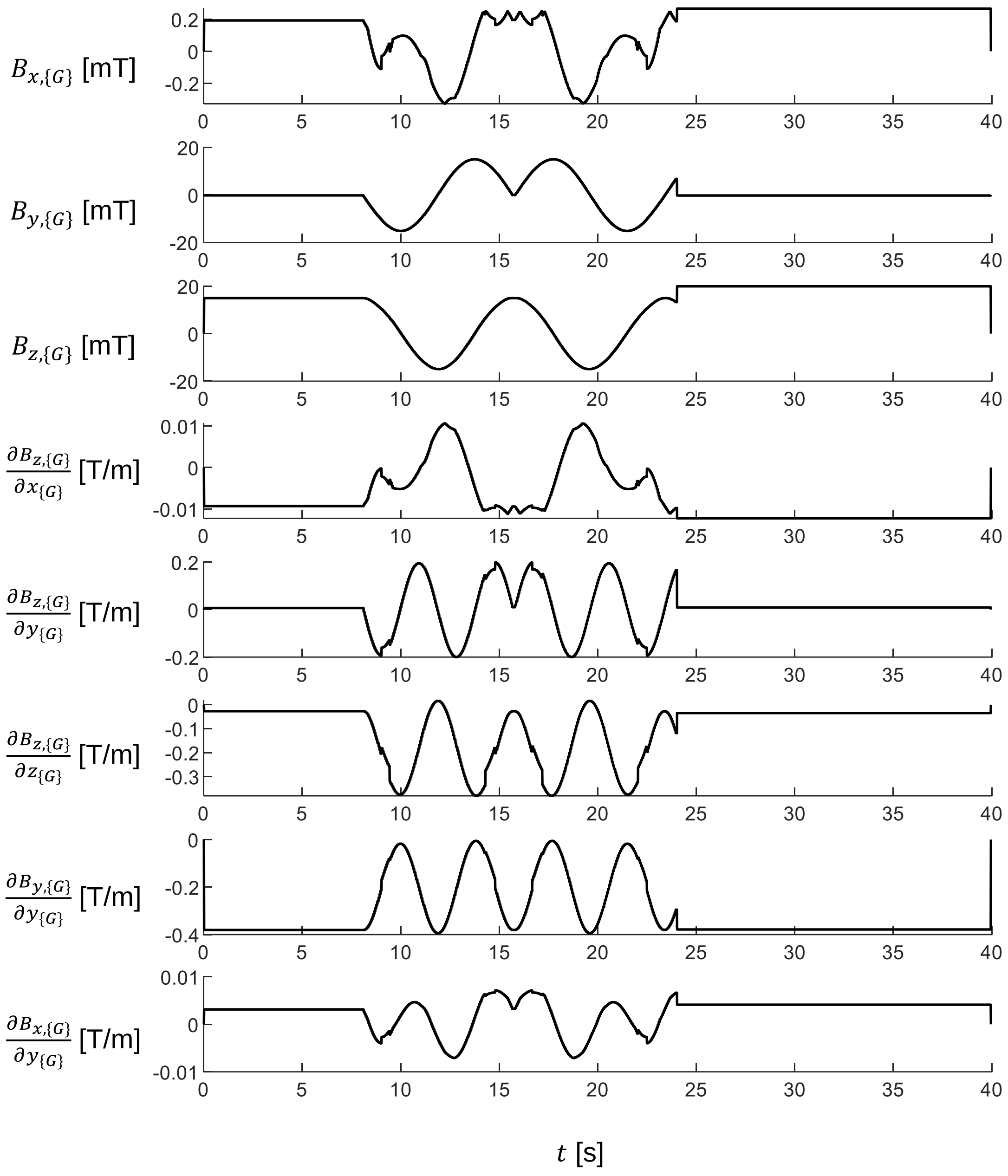

**Fig. S24 | Magnetic actuating signals for the experiment in SI Video S3 (rolling along the soft robot's length).**

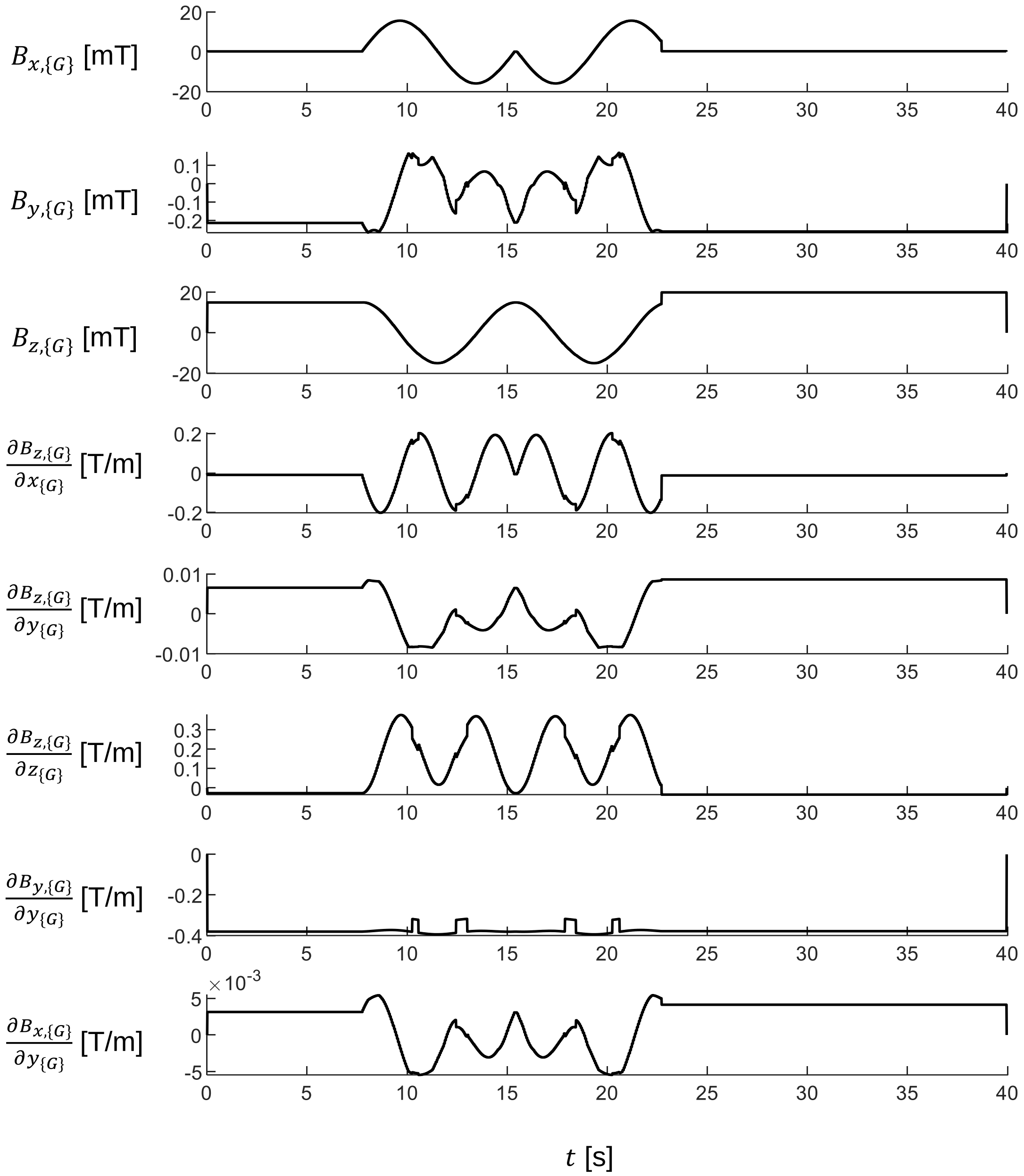

**Fig. S25 | Magnetic actuating signals for the experiment in SI Video S3 (rolling along the soft robot's width).**

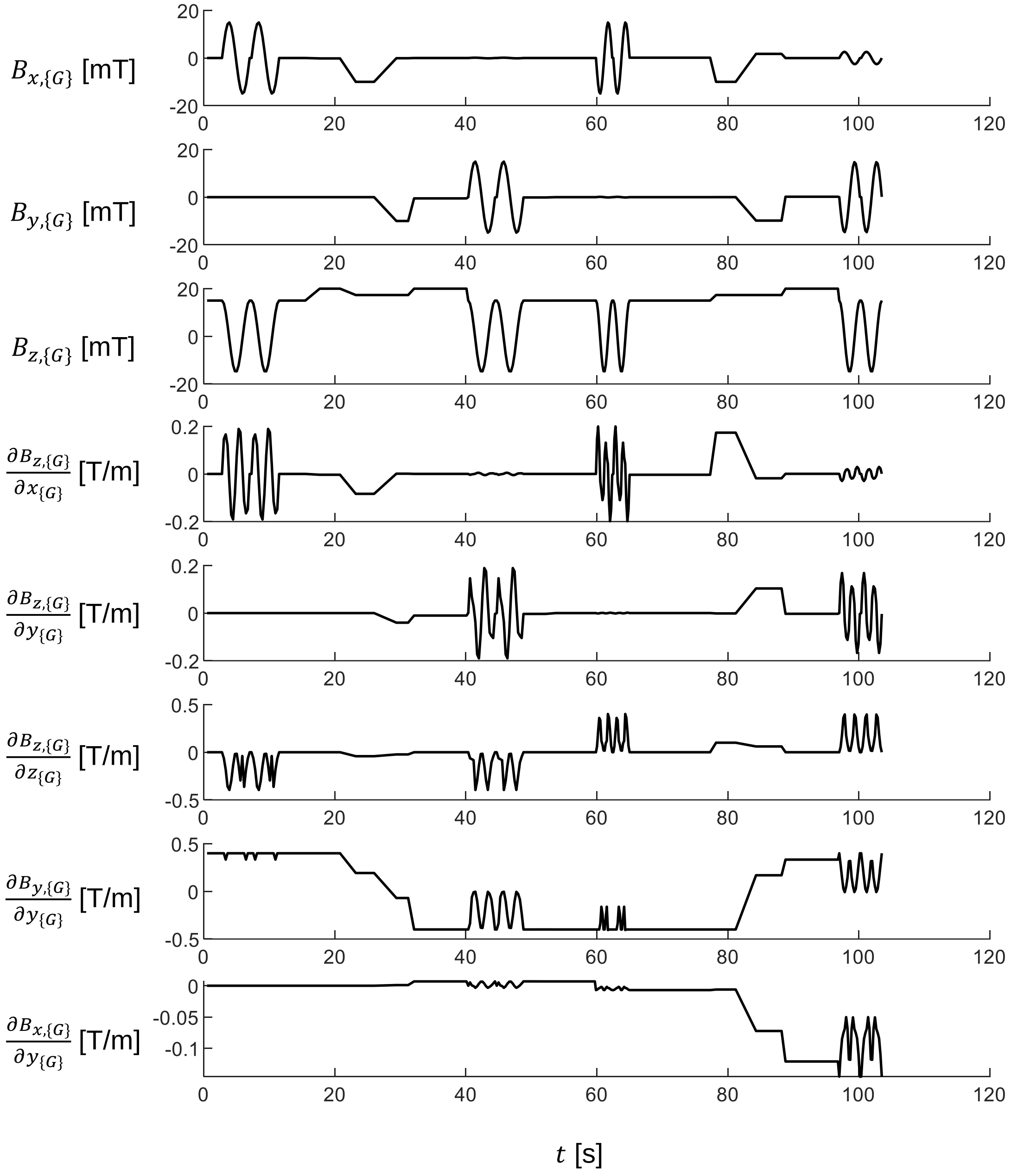

**Fig. S26 | Magnetic actuating signals for the experiment in SI Video S3 (using the rolling locomotion to track a path).**

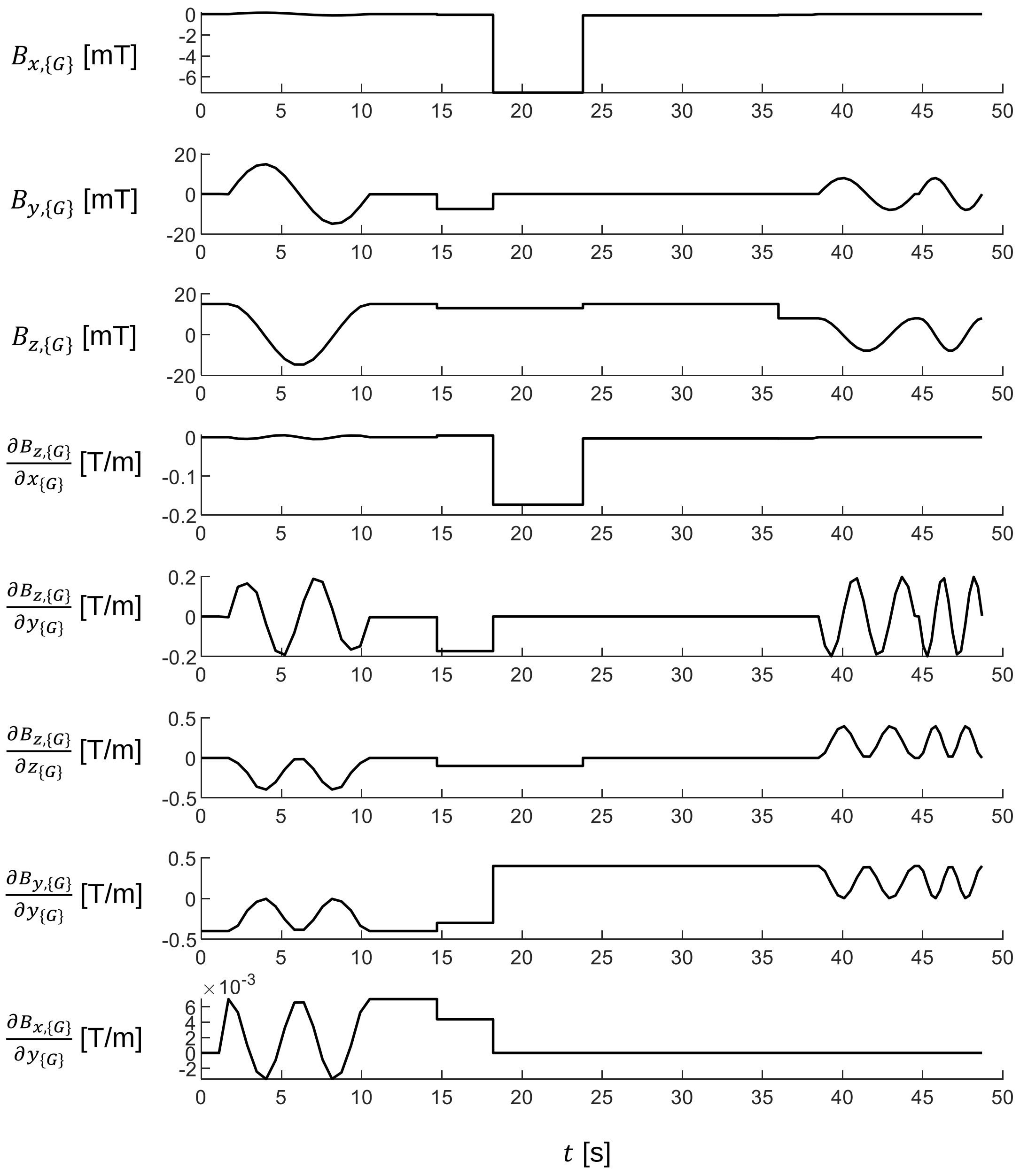

**Fig. S27 | Magnetic actuating signals for the experiment in SI Video S3 (using the rolling locomotion to negotiate across challenging barriers).**

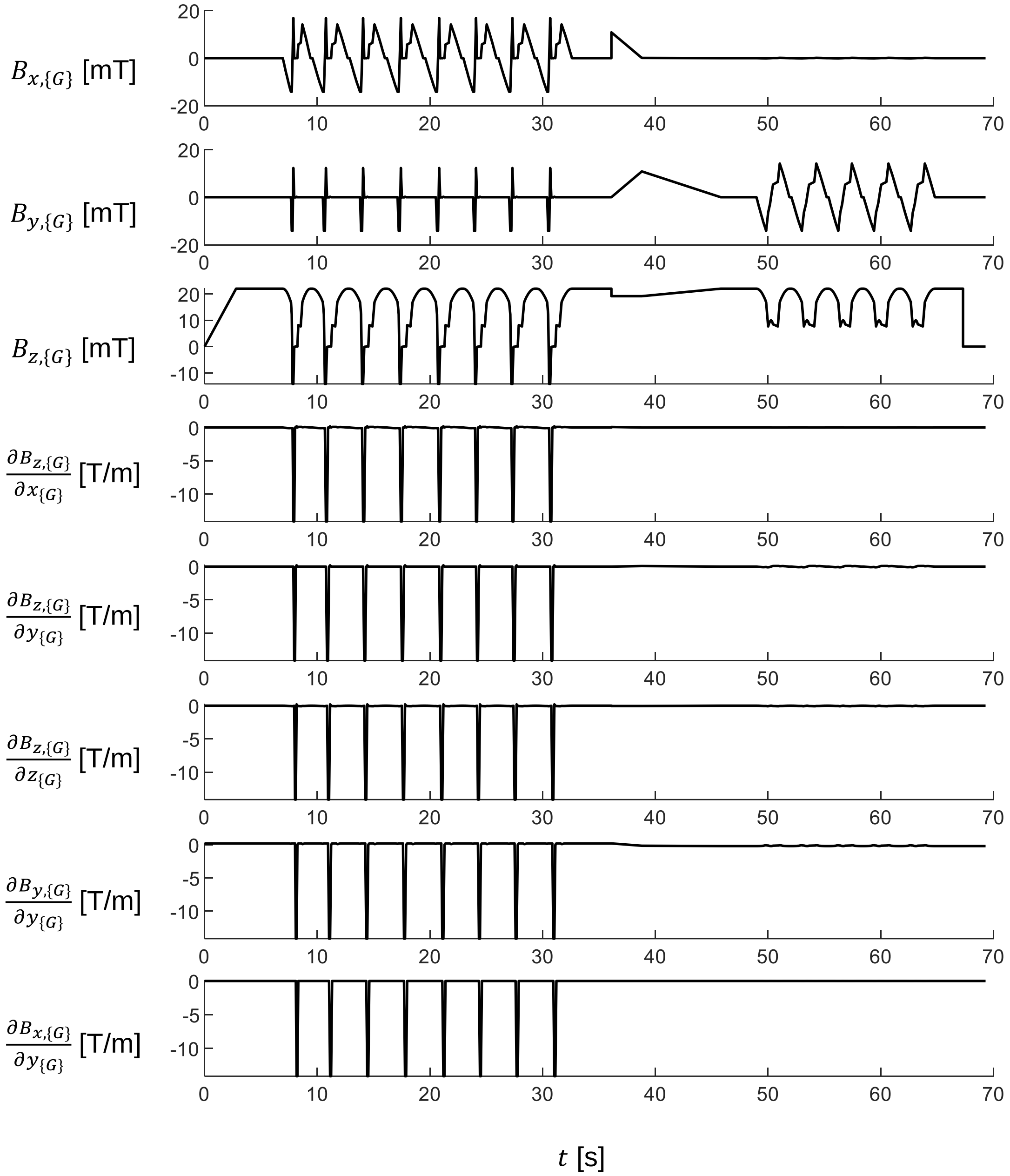

**Fig. S28 | Magnetic actuating signals for the experiment in SI Video S4 (steering the two-anchor crawling locomotion).**

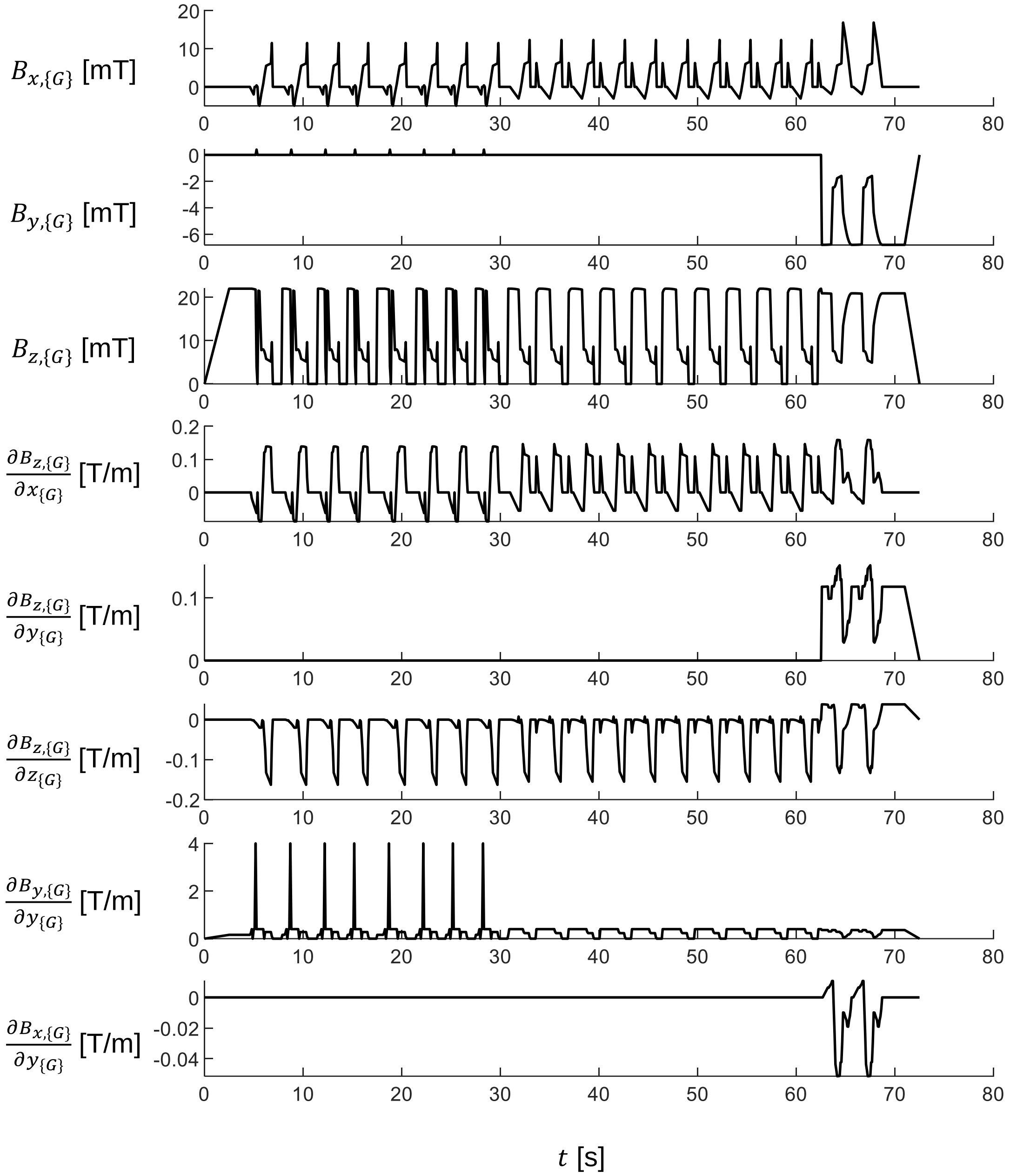

**Fig. S29 | Magnetic actuating signals for the experiment in SI Video S4 (using the two-anchor crawling gait to ascend a slope of 15°).**

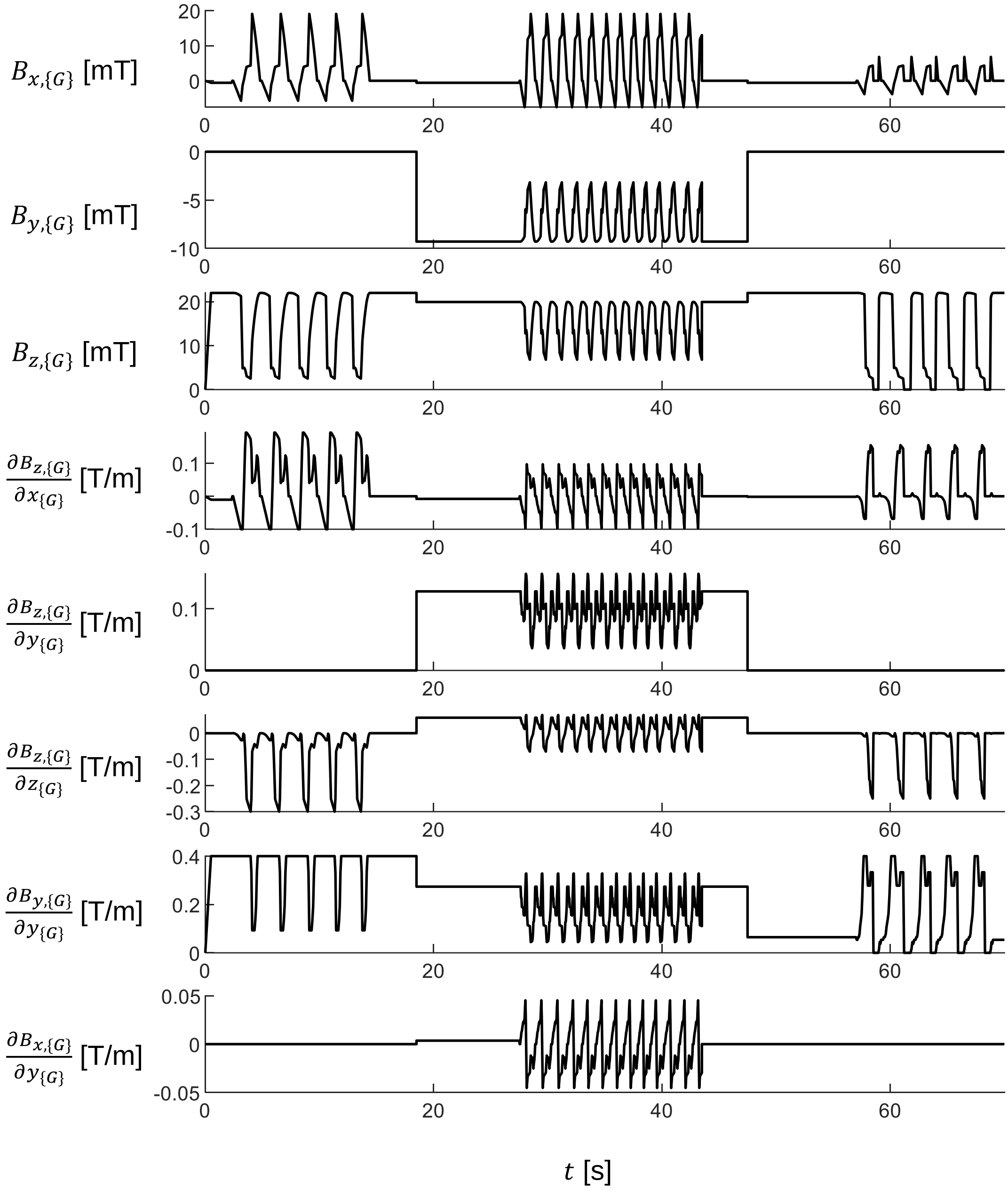

**Fig. S30 | Magnetic actuating signals for the experiment in SI Video S4 (two-anchor crawling across a challenging barrier with strict shape constraints).**

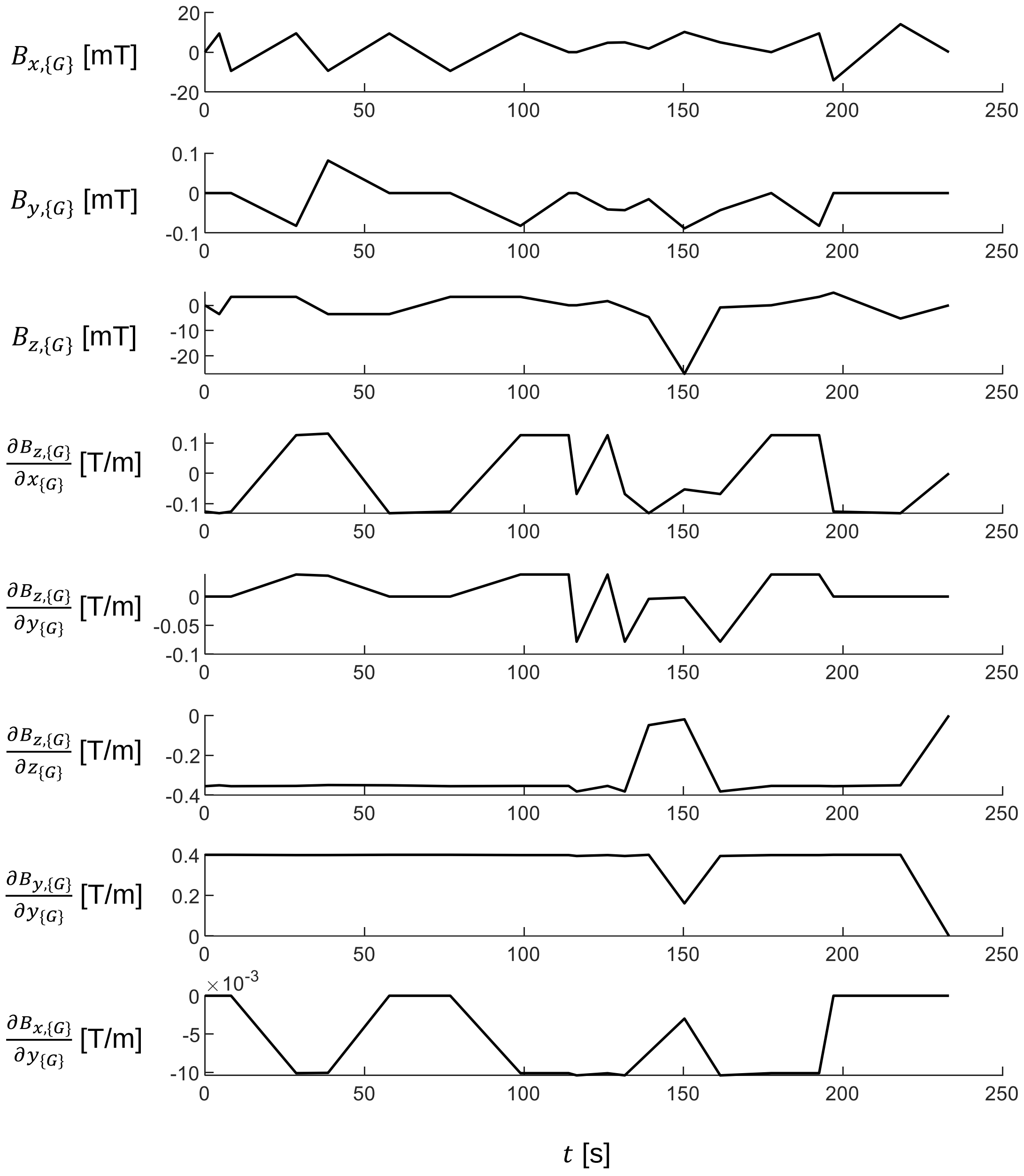

**Fig. S31 | Magnetic actuating signals for the experiment in SI Video S5 (demonstrating the drug-dispensing function).**

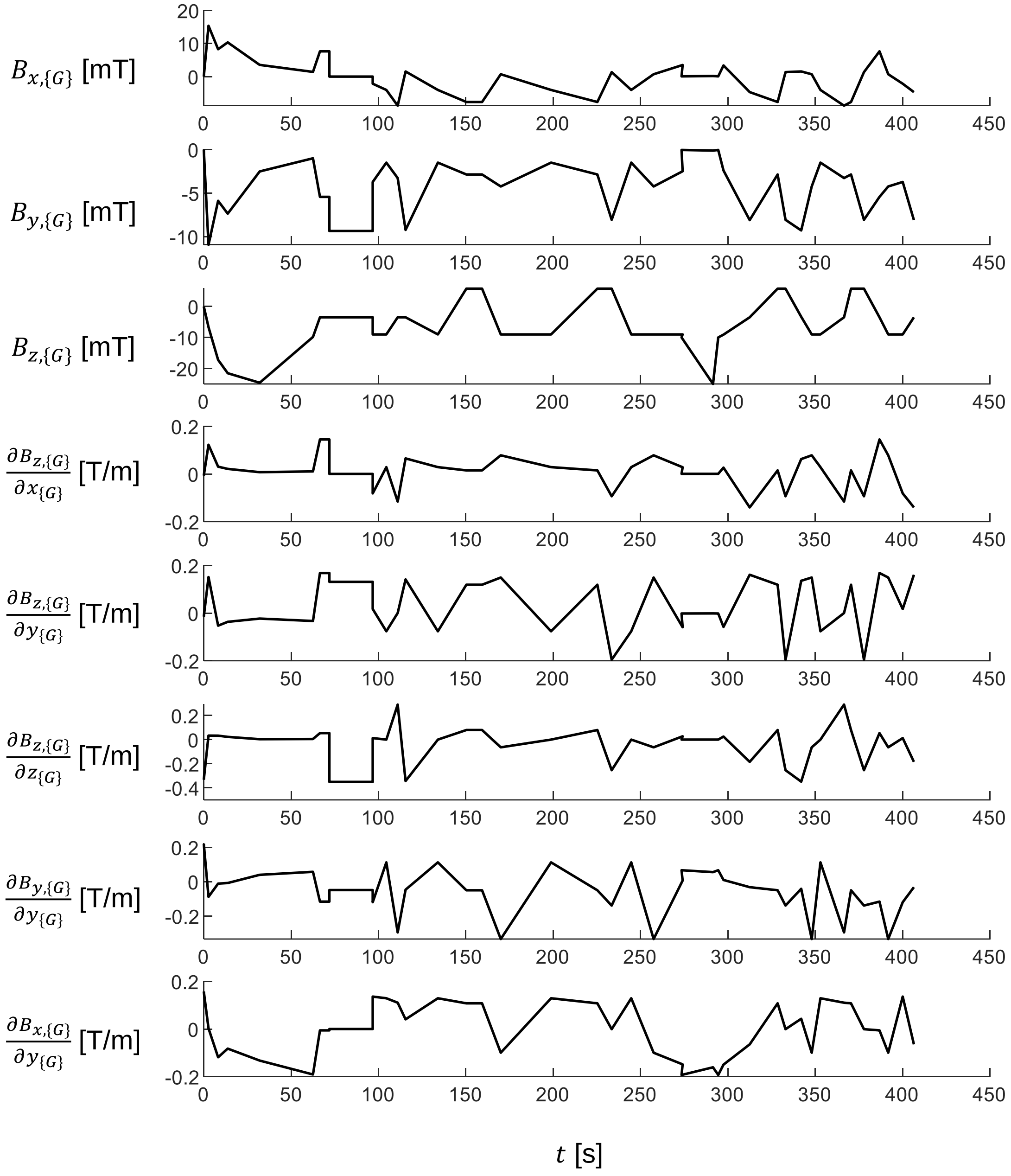

**Fig. S32 | Magnetic actuating signals for the experiment in SI Video S7 (demonstrating the pick-and-place function).**

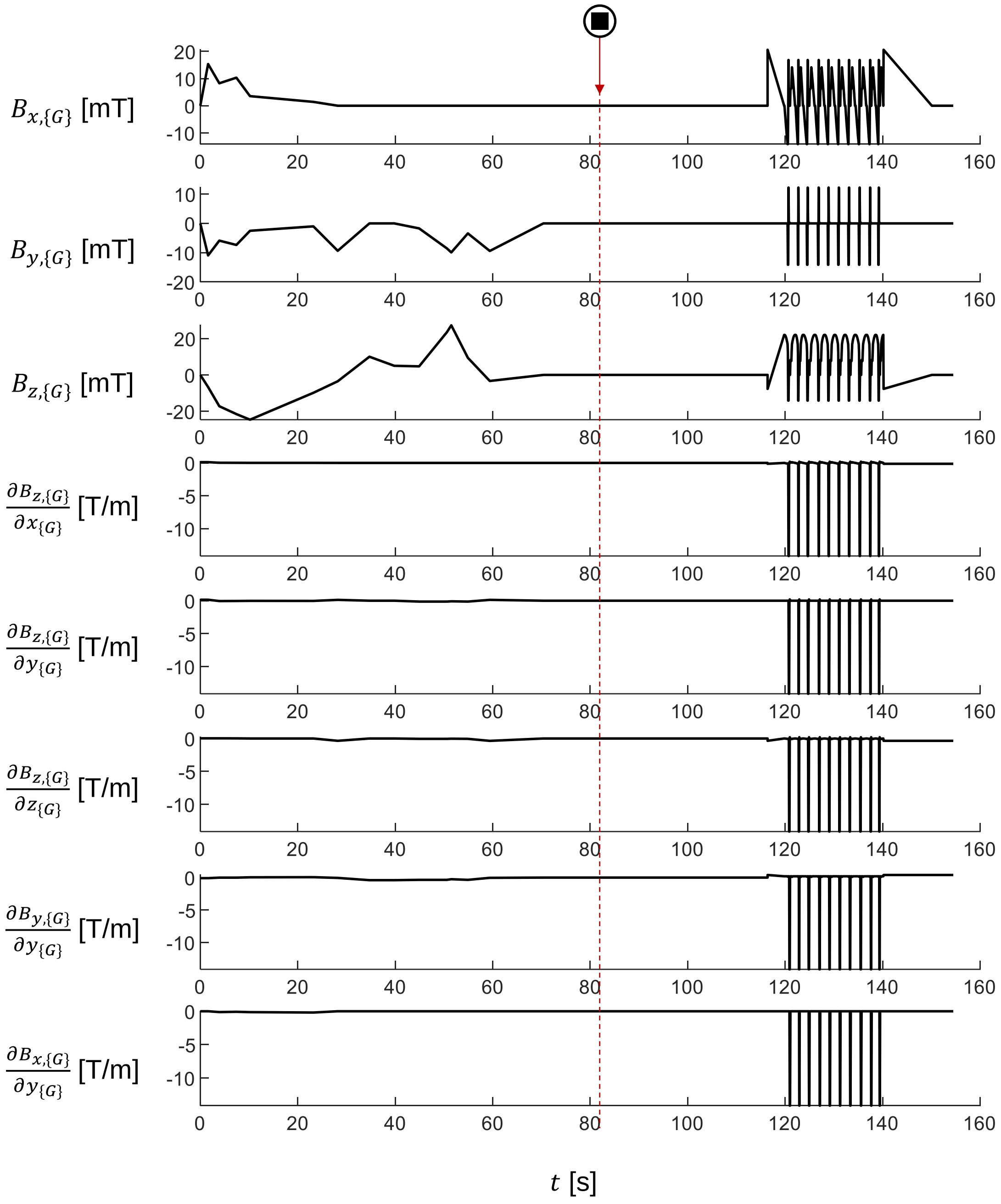

**Fig. S33 | Magnetic actuating signals for the experiment in SI Video S7 (demonstrating that the soft robot can pick, store and safely transport a synthetic excised tissue).** We indicate the time instance when the demagnetizing field is applied with a symbol that has a square in a circle. The demagnetizing field can be found in Fig. S4.

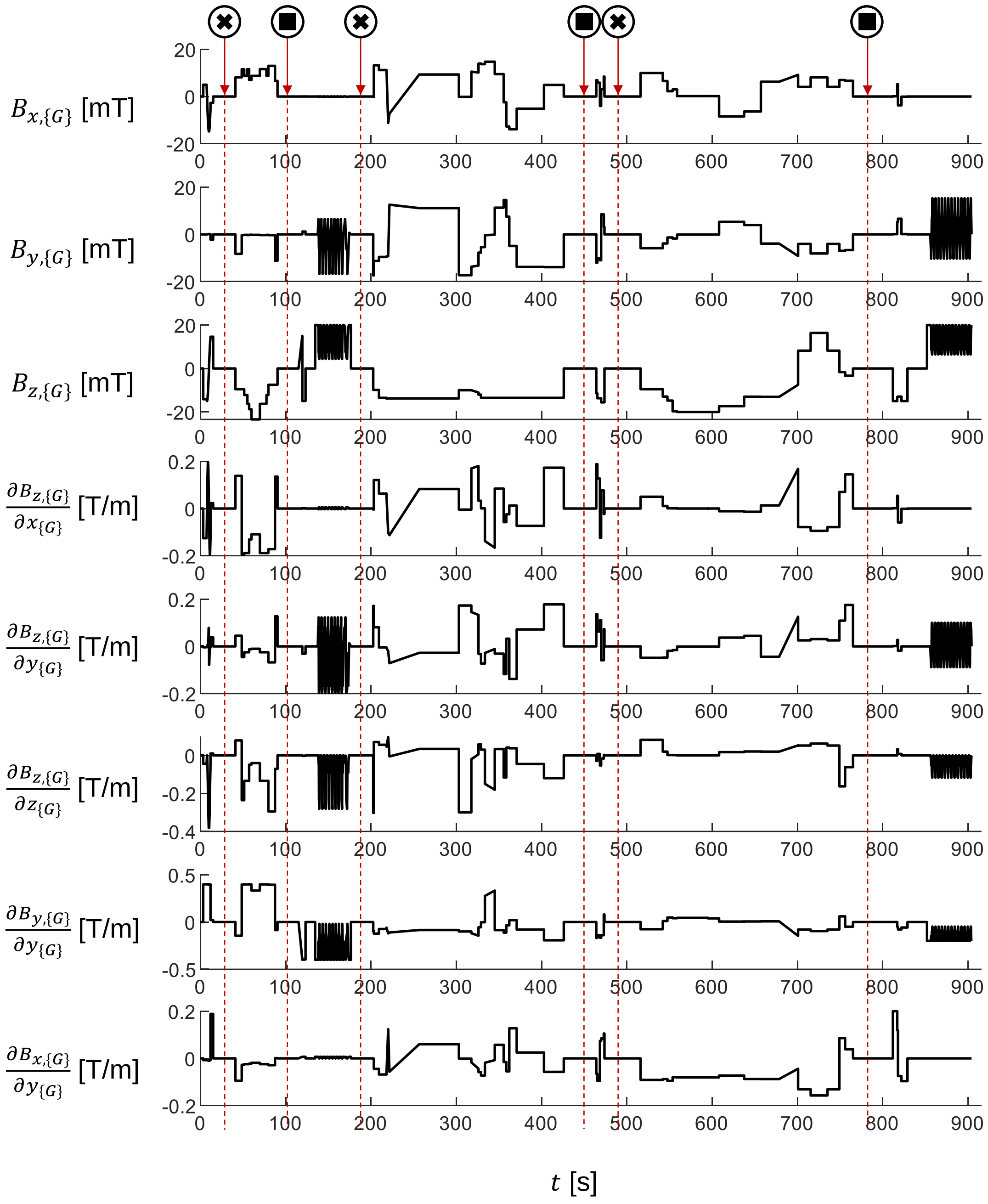

**Fig. S34 | Magnetic actuating signals for the experiment in SI Video S9 (demonstrating the soft robot's ability to perform locomotion and four surgical functions on a biological phantom).** The instances where the magnetizing and demagnetizing fields are applied are indicated with a cross in a circle and a square in a circle, respectively. (Please see Fig. S4 for the corresponding magnetic field signals they represent.)

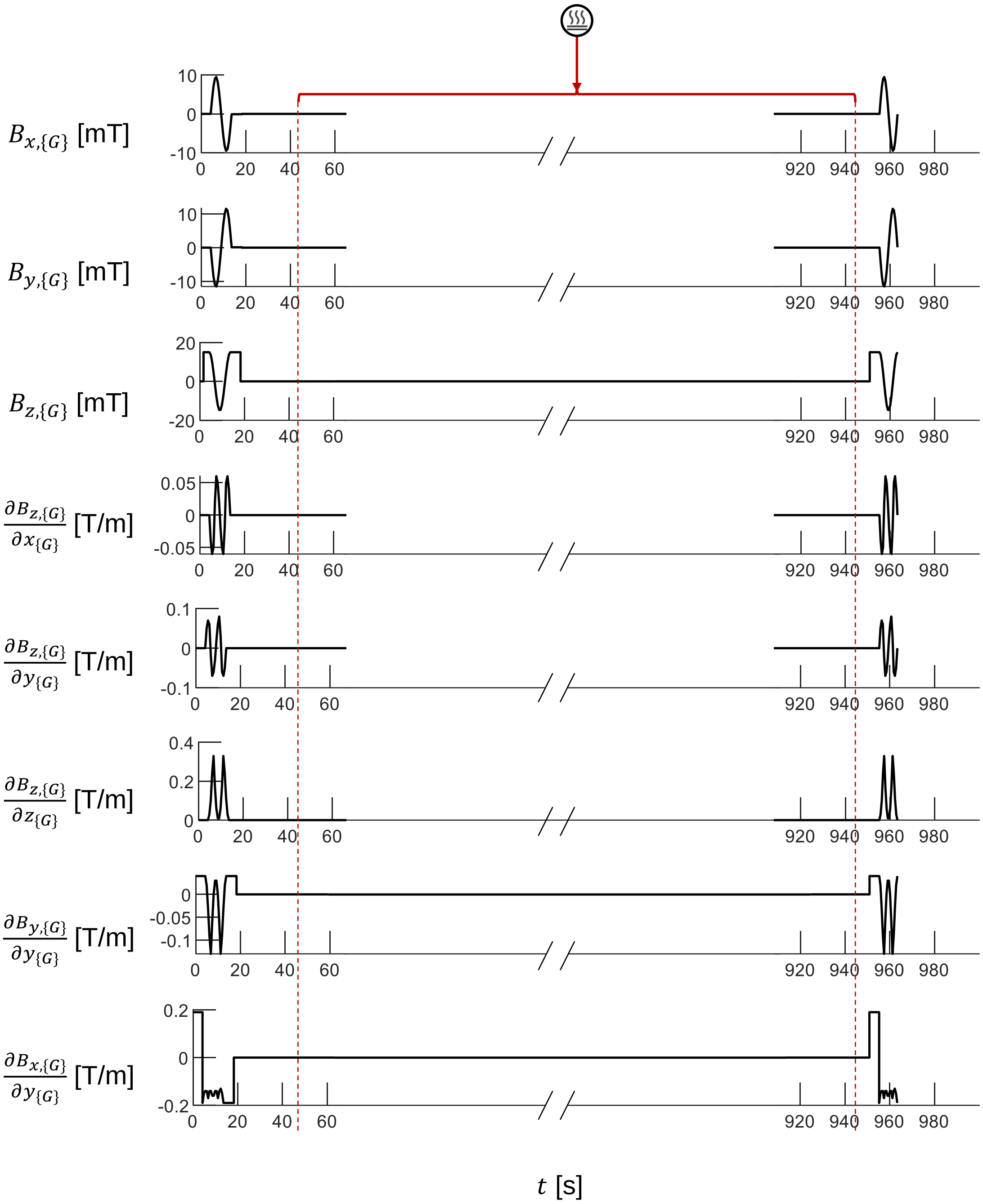

**Fig. S35 | Magnetic actuating signals for the experiment in SI Video S10 (demonstrating the soft robot's ability to perform locomotion and targeted remote heating on a biological phantom).** The period when the remote heating is executed is indicated by the heating symbol.

**Legends for SI Videos**

SI Videos S1-S10 are provided as separate files.

**SI Video S1.** Reprogramming strategy to reversibly switch the soft robot between its locomotion and function modes upon command.

**SI Video S2.** Six degrees-of-freedom motions of the soft robot.

**SI Video S3.** Rolling along the length and width of the soft robot, steering the rolling locomotion to track a path, and rolling across challenging barriers.

**SI Video S4.** Two-anchor crawling of the soft robot with steering, ascending a slope of 15°, and crossing a challenging barrier with strict shape constraints.

**SI Video S5.** Drug-dispensing mode: The soft robot rolls and dispenses a drug at the targeted location.

**SI Video S6.** Cutting mode: The soft robot penetrates biological tissues (10%wt gelatin structure).

**SI Video S7.** Gripping/storage mode: The soft robot performs a pick-and-place demonstration. The soft robot also picks, stores, and safely transports a synthetic excised tissue.

**SI Video S8.** The soft robot remotely heats a layer of thermochromic dye (blue $< 40°\mathrm{C}$ and pink $\geq 40°\mathrm{C}$) when a high frequency alternating magnetic field is applied for 1 minute.

**SI Video S9.** The soft robot performs locomotion and four reprogrammable surgical functions on a biological phantom.

**SI Video S10.** The soft robot performs locomotion and targeted remote heating on a biological phantom.